%% file: main.tex
\documentclass{article} 
\usepackage[table,RGB]{xcolor}
\usepackage{sections_iclr_cr/iclr2025_conference,times}

\usepackage[utf8]{inputenc} 
\usepackage[T1]{fontenc}    
\usepackage{url}            
\usepackage{booktabs}       
\usepackage{amsfonts}       
\usepackage{nicefrac}       
\usepackage{microtype}      
\usepackage{mathtools}
\usepackage{wrapfig}
\usepackage{bm}
\usepackage{float}
\usepackage{amsthm}
\usepackage{amsmath}
\usepackage{amssymb}
\usepackage{booktabs}
\usepackage{graphicx}
\usepackage{graphbox}
\usepackage{notes2bib}
\usepackage{multirow}
\usepackage{algorithm, algorithmic}
\usepackage{setspace}
\usepackage{makecell}
\usepackage{threeparttable}

\usepackage{tcolorbox}
\definecolor{darkerlogocolor}{RGB}{20, 0, 145}  
\newtcolorbox{ttcolorbox}[1][]{colframe=darkerlogocolor, colback=darkerlogocolor!4!white, title=#1}

\usepackage[colorlinks,linkcolor=blue]{hyperref}

\usepackage[capitalize,noabbrev]{cleveref}
\crefname{section}{Section}{Sections}
\crefname{theorem}{Theorem}{Theorems}
\crefname{lemma}{Lemma}{Lemmas}
\crefname{equation}{Equation}{Equations}
\crefname{proposition}{Proposition}{Propositions}
\crefname{claim}{Claim}{Claims}
\crefname{appendix}{Appendix}{Appendices}
\crefname{algorithm}{Algorithm}{Algorithms}
\crefname{figure}{Figure}{Figures}
\crefname{table}{Table}{Tables}
\crefname{remark}{Remark}{Remarks}
\crefname{definition}{Def.}{Definitions}
\crefname{corollary}{Corollary}{Corollaries}

\input{sections_iclr_cr/def_global}

\usepackage{etoc}
\etocdepthtag.toc{mtchapter}
\etocsettagdepth{mtchapter}{subsection}
\etocsettagdepth{mtappendix}{none}

\usepackage{xspace}
\newcommand{\revision}[1]{\textcolor{black}{#1}}
\newcommand{\propbase}{$\text{Prop}_\text{pretrained}$\xspace}
\newcommand{\propgrad}{$\text{Prop}_\text{grad}$\xspace}
\newcommand{\propamot}{$\text{Prop}_\text{amot}$\xspace}
\newcommand{\oursbase}{$\text{SMC}_\text{base}$\xspace}
\newcommand{\oursgrad}{$\text{SMC}_\text{grad}$\xspace}
\newcommand{\oursamot}{$\text{SMC}_\text{amot}$\xspace}

\title{Inference-Time Scaling of Discrete Diffusion Models via Importance Weighting and Optimal Proposal Design}

\author{%
  Zijing Ou\thanks{Equal contribution. Code is available at: \url{https://github.com/J-zin/smc_ddm_iclr}.}, \ \ Chinmay Pani\footnotemark[1], \ \ Yingzhen Li \\
  Imperial College London \\
  \texttt{z.ou22@imperial.ac.uk} \\
}

\iclrfinalcopy 
\begin{document}

\maketitle

\begin{abstract}
Discrete diffusion models have become highly effective across various domains. However, real-world applications often require the generative process to adhere to certain constraints.
To this end, we propose a Sequential Monte Carlo (SMC) framework that enables scalable inference-time control of discrete diffusion models through principled importance weighting and optimal proposal construction.
Specifically, our approach derives tractable importance weights for a range of intermediate targets and characterises the optimal proposal, for which we develop two practical approximations: a first-order gradient-based approximation and an amortised proposal trained to minimise the log-variance of the importance weights.
Empirical results across synthetic tasks, language modelling, biology design, and text-to-image generation demonstrate that our framework enhances controllability and sample quality, highlighting the effectiveness of SMC as a versatile recipe for scaling discrete diffusion models at inference time.
\end{abstract}

\input{sections_iclr_cr/01-intro}

\input{sections_iclr_cr/02-method}

\input{sections_iclr_cr/03-experiments}

\input{sections_iclr_cr/04-related_work}
\input{sections_iclr_cr/05-conclusion}

\bibliography{main}
\bibliographystyle{sections_iclr_cr/icml}

\clearpage
\appendix
\input{sections_iclr_cr/06-appendix}

\end{document}

%% file: sections_iclr_cr/def_global.tex
\usepackage{bm}
\usepackage{float}
\usepackage{amsthm}
\usepackage{amsmath}
\usepackage{amssymb}
\usepackage{booktabs}
\usepackage{graphicx}
\usepackage{graphbox}
\usepackage{notes2bib}
\usepackage{subcaption}
\usepackage{multirow}
\usepackage{algorithm, algorithmic}

\usepackage[colorlinks,linkcolor=blue]{hyperref}
\definecolor{cite_color}{HTML}{114083}
\definecolor{link_color}{RGB}{0,102,102}
\definecolor{link_color}{RGB}{153, 0,0}  
\definecolor{url_color}{RGB}{153, 102,  0}
\definecolor{emp_color}{RGB}{0,0,255}
\hypersetup{
 colorlinks,
 citecolor=cite_color,
 linkcolor=link_color,
 urlcolor=url_color}
\graphicspath{{./figs/}}

\DeclarePairedDelimiterX{\infdivx}[2]{(}{)}{%
  #1\;\delimsize\|\;#2%
  }

\newcommand*\dif{\mathop{}\!\mathrm{d}}

\def\E{{\mathbb E}}
\def\V{{\mathbb V}}

\def \x{\mathbf{x}}

\renewcommand{\mid}{|}

\newtheorem{innercustomgeneric}{\customgenericname}
\providecommand{\customgenericname}{}
\newcommand{\newcustomtheorem}[2]{%
  \newenvironment{#1}[1]
  {%
   \renewcommand\customgenericname{#2}%
   \renewcommand\theinnercustomgeneric{##1}%
   \innercustomgeneric
  }
  {\endinnercustomgeneric}
}

\newcustomtheorem{customthm}{Theorem}

\DeclareMathOperator*{\argmax}{argmax}
\DeclareMathOperator*{\argmin}{argmin}

\newtheorem{proposition}{Proposition}
\newtheorem{lemma}{Lemma}

\usepackage{thmtools}
\usepackage{thm-restate}

\usepackage{tikz}
\usetikzlibrary{decorations.pathreplacing,calc}

\newcommand{\prior}{v}
\newcommand{\mask}{\mathrm{[m]}}

\newcommand{\cat}{\mathrm{Cat}}

%% file: sections_iclr_cr/01-intro.tex
\section{Introduction}
Diffusion models \citep{sohl2015deep,ho2020denoising,song2020score} have achieved remarkable success across various domains, from image synthesis \citep{rombach2022high,esser2024scaling} to scientific applications \citep{hoogeboom2022equivariant,watson2023novo}.
Recently, advances in discrete diffusion models \citep{austin2021structured,sahoo2024simple,shi2024simplified} have established them as a powerful approach for modelling discrete data, notably in tasks such as language modelling \citep{nie2025large,zhang2025target} and code generation \citep{gat2024discrete,gong2025diffucoder}.

Despite their impressive capabilities, pretrained diffusion models often need to generate samples that meet specific downstream constraints. 
For example, text-to-image generation requires images aligned with human preferences \citep{black2023training,fan2023dpok,uehara2024feedback}, while protein generation demands stability or desired binding affinity \citep{verkuil2022language,uehara2025reward}.
To address this challenge, existing approaches mainly fall into two categories: i) fine-tuning and ii) guidance methods.
Fine-tuning methods, including techniques such as steering \citep{rector2024steering}, reinforcement learning \citep{zekri2025fine}, and direct backpropagation \citep{wang2024fine}, have demonstrated promising results. Nevertheless, these methods often suffer from reward over-optimisation, which can compromise sample quality and diversity.
On the other hand, guidance and sampling methods \citep{li2024derivative,gruver2023protein,nisonoff2024unlocking,guo2024plug,uehara2025rewarditer} provide training-free alternatives that are easier to deploy, but they often suffer from reward under-optimisation. This limits their ability to enforce correct alignment, resulting in outputs that may not fully meet complex constraints.

In this paper, with a primary focus on discrete diffusion models, we propose a Sequential Monte Carlo (SMC) \citep{del2006sequential} framework for test-time inference.
By leveraging SMC, an asymptotically unbiased sampler, our approach enables test-time scaling, effectively addressing the over-optimisation issues commonly encountered by fine-tuning methods.
Moreover, we propose a learnable amortised proposal to approximate the optimal SMC proposal, which mitigates the under-optimisation problems often associated with guidance-based methods, thereby improving both scalability and efficacy during inference.
In summary, our contributions include:
\begin{itemize}
    \item We propose a simple SMC framework for discrete diffusion models. By leveraging tractable importance weights, we show that SMC provides a general recipe for test-time scaling, enhancing classifier-free guidance and enabling effective reward alignment.
    \item We propose two approximately optimal proposals: a first-order approximation and a learnable amortised proposal. The latter is optimised by minimising the log-variance of importance weights, leading to substantial improvements in the effectiveness of SMC.
    \item We demonstrate the versatility of the proposed approach across a broad range of applications, including language modelling, biology design, and text-to-image generation, highlighting its ability to consistently improve performance and generalise across diverse domains.
\end{itemize}

%% file: sections_iclr_cr/02-method.tex
\section{Background}

We first introduce the main preliminaries: discrete diffusion models and Sequential Monte Carlo.

\subsection{Discrete Diffusion Models}
\label{sec: discrete-diffusion-models}

Discrete diffusion models \citep{austin2021structured} define a forward nosing process that interpolates between the original data distribution and a fixed prior $\prior \in \Delta^V$ on the $V$-simplex:
\begin{align} \label{eq:diffusion-forward}
    p (x_t | x_0) = \cat(x_t ; \alpha_t x_0 + (1-\alpha_t) \prior),
\end{align}
where $\alpha_t$ is a monotonically decreasing schedule from 1 to 0 such that $x_T \sim \cat(v)$.
Masked diffusion models \citep{sahoo2024simple, shi2024simplified} are a special case that use a mask token $\mask$ as the prior, with the induced posterior taking the form of
\begin{equation}
    p(x_{t-1} | x_t, x_0) =
    \begin{cases}
        \cat(x_{t-1}; x_t) & x_t \neq \mask,\\
        \cat\left(x_{t-1}; \frac{(1-\alpha_{t-1})\mask + (\alpha_{t-1} - \alpha_t) x_0}{1 - \alpha_t}\right) & x_t = \mask
    \end{cases}
\end{equation}
Since $x_0$ is not available during inference, the reverse unmasking process is parametrised as $p_\theta(x_{t-1} | x_t) = p(x_{t-1} | x_t, \mu_\theta(x_t))$,
where $\mu_\theta(x_t)$ is a denoising model that predicts the clean data $x_0$. The model is trained by minimising the cross-entropy loss
\begin{align} \label{eq:mdm-loss}
    \mathcal{L}(x_0;\theta) = \sum_{t=1}^{T} \frac{\alpha_t^\prime}{1 - \alpha_t}  \E_{p(x_t | x_0)} [-\log (x_0^T \mu_\theta (x_t))] \dif t,
\end{align}
which is equivalent, if $T \!\to\! \infty$, to the negative evidence lower bound of the log-likelihood $\log p_\theta(x_0)$. 

\subsection{Importance Sampling and Sequential Monte Carlo}
Consider the Monte Carlo integration problem $\E_{\pi(x_t)}[\delta(x_t)]$, where sampling from the target distribution $\pi$ is intractable. Importance sampling \citep{robert1999monte} alleviates this issue by introducing a proposal distribution $q$, allowing the expectation to be rewritten as
\begin{align}
    \E_{\pi(x_t)}[\delta (x_t)] = \E_{q(x_{t:T})}\left[ \frac{\pi(x_{t:T})}{q(x_{t:T})} \delta(x_t) \right] \approx \frac{1}{N} \sum_{i=1}^N \frac{\pi(x_{t:T}^{(i)})}{q(x_{t:T}^{(i)})} \delta(x_t^{(i)}), \quad x_{t:T}^{(i)} \sim q(x_{t:T}).
\end{align}
While conceptually simple, importance sampling often suffers from high variance. To address this limitation, Sequential Monte Carlo (SMC) \citep{del2006sequential} extends importance sampling by incorporating resampling and sequential weighting strategies across the path, thereby reducing variance in practice.
In SMC, a key intuition is the recursive formulation of the importance weight
\begin{align} \label{eq:iw_recursive}
    w_{t-1} (x_{t-1:T}^{(i)}) \triangleq \frac{\pi(x_{t-1:T}^{(i)})}{q(x_{t-1:T}^{(i)})} = \frac{\pi(x_{t-1}^{(i)} | x_{t:T}^{(i)})\pi(x_{t:T}^{(i)})}{q(x_{t-1}^{(i)} | x_{t:T}^{(i)})q(x_{t:T}^{(i)})} = \frac{\pi(x_{t-1}^{(i)})}{\pi(x_t^{(i)})} \frac{\gamma(x_t^{(i)} | x_{t-1}^{(i)})}{q(x_{t-1}^{(i)} | x_t^{(i)})} w_t (x_{t:T}^{(i)}),
\end{align}
where we leverage the Markovian assumption that $\pi(x_{t:T}^{(i)}) = \pi(x_t^{(i)})\prod_{k=t}^{T-1} \gamma(x_{t+1}^{(i)}|x_t^{(i)})$ for arbitrary forward kernel $\gamma$ and thus $\pi(x_{t-1}^{(i)} | x_t^{(i)}) = \pi(x_{t-1}^{(i)}) \gamma(x_t^{(i)} | x_{t-1}^{(i)}) / \pi(x_t^{(i)})$.
The recursion of importance weight underlies the iterative procedure of SMC. Concretely, The procedure initialises begins by $N$ particles $x_T^{(i)} \sim q(x_T)$ with weights $w_T^{(i)} \leftarrow \pi(x_T^{(i)}) / q(x_T^{(i)})$. For each step $t = T, \dots, 1$ and particles $i=1,\dots,N$, SMC proceeds as follows: i) resample ancestor according to the weights $\{w_{t}^{(i)}\}_{i=1}^N$; ii) propagate new particles via $x_{t-1}^{(i)} \sim q(x_{t-1} | x_{t})$; and iii) updating the weights as $w_{t-1}^{(i)} \!\leftarrow\! [\pi(x_{t-1}^{(i)}) \pi(x_{t}^{(i)} | x_{t-1}^{(i)})] / [\pi(x_{t}^{(i)}) q(x_{t-1}^{(i)} | x_{t}^{(i)})]$.
The resulting collection of weighted particles provides an asymptotically consistent approximation of the intermediate target distribution $\pi(x_t)$.

\section{Sequence Monte Carlo for Discrete Diffusion Models}
\label{sec:methodology}

Given a pretrained discrete diffusion model $p_\theta (x_t)$, we consider sampling from modified target distributions that enable inference-time control. These targets include: i) product distributions, a general form underlying classifier free guidance \citep{ho2022classifier}, defined as $\pi (x_t) \propto p_{\theta_1}^\alpha (x_t) p_{\theta_2}^\beta (x_t)$; and ii) reward-tilting distributions, expressed as $\pi (x_t) \propto p_\theta (x_t) \exp(r(x_t))$.
In the following section, we introduce how to construct tractable importance weights by carefully selecting forward kernels $\gamma (x_t \mid x_{t-1})$, and show their connection to existing SMC formulations for continuous-time discrete diffusion models. We then discuss the choice of proposal distributions, which play a central role in balancing variance reduction with computational efficiency.

\subsection{Importance Weight: Tractability with Pretrained Diffusion Models}
To perform SMC, one must evaluate the importance weight from \cref{eq:iw_recursive} at each step $t$. While the forward kernel $\gamma (x_t | x_{t-1})$ and the proposal $q(x_{t-1} | x_t)$ can be chosen flexibly, the ratio of intermediate targets $\frac{\pi(x_{t-1})}{\pi(x_t)}$ is generally intractable in diffusion models.
With a well-trained diffusion model $p_\theta$, however, this ratio can be approximated via detailed balance $\frac{p_\theta (x_{t-1})}{p_\theta (x_{t})} \approx \frac{p_\theta (x_{t-1} | x_t)}{p(x_t | x_{t-1})}$, where $p(x_t | x_{t-1})$ denotes the forward noising process and $p_\theta (x_{t-1} | x_t)$ is the learned reverse counterpart. Under this approximation, the importance weight for the product target takes the form
\begin{align} \label{eq:important-weight-anneal-comp}
    \text{product:}\ \frac{p_{\theta_1}^\alpha (x_{t-1} | x_t) p_{\theta_2}^\beta (x_{t-1} | x_t)}{p_{1}^\alpha (x_t | x_{t-1}) p_{2}^\beta (x_t | x_{t-1})} \frac{\gamma(x_t | x_{t-1})}{q(x_{t-1} | x_t)}.
\end{align}
Although tractable, this weight inevitably introduces approximation error unless the reverse model is perfectly trained, due to the mismatch between the forward and backward processes.
In contrast, for the reward-tilting, the error can be eliminated by setting $\gamma(x_t | x_{t-1}) = p(x_t | x_{t-1})$, yielding
\begin{align} \label{eq:important-weight-tilting}
    \text{reward-tilting:}\ \frac{\exp(r(x_{t-1}))}{\exp(r(x_t))} \frac{p_\theta (x_{t-1} | x_t)}{p (x_t | x_{t-1})} \frac{\gamma(x_t | x_{t-1})}{q(x_{t-1} | x_t)} =  \frac{\exp(r(x_{t-1}))}{\exp(r(x_t))}\frac{p_\theta (x_{t-1} | x_t)}{q(x_{t-1} | x_t)}
\end{align}
\begin{wrapfigure}{r}{0.38\linewidth}
    \vspace{-4mm}
    \centering
        \begin{minipage}[t]{1.\linewidth}
            \centering
            \begin{minipage}[t]{0.48\linewidth}
                \centering
                \includegraphics[width=.99\linewidth]{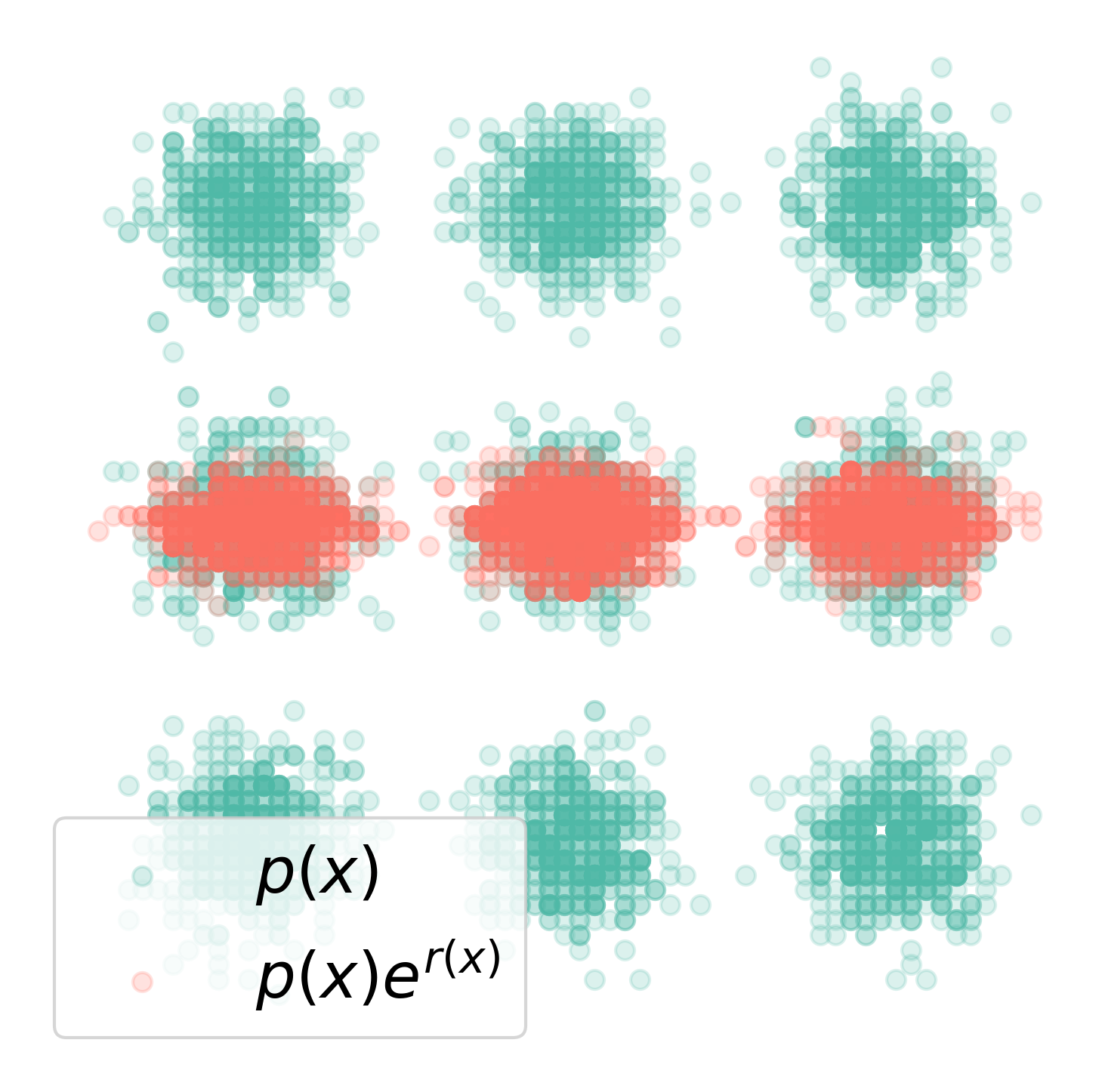}
            \end{minipage}
            \begin{minipage}[t]{0.48\linewidth}
                \centering
                \includegraphics[width=.99\linewidth]{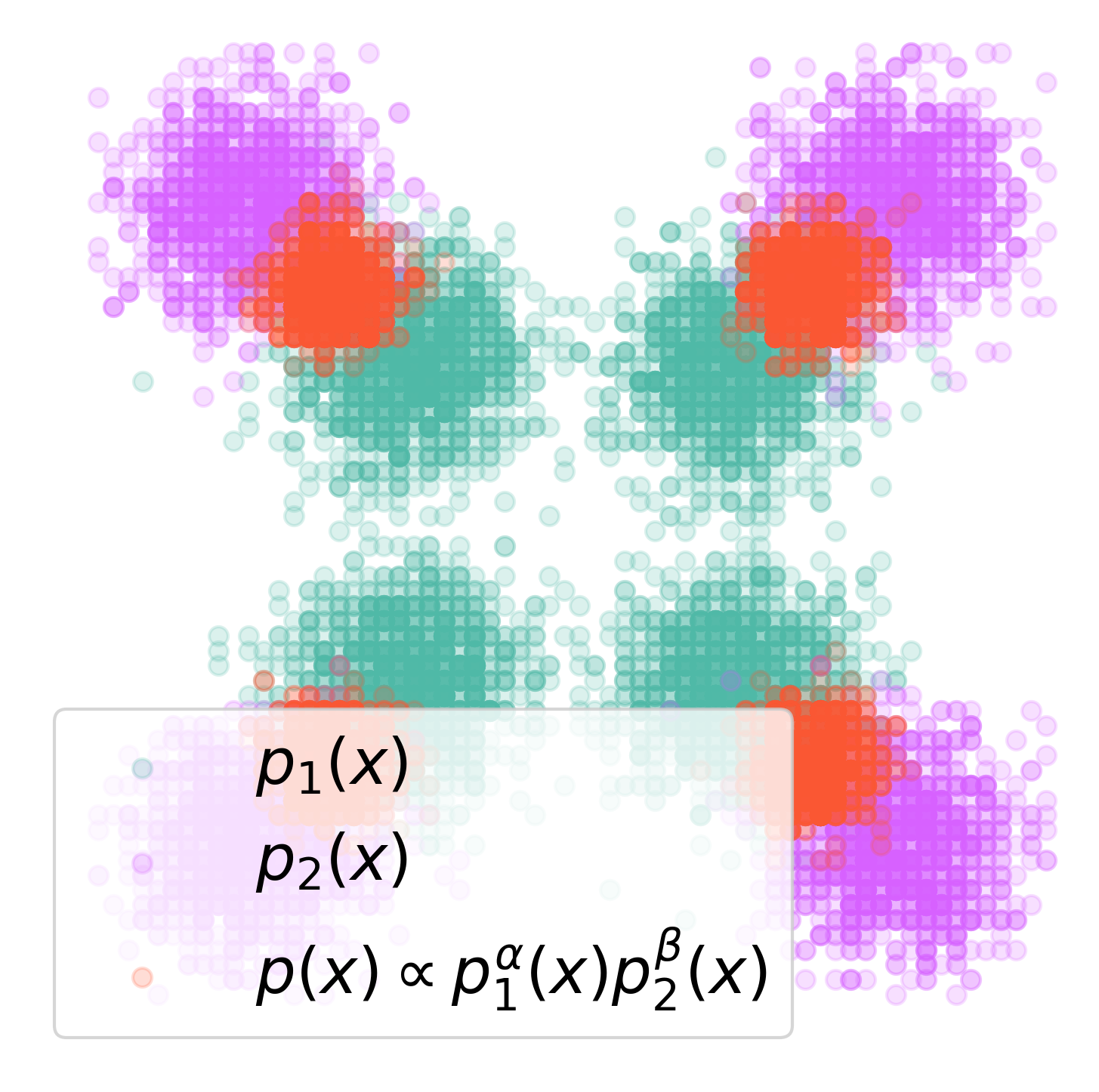}
            \end{minipage}
        \end{minipage}
    \vspace{-4mm}
    \caption{SMC results for the reward-tilting and product target distributions.}
    \label{fig:smc_diff_toy}
    \vspace{-4mm}
\end{wrapfigure}
It is noteworthy that this cancellation is not applicable to the product distributions, since the normalising constant of $\gamma$ can not be cancelled even if we choose $\gamma \propto p_1^\alpha (x_t | x_{t-1}) p_2^\beta (x_t | x_{t-1})$. 
Nevertheless, as illustrated in \cref{fig:smc_diff_toy}, SMC with these tractable importance weights performs well across both two settings on 2D toy examples.
Moreover, although we primarily focus on discrete-time diffusion, the proposed SMC method can be extended seamlessly to the continuous-time setting, as established in the following proposition.
\begin{restatable}[SMC for Continuous-Time Discrete Diffusion]{proposition}{restapropcontsmc}  \label{prop:smc_cont}
    Let $R_t$ be the rate matrix generating the forward transition kernel $\gamma(x_t \mid x_{t-\Delta t})$, and $\hat{R}_t$ be its counterpart associated with the backward proposal kernel $q(x_{t-\Delta t} \mid x_t)$, where $\Delta t \to 0$ is the infinitesimal time increment. Then, the importance weight at time $t$ is given by $w_t = \int_1^t -\partial_s \log \pi (x_s) + \sum_{y_s} R_s (x_s, y_s) \frac{\pi (y_s)}{\pi (x_s)} \dif s$, if the forward kernel $\gamma$ is chosen such that the rate matrices satisfy detailed balance $\hat{R}_t (x_t, y_t) \pi(x_t) = {R}_t (y_t, x_t) \pi(y_t)$.
\end{restatable}
\cref{prop:smc_cont} coincides with the importance weight used in \cite{holderrieth2025leaps}, which focus on sampling scenarios where the intermediate target $\pi$ is tractable up to an unnormalized constant. In contrast, our setting concerns test-time control of pretrained diffusion models, where $\pi$ is intractable. In \cref{sec:appendix-conti-smc-ctmc}, we further connect our result to \cite{lee2025debiasing}, who also study reward tilting, but we provide a derivation from the perspective of discrete-time diffusion.

\subsection{Choices of Proposal: the Way to Reduce Variance}
While the proposal $q(x_{t-1} | x_t)$ offers substantial flexibility, the statistical efficiency of SMC is highly sensitive to its choice: suboptimal proposals induce high-variance importance weights, which in turn precipitate particle degeneracy and hinder adequate exploration of the state space \citep{del2006sequential}. Conversely, an appropriately constructed proposal substantially mitigates weight variance, thereby enhancing the effective sample size and ensuring stability of the inference procedure.
The following proposition characterises the minimum variance choice of proposal:
\begin{restatable}[Locally Optimal Proposal]{proposition}{restaproptwo} \label{prop:locally-opt-prop}
    Given the incremental importance weight  as in \cref{eq:iw_recursive} $w_{t-1}(x_{t-1}, x_t) = \frac{\pi(x_{t-1}) \gamma(x_t | x_{t-1})}{\pi(x_t) q(x_{t-1} | x_t)}$, the proposal distribution that minimises the variance of $w_{t-1}$, often referred to as the \emph{locally optimal proposal}, is $q(x_{t-1} | x_t) \propto \pi(x_{t-1}) \gamma(x_t | x_{t-1})$.
\end{restatable}
Building on \cref{prop:locally-opt-prop}, one readily verifies that the optimal proposal distribution for the case of product target is tractable (see the remark in \cref{sec:appendix-optimal-proposal} for discussion), under the choices of forward kernels $\gamma \propto p_{1}^\alpha (x_t | x_{t-1}) p_{2}^\beta (x_t | x_{t-1})$.
In contrast, for the reward-tilting, the locally optimal proposal takes the form $q \propto \exp(r(x_{t-1})) p_\theta (x_{t-1} | x_t)$, which is generally computationally infeasible due to the inaccessibility of the corresponding normalising constant.
Consequently, practical implementations must resort to approximations that balance variance reduction with computational feasibility.
In what follows, we introduce two approximation strategies tailored to the reward-tilting setting: i) a gradient-based method to achieve first-order approximation, and ii) a neural proposal trained to minimise the log-variance of the importance weight.

\subsubsection{Approximated Optimal Proposal via First-Order Approximation}
In reward tilting, evaluating the locally optimal proposal requires computing the normalising constant $Z = \sum_{x_{t-1}} \exp(r(x_{t-1}))p_\theta(x_{t-1} | x_t)$. This computation entails $\mathcal{O}(|\mathcal{X}|)$ forward pass through the reward model at each denoising step, which significantly slows down the generation speed, rendering the method impractical for large discrete state spaces.
To mitigate this issue, we adopt the approach of \cite{grathwohl2021oops,zhang2022langevin}, treating $r(x_t)$ as a function defined over continuous real-valued inputs, while evaluating it on the discretised domain of interest. This allows us to apply a first-order Taylor expansion to approximate the reward: $r(x_{t-1}) \approx r(x_t) + (x_{t-1} - x_t)^T \nabla_x r(x_t)$, which in turn yields a first-order approximation to the locally optimal proposal:
\begin{align} \label{eq:approx-opt-prop}
    q(x_{t-1} | x_t) \propto p_\theta (x_{t-1} | x_t) \exp( x_{t-1}^T \nabla_x r(x_t)).
\end{align}
This approximation improves computational efficiency by requiring the reward function $r$ to be evaluated and differentiated only once at $x_t$, instead of repeatedly across all states.
Nevertheless, it assumes differentiable rewards and remains costly when the reward model is large.
Motivated by \cite{richter2020vargrad,richter2023improved} and the amortisation technique in variational inference \citep{dayan1995helmholtz,kingma2013auto}, we propose learning an amortised network that approximates the optimal proposal, resulting in a transition kernel that directly transports between successive intermediate targets as in \cite{matthews2022continual}. This reduces computation to a single network evaluation, thereby significantly enhancing the efficiency of SMC.

\begin{figure*}[!t]
\vspace{-4mm}
\centering
    \begin{minipage}[t]{0.48\linewidth}
    \centering
    \begin{algorithm}[H]
    \caption{Training Optimal Proposal} \small
    \label{alg:train-proposal} 
    \setstretch{1.33}
    \begin{algorithmic}[1] 
        \STATE Rollout trajectory $\{x_t\}_t$ with $q_{\mathrm{ref}}(x_{t-1} | x_t)$
        \STATE Compute gradient with \cref{eq:log-var-upper-bound}\\ $g_\phi, g_\psi \leftarrow \frac{1}{T} \nabla_{\phi, \psi} \sum_t \mathcal{L}_{\phi, \psi}(x_{t-1}, x_t)$
        \STATE Update $\phi, \psi$ using $g_\phi, g_\psi$ \\ \vspace{-.4mm} $\phi \leftarrow \phi - \eta g_\phi, \quad \psi \leftarrow \psi - \eta g_\psi$ \vspace{-.4mm}
    \end{algorithmic}
    \end{algorithm}
    \end{minipage}
    \quad
    \begin{minipage}[t]{0.48\linewidth}
    \centering
    \begin{algorithm}[H]
    \setstretch{1.25}
    \caption{SMC Sampling Procedure} \small
    \label{alg:smc-sampling} 
    \begin{algorithmic}[1] 
        \STATE Propose particles $x_{t-1}^{(i)} \sim q(x_{t-1} | x_t)$
        \STATE Compute importance weight with \cref{eq:iw_recursive} \\ $w^{(i)}_{t-1} \!=\! \frac{\pi(x_{t-1})}{\pi(x_t)} \frac{\gamma(x_t | x_{t-1})}{q(x_{t-1} | x_t)}, \Tilde{w}_{t-1}^{(i)} \!=\! \frac{w^{(i)}_{t-1}}{\sum_{i} w^{(i)}_{t-1}}$
        \STATE Resample $\!x_{t-1}^{(i)} \!\!\sim\!\! \mathrm{Multinormial}(x_{t-1}^{(i)}; \Tilde{w}_{t\-1}^{(i)})$
    \end{algorithmic}
    \end{algorithm}
    \end{minipage}
\label{fig:algorithms}
\end{figure*}

\subsubsection{Amortised Optimal Proposal via Log-Variance Minimisation}
To train a network $q_\phi$ to approximate the locally optimal proposal, a natural approach is to minimise the log-variance of the importance weight:
\begin{align}
    \min_\phi \mathbb{V}_{q_{\mathrm{ref}}(x_{0:T})} \left[ \sum_t \log  \frac{\exp(r(x_{t-1}))}{\exp(r(x_t))}\frac{p_\theta (x_{t-1} | x_t)}{q_\phi (x_{t-1} | x_t)} 
    \right] \triangleq \mathcal{L}_{\text{log-var}}(\phi)
\end{align}
where $q_\mathrm{ref}$ is an arbitrary reference distribution that has the same support as $p_\theta$ and $q_\phi$.
The following corollary establishes the validity of the proposed log-variance objective.
\begin{restatable}{corollary}{restatecorollary} \label{cor:unique-optimal}
    The locally optimal proposal $q^* \propto \pi(x_{t-1}) p_\theta (x_{t-1} | x_t)$ that achieves the minimum variance of the important weight $\mathbb{V}_{q} \left[\frac{\pi(x_{t-1}) \gamma(x_t | x_{t-1})}{\pi(x_t) q(x_{t-1} | x_t)}\right]$ is unique.
\end{restatable}
Although conceptually simple, naive Monte Carlo estimation of $\mathcal{L}_{\text{log-var}}$ suffers from high variance and computational cost. To alleviate these issues, we introduce an auxiliary network $F_\psi: \mathbb{R} \rightarrow \mathbb{R}$, parameterised by $\psi$, that estimates the mean of the log-weight. This yields the refined objective:
\begin{align} \label{eq:log-var-upper-bound}
    \mathcal{L}(\phi, \psi) = \mathbb{E}_{t, q_{\mathrm{ref}}(x_{t-1}, x_t)} \left| \log \frac{\exp(r(x_{t-1}))}{\exp(r(x_t))}\frac{p_\theta (x_{t-1} | x_t)}{q_\phi (x_{t-1} | x_t)} - F_\psi (t) \right|^2,
\end{align}
which provably upper bounds the log-variance loss. To be specific, the following proposition holds:
\begin{restatable}{proposition}{restateproplogvarobj}
    For any reference distribution $q_{\mathrm{ref}}$, we have $\mathcal{L}_{\text{log-var}}(\phi) \leq T^2 \mathcal{L}(\phi, \psi)$. Moreover, the minimiser of $\mathcal{L}$ is unique and attains its optimum when $q_\phi \propto \exp(r(x_{t-1})) p_\theta (x_{t-1} | x_t)$.
\end{restatable}
We outline the training and sampling procedures in \cref{alg:train-proposal,alg:smc-sampling}. For clarity, we designate \oursbase, \oursgrad, and \oursamot to denote, respectively, the variants employing the pretrained diffusion proposal, the first-order approximated proposal, and the learned amortised proposal.
We further denote the first-order approximated and amortised proposals by \propgrad and \propamot, which coincide with their corresponding SMC methods when restricted to a single particle.

\subsection{Sequential Monte Carlo Recipe: Practical Implementation} \label{sec:smc-recipe}
Building on the theoretical characterisation of optimal proposals, we next present a practical SMC recipe. An essential ingredient of the proposed framework is the introduction of a twisted intermediate target for reward-tilting: $\pi(x_t) \propto p_\theta (x_t) \exp \left( \frac{\lambda_t}{\alpha} r(x_t) \right)$, where $\alpha > 0$ is a KL-regularisation coefficient. This construction is motivated by the following identity
\begin{align}
    \pi = \argmax_{\pi} \E_{\pi} [r(x_t)] - \alpha \mathbb{KL}(\pi \Vert p_\theta) \propto p_\theta (x_t) \exp \left( \frac{r(x_t)}{\alpha} \right).
\end{align}
Here,  $\lambda_t \in [0, 1]$ acts a temperature parameter that smoothly interpolates between the prior $p_\theta (x_t)$ ($\lambda_t = 0$) and the fully reward-augmented target ($\lambda_t = 1$). By gradually increasing $\lambda_t$ over denoising steps, the influence of the reward is tempered, thereby improving stability during sampling.
In scenarios where the reward is only defined on clean data, following \cite{wu2023practical,kim2025alignment}, we approximate the optimal intermediate target as
\begin{align} \label{eq:reward-est}
    \pi(x_t) \propto p_\theta (x_t)  \exp \left( \frac{\lambda_t}{\alpha} \hat{r}(x_t) \right), \quad \hat{r}(x_t) = \frac{1}{M} \sum_{m=1}^M r(x_0^{(m)}), \quad x_0^{(m)} \sim p_\theta (x_0 | x_t).
\end{align}
However, categorical sampling from $p_\theta$ renders $\hat{r}(x_t)$ non-differentiable w.r.t. $x_t$. To resolve it, we employ the reparameterisation trick with Gumbel-Softmax \citep{jang2016categorical} to enable differentiability (see \cref{sec:appendix-gumbel-softmax} for details), thereby making the approximated proposal applicable as in \cref{eq:approx-opt-prop}.
For completeness, \cref{sec:appendix-ratio-low-confidence} provides a further discussion of the computation of importance weights in \cref{eq:important-weight-tilting} under low-confidence sampling, where the ratio $\frac{p_\theta (x_{t-1} | x_t)}{q(x_{t-1} | x_t)}$ is not explicitly tractable. This extension ensures that the proposed SMC algorithm remains suitable for recent state-of-the-art discrete diffusion models for language modelling \citep{nie2025large} and text-to-image generation \citep{bai2024meissonic}, where low-confidence sampling \citep{chang2022maskgit} is commonly used.

%% file: sections_iclr_cr/03-experiments.tex
\section{Experiments}
To support our theoretical discussion, we first showcase the effectiveness of the proposed methods through a synthetic experiment. We then evaluate it across a wide range of applications, including language modelling, biological design, and text-to-image generation. 
Detailed experimental settings and additional results are provided in \cref{sec:appendix_exp}.

\subsection{Synthetic Experiemnts}
We begin with the empirical evaluation with two synthetic experiments: binary MNIST and a two-dimensional discretised mixture of Gaussians (MoG), each dimension comprising 64 categorical states.
A discrete diffusion model is first pretrained on the clean data. We evaluate the proposed SMC-based reward-tilting variants in comparison with two non-SMC baselines: \propgrad, corresponding to the first-order approximation in \cref{eq:approx-opt-prop}, and \propamot, which utilises an amortised proposal trained according to the objective in \cref{eq:log-var-upper-bound}.
Performance is assessed using the earth mover’s distance (EMD), alongside the evaluation of the reward on the generated samples.

\begin{figure}[!t]
    \centering
    \includegraphics[width=\linewidth]{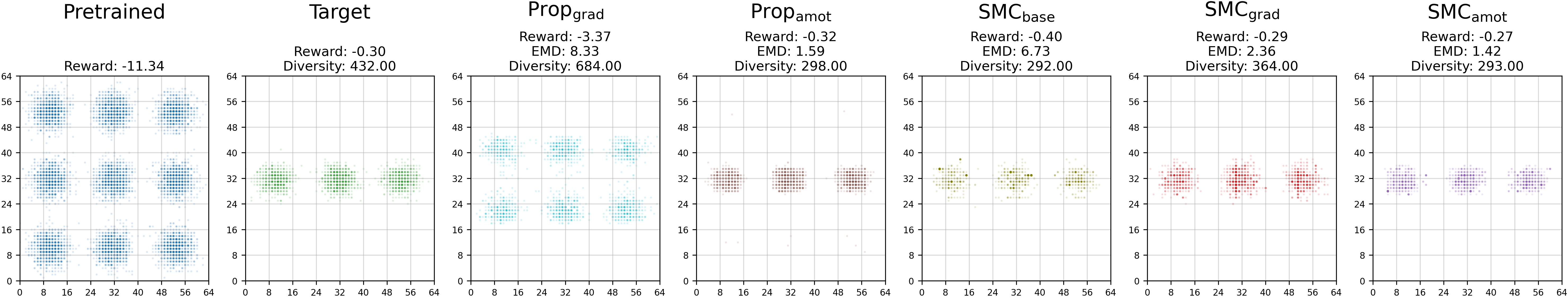}
    \vspace{-6mm}
    \caption{Comparisons on reward-tilted discreteised MoGs. We consider the reward function as $r(X, Y) \!=\! -\hat X^2/100-\hat Y^2$, where $\hat X \!=\! 12(X/63 - 1/2)$ and $\hat{Y} \!=\! 12(Y/63-1/2)$.}
    \label{fig:gmm-comparison}
    \vspace{-2mm}
\end{figure}

\begin{figure}[!t]
    \centering
    \includegraphics[width=\linewidth]{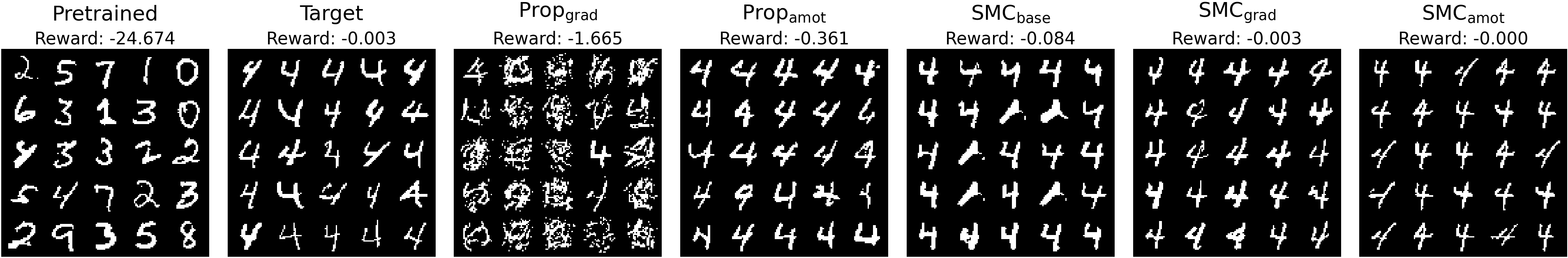}
    \vspace{-6mm}
    \caption{Comparisons on reward-tiled binary MNIST. We train a classier $p_{\text{clf}}(y|x)$ on the clean data, and the reward is given by $r(x) = \log p_{\text{clf}}(y_{\text{target}}|x)$, where $y_{\text{target}}$ denotes the target digit.}
    \label{fig:smc_mnist}
\end{figure}

The results, shown in \cref{fig:gmm-comparison,fig:smc_mnist}, indicate that, compared to the non-SMC baselines, the SMC-based methods achieve superior performance, demonstrating the effectiveness of the proposed approach. Specifically, \oursamot attains the highest rewards and lowest EMD, though at the cost of reduced sample diversity, likely due to reward over-optimisation. In contrast, \oursgrad maintains comparable sample quality while preserving high diversity, highlighting the effectiveness of the proposed approximated optimal proposal. Furthermore, \propamot significantly outperforms \propgrad, underscoring the benefit of the log-variance minimisation objective.
We further demonstrate the reward curves over training in \cref{fig:gmm_rewards_training,fig:mnist_rewards_training}, which shows that the proposed method can achieve stable reward convergence, confirming the efficiency and robustness in learning optimal proposal.


\subsection{Language Modelling}
We further evaluate our approach on language modelling, focusing on toxic text generation \citep{singhal2025general}, an undesirable behaviour of language models, where the pretrained model MDLM \citep{sahoo2024simple} produces only $0.8\%$ of samples flagged as toxic.
To assess sample quality, we use four metrics: i) Toxic, based on the same reward model applied during inference \citep{logacheva2022paradetox}; ii) Toxic (Holdout), measured by a holdout toxicity classifier trained on a multilingual mixture of datasets \citep{dementieva2024overview}; iii) generative perplexity with GPT2-XL \citep{radford2019language}; and iv) distinct uni/bi/trigrams (Dist-1/2/3). The first two metrics evaluate alignment with the reward, while the latter two measure semantic quality and diversity.

Following \cite{han2022ssd},  we generate sequences of length 100 with 100 denoising steps condition on the given starting prompts, and report results averaged over 300 independent runs corresponding to 15 prompts with 20 generations per prompt.
The results are summarised in \cref{tab:toxicity-results}, with an extended version provided in \cref{tab:toxicity-results-expanded}.
We observe that SMC with proposals closer to the optimal achieves better performance on the toxicity metrics, reflecting stronger alignment with the reward model. 
Among the non-SMC baselines, $\text{Prop}_{\text{amot}}$ yields the best performance, highlighting the effectiveness of the log-variance minimisation objective.
To further assess its effect, we plot the training dynamics in \cref{fig:toxicity_rewards_training}, which shows the reward steadily improving as training progresses.
Notably, although the learned proposal sacrifices a small degree of performance on perplexity and diversity, we demonstrate in \cref{fig:toxic-text-gen-samples-comp} that it consistently generates coherent and semantically meaningful sequences, indicating that alignment improvements need not come at the expense of sample quality.

\subsection{Biology Design}
In this experiment, we evaluate our method on DNA sequence design. Specifically, we adopt the pretrained model and the reward model from \cite{wang2024fine}, which are trained on $\sim$700k DNA sequences.
To evaluate the performance, we consider five metrics: i) predicted activity (\textit{Pred-Activity}); ii) chromatin accessibility classification accuracy (\textit{ATAC-Acc}); iii) 3-mer Pearson correlation with dataset sequences (\textit{3-mer Corr}); iv) JASPAR motif frequency correlation (\textit{JASPAR Corr}); and v) approximate log-likelihood under the pretrained model (\textit{App-Log-Lik}).
For further details on these evaluation metrics, we refer the reader to \cite{wang2024fine}.

\begin{figure}[!t]
    \centering
    \begin{minipage}[t]{\linewidth}
        \centering
        \includegraphics[width=.99\linewidth]{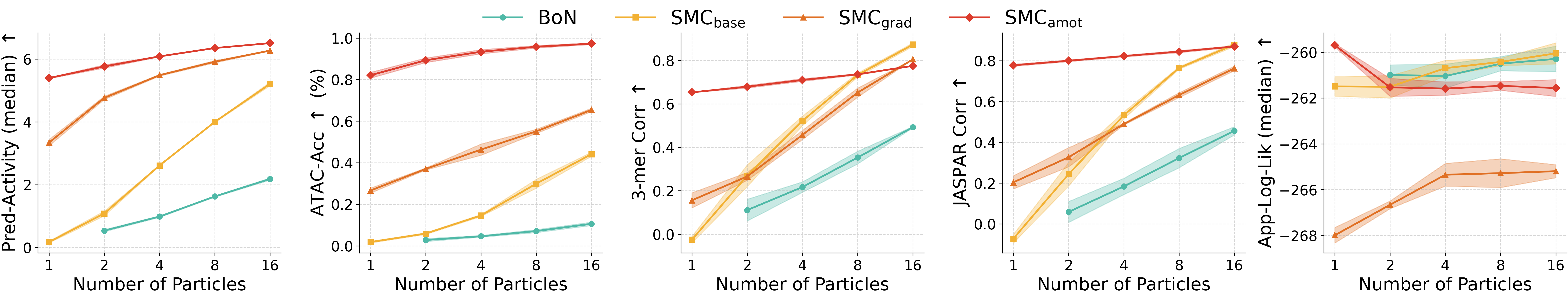}
    \end{minipage}
    \vspace{-4mm}
    \caption{Results of DNA sequence design. Both the pretrained discrete diffusion model and the reward models are adopted from \cite{wang2024fine}.}
    \label{fig:smc_dna}
    \vspace{-4mm}
\end{figure}

As shown in \cref{fig:smc_dna}, the performance improves consistently with an increasing number of particles, suggesting that SMC benefits from a larger particle set by providing a more accurate approximation of the target distribution.
Compared to \oursbase, \oursamot achieves higher \textit{Pred-Activity} and \textit{ATAC-Acc}, while performing slightly worse on the other three metrics. This can be attributed to the more mode-seeking behaviour of their proposals, which emphasises high-probability regions at the expense of overall diversity.
Nonetheless, we observe that \oursamot with larger particle sets attains more favourable overall performance, indicating that a learnable amortised proposal can effectively leverage the flexibility of SMC to balance quality and diversity.
In \cref{sec:appendix-comp-baselines}, we present additional comparisons with baseline methods, further demonstrating the effectiveness of our approach.

\begin{table}[!t]
\centering
\small
\caption{The results of toxic text generation. We use a widely adopted toxicity classifier as the reward \citep{logacheva2022paradetox}, while the pretrained language model is MDLM \citep{sahoo2024simple}.}
\label{tab:toxicity-results}
\vspace{-2mm}
\resizebox{1.0\linewidth}{!}{
\begin{tabular}{c c c c c c}
\toprule
{\# Particles} & {Method} & {Toxic \(\uparrow\)} & {Toxic (Holdout) \(\uparrow\)} & {PPL (GPT2-XL) \(\downarrow\)} & {Dist-1/2/3 \(\uparrow\)} \\
\midrule
\multirow{3}{*}{N = 1} & Pretrained
&  0.8\% &  5.2\% & {121.1} & 56/92/96 \\
& \propgrad
& 58.0\% & 58.3\% & 216.7 & 58/93/96 \\
& \propamot
& 63.7\% & 75.7\% & 131.9 & 53/89/94 \\
\midrule
\multirow{4}{*}{N = 8} & BoN
&  6.3\% & 16.7\% & 127.4 & 56/91/96 \\
&\oursbase
& 26.7\% & 40.0\% & 132.3 & 57/92/96 \\
&\oursgrad
& {95.0\%} & {86.3\%} & 132.1 & 57/92/96 \\
&\oursamot
& 100.0\% & 99.7\% & 147.6 & 44/81/91 \\
\bottomrule 
\end{tabular}
}
\end{table}

\subsection{Image Generation}
In this section, we evaluate our method on image generation. We begin by demonstrating that SMC yields improvements over classifier-free guidance, which can be viewed as a special case of the product target in \cref{eq:important-weight-anneal-comp}. Subsequently, we present large-scale experiments to illustrate the applicability of the proposed methods to text-to-image generation at scale.

\textbf{Improving CFG with MaskGit \citep{chang2022maskgit}.}
Given a pretrained diffusion model $p_\theta$, classifier-free guidance (CFG) generates samples according to $p_\theta (x_{t-1} | x_t, c)^\alpha p_\theta (x_{t-1} | x_t)^{1-\alpha}$, where $\alpha$ is the CFG coefficient. CFG has been shown to enhance sample quality substantially \citep{ho2022classifier}. By incorporating the importance weight defined in \cref{eq:important-weight-anneal-comp}, we can further improve CFG within the proposed SMC framework.

\begin{wraptable}{r}{0.5\linewidth}
\caption{Comparisons of different numbers of particles with CFG=1.25 on ImageNet256.}
\label{tab:fid_is_diff_particles_cfg1.25}
\vspace{-2mm}
\centering
\resizebox{1.0\linewidth}{!}{
\begin{tabular}{lrrrrrr}
    \toprule
    & \multicolumn{3}{c}{FiD $\downarrow$} & \multicolumn{3}{c}{IS $\uparrow$} \\
    \cmidrule(lr){2-4} \cmidrule(lr){5-7} 
    \# steps & 8 & 16 & 32 & 8 & 16 & 32 \\
    \midrule
    N = 1 & 24.64 & 14.94 & 12.02 & 62.8 & 90.7 & 107.5 \\
    N = 2 & 21.08 & 12.55 & 10.26 & 74.0 & 106.6 & 126.2 \\
    N = 4 & 18.08 & 10.92 & 9.29 & 87.4 & 123.5 & 146.5 \\
    N = 8 & 16.26 & 9.93 & 8.98 & 96.4 & 139.4 & 159.8 \\
    N = 16 & 14.56 & 9.59 & 8.76 & 107.4 & 149.4 & 170.7 \\
\bottomrule
\end{tabular}
}
\vspace{-4mm}
\end{wraptable}
Specifically, we perform experiments with MaskGit \citep{chang2022maskgit} trained on ImageNet256 \citep{deng2009imagenet}. 
To ensure tractable importance weight computation, we adopt the ReMDM sampling scheme \citep{wang2025remasking} instead of the low-confidence sampling strategy from \citep{chang2022maskgit} (see \cref{tab:fid_is_diff_sampling_method_cfg1.5,tab:fid_is_diff_sampling_method_cfg1.75,tab:fid_is_diff_sampling_method_cfg1.25} for a comparison).
The result is presented in \cref{tab:fid_is_diff_particles_cfg1.25}. 
It shows that with fewer denoising steps, increasing the number of particles leads to a substantial improvement in sample quality, as measured by FID and Inception Score (IS) with $50,000$ generated images, thereby demonstrating the effectiveness of the proposed SMC method.
However, as the number of denoising steps increases, the benefit of using more particles diminishes. This can be attributed to the role of SMC in correcting sampling inaccuracies: with sufficient denoising steps, the sampling process itself becomes accurate enough, leaving limited room for further improvement through additional particles.
In addition, we report results with different CFG coefficients in \cref{tab:fid_is_diff_particles_cfg1.5,tab:fid_is_diff_particles_cfg1.75}. Interestingly, for larger CFG coefficients, increasing the particle count tends to decrease FID while increasing IS. This behaviour is expected, since stronger CFG reduces sample diversity. With more accurate sampling under SMC, this reduction in diversity becomes more apparent, leading to lower FID but higher IS.

\textbf{Improving Text-to-Image Generation with Meissonic \citep{bai2024meissonic}.}
We evaluate the scalability of the proposed methods on text-to-image generation using Meissonic \citep{bai2024meissonic} as the base model.
Our experiments consider three text–image alignment rewards: Human Preference Score (HPSv2) \citep{wu2023human}, Aesthetic Score \citep{laion2024aesthetics}, and ImageReward \citep{xu2023imagereward}.
For the prompt distributions, we use photo and painting prompts from the Human Preference Dataset (HPDv2) \citep{wu2023human} for HPSv2, the DrawBench prompt set for ImageReward, and a curated set of 45 simple animal prompts for Aesthetic Score, follwoing \cite{black2023training}.

\begin{figure}[!t]
    \centering
    \begin{minipage}{0.9\linewidth}
        \centering
        \includegraphics[width=.99\linewidth]{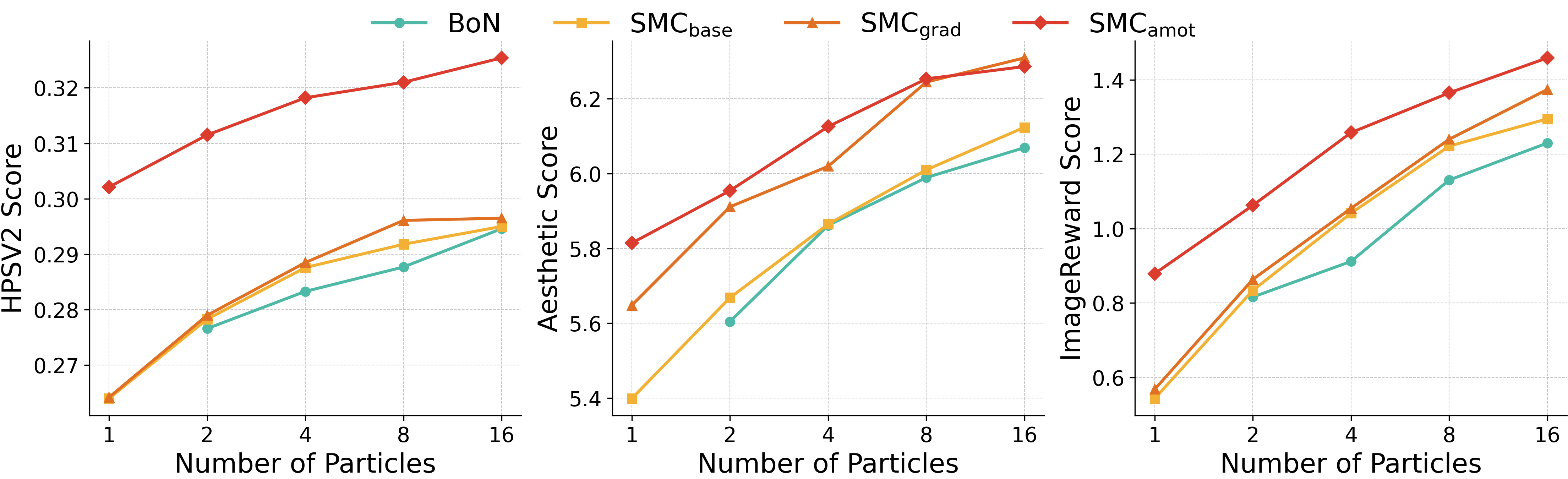}
    \end{minipage}
    \vspace{-2mm}
    \caption{The results of text-to-image generation across different reward models.}
    \label{fig:t2i-metrics}
\end{figure}

\begin{figure}[!t]
    \begin{minipage}{0.99\linewidth}
        \centering
        {\scriptsize ---\ A helmet-wearing monkey skating. --->}
        \includegraphics[width=.99\linewidth]{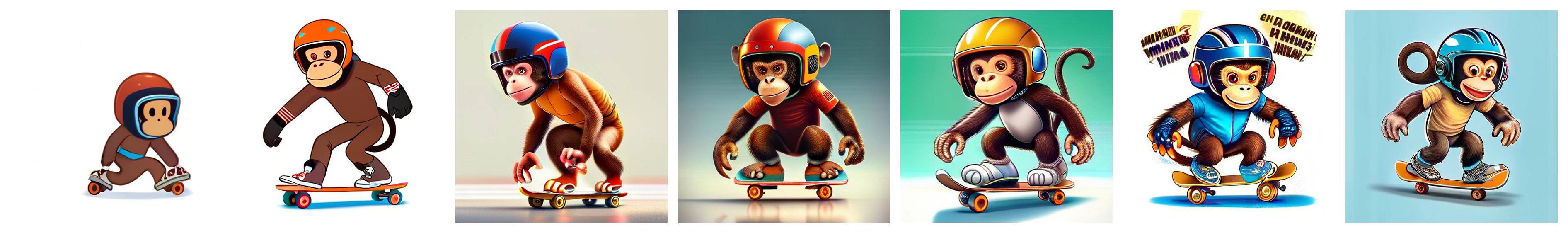}
    \end{minipage}
    \begin{minipage}{0.99\linewidth}
        \centering
        {\scriptsize ---\ The image features a castle surrounded by a dreamy garden with roses and a cloudy sky in the background. --->}
        \includegraphics[width=.99\linewidth]{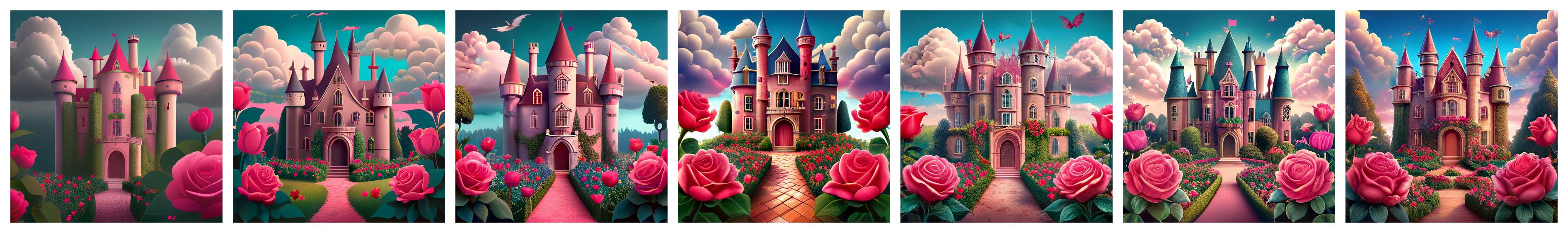}
    \end{minipage}
    \vspace{-2mm}
    \caption{Visualisation of alignment dynamics over the training progress, with images generated every 100 training steps. The generated images become more faithful to the text prompt.}
    \label{fig:t2i-hpsv2-training}
    \vspace{-4mm}
\end{figure}

The results are shown in \cref{fig:t2i-metrics}.
We observe that performance consistently improves with an increasing number of particles, and \oursamot outperforms all other methods, which highlights the benefit of the proposed SMC framework. 
In \cref{fig:t2i-hpsv2-training}, we visualise the alignment dynamics for the HPSv2 task, showing that the generated images progressively align more faithfully with the prompts, thereby validating the effectiveness of the proposed log-variance minimisation objective.
Furthermore, \cref{fig:hpsv2_rewards_training,fig:aesthetic_rewards_training,fig:imagereward_rewards_training} present the convergence of the reward during training, and further qualitative examples in \cref{sec:appendix-illustrate-samples} collectively reinforce the validity of our approach.

%% file: sections_iclr_cr/04-related_work.tex
\section{Related Work}

\textbf{Discrete Diffusion Models.}
Discrete diffusion models (DDMs) were originally introduced in \cite{austin2021structured,sun2022score,campbell2022continuous}, grounded in the framework of continuous-time Markov chains \citep{norris1998markov}. More recently, masked diffusion models (MDMs) \citep{lou2023discrete,shi2024simplified,sahoo2024simple}, a special case of DDMs, have shown strong performance in language modelling \citep{zhang2025target,nie2025large}. In addition, MDMs have achieved promising results in math reasoning \citep{zhao2025d1}, image synthesis \citep{bai2024meissonic}, code planning \citep{gat2024discrete,gong2025diffucoder}, and biological sequence generation \citep{campbell2024generative}, yielding performance comparable to continuous diffusion \citep{rombach2022high} and autoregressive models \citep{radford2019language}.
In contrast to these approaches, which primarily study large-scale pretraining, our work focuses on test-time inference and post-training alignment \citep{uehara2025reward}, where access to training data is not available.

\textbf{Test-time Alignment of Discrete Diffusion Models.}
Existing alignment methods mainly fall into two categories: classifier guidance \citep{dhariwal2021diffusion} and RL-based fine-tuning \citep{black2023training}.
Although the score is ill-defined in discrete distributions, several works \citep{vignac2022digress,nisonoff2024unlocking,schiff2024simple} employ a first-order approximation to the target distribution, which resembles the insight underlying our approximated proposal \propgrad in \cref{eq:approx-opt-prop}.
\revision{
Alternatively, \cite{rout2025test} perform guidance on the embedding space, mitigating the issue of ill-defined gradients. \cite{chen2025rfg} introduce reward-free guidance, analogous to classifier-free guidance \citep{ho2022classifier} but designed for masked diffusion models.
Moreover, \cite{tang2025tr2} propose tree search guided finetuning, which is related to the searching-based scaling methods on continuous diffusion \citep{ma2025scaling,zhang2025inference,jain2025diffusion,ramesh2025test}.
}
Beyond guidance approaches, sampling-based techniques have also demonstrated promising performance, including value-based sampling \citep{li2024derivative}, importance sampling \citep{guo2024plug}, and iterative refinement strategies \citep{uehara2025rewarditer}.
While training-free and relatively efficient to deploy, these methods often face challenges in scalability and robustness.
More recently, RL-based fine-tuning methods \citep{zekri2025fine,zhao2025d1,gong2025diffucoder} have gained significant traction, fueled by the remarkable success of Group Relative Policy Optimisation (GRPO) \citep{shao2024deepseekmath} in large language models \citep{guo2025deepseek}. In parallel, steering-based \citep{rector2024steering} approaches leveraging GFlowNets \citep{bengio2023gflownet} and direct backpropagation methods \citep{wang2024fine} have also demonstrated strong potential for test-time alignment.
Distinct from these directions, our amortised proposal \propamot introduces an alternative perspective for fine-tuning pretrained discrete diffusion models: it minimises the log-variance of importance weights, a criterion that has been rarely investigated in previous work.

\textbf{Sequential Monte Carlo for Generative Modelling.}
SMC has emerged as a versatile framework for probabilistic modelling, providing effective tools for sampling and inference across a wide range of applications, including particle filtering \citep{johansen2009tutorial}, Bayesian experimental design \citep{ryan2016review}, and probabilistic planning \citep{piche2018probabilistic}.
Most recently, SMC has been combined with diffusion models \citep{chen2024sequential,he2025rne,skreta2025feynman,wu2025reverse,ou2026diffusion}, transforming it into a powerful neural sampler capable of drawing from complex Boltzmann distributions.
These developments have also extended SMC’s reach to discrete domains, as demonstrated by \citet{holderrieth2025leaps,lee2025debiasing}.
Beyond classical sampling tasks, SMC has further expanded to the improvement of generative models at test time.
A seminal step in this direction was taken by \citet{zhao2024probabilistic}, who introduced SMC as a principled probabilistic inference framework for addressing capability and safety challenges in large language models (LLMs). Subsequent works \citep{feng2024step,puri2025rollout} successfully applied this idea to enhance mathematical reasoning in LLMs, while others explored its use in reward-guided adaptation of pretrained diffusion models \citep{trippe2022diffusion,wu2023practical,cardoso2023monte,dou2024diffusion,kim2025alignment,yoon2025psi,chen2025solving,ren2025driftlite}.
Our work is most closely related to \citet{singhal2025general,dang2025inference,hasan2025discrete}, who employ SMC for test-time alignment of discrete diffusion models. However, their approaches treat pretrained diffusion models as fixed proposal distributions. By contrast, we take a closer look at the role of proposal choice, systematically investigating its impact and providing empirical evidence for a key insight: proposals that better approximate the optimal, which minimises the variance of importance weights, consistently lead to better performance.

%% file: sections_iclr_cr/05-conclusion.tex
\section{Conclusion}
In this paper, we introduced a Sequential Monte Carlo (SMC) framework tailored for discrete diffusion models. By exploiting tractable importance weights, we established SMC as a powerful and principled recipe for test-time scaling. A central insight of our work is that the proposal distribution is crucial for unlocking the full potential of SMC. Building on this observation, we developed two approximately optimal proposals: a first-order approximation and a learnable amortised proposal trained to approximate the optimal proposal by minimising the log-variance of importance weights.
Extensive experiments across diverse domains demonstrated the effectiveness and scalability of our approaches.
We hope this work inspires future studies on more efficient test-time scaling and post-training alignment strategies for discrete diffusion models.

\section{Acknowledgements}
ZO is supported by the Lee Family Scholarship.
We would like to thank the anonymous reviewers for their constructive and valuable suggestions.
ZO also thanks Ruixiang Zhang for insightful discussions and for his support with the experimental setup.

%% file: sections_iclr_cr/06-appendix.tex
\newpage 
\appendix

\begin{center}
\LARGE
\textbf{Appendix for ``Inference-Time Scaling of Discrete Diffusion Models via Importance Weighting and Optimal Proposal Design''}
\end{center}

\etocdepthtag.toc{mtappendix}
\etocsettagdepth{mtchapter}{none}
\etocsettagdepth{mtappendix}{subsection}
{\small \tableofcontents}

\section{Abstract Proof and Derivations}

\subsection{A Brief Recap of SMC}

\revision{
In this section, we provide a brief overview of Sequential Monte Carlo (SMC).
For a target distribution $\pi(x_t)$, we consider the problem of estimating the expectation of a test function $\delta$, namely $\E_{\pi(x_t)}[\delta(x_t)]$. When $\delta (\cdot)$ is taken to be the Dirac delta function, estimating this expectation reduces to constructing an empirical approximation of the distribution $\pi(x_t)$.
}

\revision{
To estimate the expectation, importance sampling introduces a proposal distribution $q$, which is easy to sample from, and proposes an estimator as follows
\begin{align}
    \E_{\pi(x_t)}[\delta(x_t)] \!=\! \E_{q(x_{t:T})} \left[ \frac{\pi(x_{t:T})}{q(x_{t:T})} \delta(x_t) \right] \!\approx\! \sum_{i=1}^N w_t^{(i)} \delta(x_t^{(i)}), \ \text{where}\ w_t^{(i)} \!=\! \frac{\pi(x_{t:T}^{(i)})}{q(x_{t:T}^{(i)})}, x_{t:T}^{(i)} \sim q (x_{t:T}) \nonumber
\end{align}
The key ingredients of SMC are the target distribution $\pi(x_{t:T})$ and the proposal distribution $q(x_{t:T})$.
Here we consider the target distribution as a Markovian model associated with a sequence of forward transition kernels $\gamma$: $\pi(x_{t:T}) = \pi(x_t) \prod_{s = t}^{T-1} \gamma (x_{t+1} | x_t)$; and the proposal distribution as $q(x_{t:T}) = \pi(x_T) \prod_{s = t}^{T-1} q(x_t | x_{t+1})$. Substituting these into the importance weights gives
\begin{align}
    w_t
    &= \frac{\pi(x_t) \prod_{s = t}^{T-1} \gamma (x_{t+1} | x_t)}{\pi(x_T) \prod_{s = t}^{T-1} q(x_t | x_{t+1})} \nonumber \\
    &=  \frac{\pi(x_t) \gamma (x_{t+1} | x_t) }{\pi(x_{t+1}) q(x_t | x_{t+1}) } 
    \frac{\pi(x_{t+1}) \prod_{s = t+1}^{T-1} \gamma (x_{t+1} | x_t)}{\pi(x_T) \prod_{s = t+1}^{T-1} q(x_t | x_{t+1})} \nonumber \\
    &= \frac{\pi(x_t) \gamma (x_{t+1} | x_t) }{\pi(x_{t+1}) q(x_t | x_{t+1})}  w_{t+1}
\end{align}
This recursive structure allows importance weights to be computed incrementally. SMC augments this with a resampling step to mitigate weight degeneracy.
For $N$ particles, SMC proceeds as follows:
\begin{itemize}
    \item[1.] Initialise: $x_T^{(i)} \sim \pi (x_T), w_T^{(i)} = 1$.
    \item[2.] For $t = T, \dots, 1$:
        \begin{itemize}
            \item[(a)] Propagate: $x_{t-1}^{(i)} \sim q(x_{t-1} | x_t^{(i)})$.
            \item[(b)] Update weights: $\frac{\pi(x_t^{(i)}) \gamma (x_{t+1}^{(i)} | x_t^{(i)}) }{\pi(x_{t+1}) q(x_t^{(i)} | x_{t+1}^{(i)})}$.
            \item[(c)] Resample particles according to $\left\{\frac{w_{t-1}^{(i)}}{\sum_{j=1}^N w_{t-1}^{(j)}} \right\}_{i=1}^N$; then reset all weights to $w_{t-1}^{(i)} = 1$.
        \end{itemize}
\end{itemize}
The resulting set of particles $\{x_0^{(i)}, w_0^{(i)}\}_{i=1}^N$ forms an empirical approximation of the target $\pi(x_0)$.
}

\subsection{Proof of Locally Optimal Proposal} \label{sec:appendix-optimal-proposal}

\restaproptwo*
\begin{proof}
    We first present an intuitive argument to aid understanding, and subsequently provide the formal proof.

    \textit{Intuitive argument.} 
    The optimal proposal distribution is characterised as the one that minimises the variance of the importance weights. In the degenerate case of zero variance, the importance weight must be constant:  
    $\frac{\pi(x_{t-1}) \gamma(x_{t-1} | x_t)}{\pi(x_{t-1}) q(x_{t-1} | x_t)} = c$, for some constant $c > 0$. Rearranging yields
    \begin{align}
        q^*(x_{t-1} | x_t) = \frac{1}{c} \frac{\pi(x_{t-1})}{\pi(x_t)} \gamma(x_{t-1} | x_t) \propto \pi(x_{t-1}) \gamma(x_{t-1} | x_t),
    \end{align}
    where $c = \frac{1}{\pi(x_t)} \sum_{x_{t-1}} \pi(x_{t-1}) \gamma(x_{t-1} | x_t)$ is the normalising constant.
    
    \textit{Formal proof.} 
    The optimal proposal can be obtained by minimising the variance of the incremental importance weight $w(x_{t-1}, x_t) = \frac{\pi(x_{t-1}) \gamma(x_{t-1} | x_t)}{\pi(x_{t-1}) q(x_{t-1} | x_t)}$:
    \begin{align}
        q^* 
        &= \argmin_{q} \mathbb{E}_q \left[ w(x_{t-1}, x_t) - \mathbb{E}_q [w(x_{t-1}, x_t)] \right]^2 + a \left( \sum_{x_{t-1}} q(x_{t-1} | x_t) - 1\right) \nonumber \\
        &= \argmin_{q} \mathbb{E}_q \left[ w(x_{t-1}, x_t)^2 \right] - \mathbb{E}_q [w(x_{t-1}, x_t)]^2 + a(x_t) \left( \sum_{x_{t-1}} q(x_{t-1} | x_t) - 1 \right), \nonumber \\
        &= \argmin_{q} \sum_{x_{t-1}} \underbrace{w(x_{t-1}, x_t)^2 q(x_{t-1} | x_t) + a(x_t) q(x_{t-1} | x_t)}_{:= F(q)}  + c, \nonumber
    \end{align}
    where $c$ denotes a constant w.r.t. $q$ and we introduce a Lagrange multiplier $a (x_t) > 0$ for the constraint $\sum_{x_{t-1}} q(x_{t-1} | x_t) = 1$. Using the calculation of variation, where the functional $F$ should satisfy the Euler-Lagrange equation $\frac{\partial F}{\partial q} - \frac{\dif}{\dif x} \frac{\partial F}{\partial q^\prime} = 0$, we have
    \begin{align}
        \frac{\partial F}{\partial q} = - \left( \frac{\pi(x_{t-1}) \gamma(x_{t-1} | x_t)}{\pi(x_{t-1}) q(x_{t-1} | x_t)} \right)^2 + a(x_t) = 0 \ 
        \Rightarrow \ q^* (x_{t-1} | x_t) = \frac{\pi(x_{t-1}) \gamma(x_{t-1} | x_t)}{\pi(x_t) \sqrt{a(x_{t})}}. \nonumber
    \end{align}
    The term $\frac{1}{\pi(x_t)) \sqrt{a(x_t)}}$ is a normalisation constant that does not depend on the $x_{t-1}$. We can find its value by enforcing the constraint $\sum_{x_{t-1}} q^* (x_{t-1} | x_t) = 1$. This shows that the optimal proposal is $q^* (x_{t-1} | x_t) \propto \pi(x_{t-1}) \gamma(x_{t-1} | x_t)$.
\end{proof}
\textbf{Remark.} Given \cref{prop:locally-opt-prop}, we can derive the form of the locally optimal proposal under different settings. Specifically, using the importance weight defined in \cref{eq:important-weight-anneal-comp,eq:important-weight-tilting}, let the forward kernel $\gamma(x_t \mid x_{t-1})$ be specified as
\begin{align}
    \text{product:}\ \gamma \propto p_1^\alpha (x_t | x_{t-1}) p_2^\beta (x_t | x_{t-1}) \quad \text{reward-tilting:}\ \gamma \propto p(x_{t-1} | x_t) \nonumber
\end{align}
The corresponding importance weights are then
\begin{align}
    \text{product:}\ \frac{p_{\theta_1}^\alpha (x_{t-1} | x_t) p_{\theta_2}^\beta (x_{t-1} | x_t)}{Z(x_{t-1}) q(x_{t-1} | x_t)} \quad \text{reward-tilting:}\ \frac{\exp(r(x_{t-1}))}{\exp(r(x_t))}\frac{p_\theta (x_{t-1} | x_t)}{q(x_{t-1} | x_t)} \nonumber
\end{align}
where $Z(x_{t-1}) = \sum_{x_t} p_1^\alpha (x_t | x_{t-1})p_2^\beta(x_t | x_{t-1})$ is the normalising constant.
By \cref{prop:locally-opt-prop}, the corresponding locally optimal proposals are
\begin{align}
    \text{product:}\ q \propto \frac{ p_{\theta_1}^\alpha (x_{t-1} | x_t) p_{\theta_2}^\beta (x_{t-1} | x_t)}{Z(x_{t-1})} \quad  \text{reward-tilting:}\ q \propto \exp(r(x_{t-1})) p_\theta (x_{t-1} | x_t) \nonumber
\end{align}
The normalising constant $Z$ is tractable, since $p(x_t \mid x_{t-1})$ is a simple forward noising distribution (induced by \cref{eq:diffusion-forward}) that does not involve network evaluation. In contrast, for the reward-tilting, the dependence on the reward function $r$, which is defined via a neural network, renders the optimal proposal intractable in general. This necessitates the development of approximation techniques.

\subsection{Proof of Log-Variance Minimisation Objective}

\restatecorollary*
\begin{proof}
    Recall that the variance is given by $\mathbb{V}_q [w] = \mathbb{E}_q [w^2] - (\mathbb{E}_q [w])^2$. As shown in the proof of \cref{prop:locally-opt-prop}, the term $\mathbb{E}_q [w]$ is constant w.r.t. the choice of $q$. Therefore, minimising the variance is equivalent to minimise the expected square of the weights, $\mathbb{E}_q [w^2]$, which we will call $F(q)$:
    \begin{align}
        F(q) = \mathbb{E}_q [w^2] = \sum_{x_{t-1}} \frac{1}{q(x_{t-1} | x_t)} \left( \frac{\pi(x_{t-1}) \gamma (x_t | x_{t-1})}{\pi(x_t) } \right)^2.
    \end{align}
    To simplify the notation, let $q_i = q(x_{t-1} = i | x_t)$ and $C_i = \frac{\pi(x_{t-1}=i) \gamma (x_t | x_{t-1} = i)}{\pi(x_t)}$. The optimal proposal is $q^* = C_i / Z$, where $Z = \sum_j C_j$. We then can rewrite the objective function $F(q)$ as
    \begin{align}
        F(q) = \sum_i \frac{C_i^2}{q_i} = \sum_i \frac{\left( Z q_i^* \right)^2}{q_i} = Z^2 \sum_i \frac{(q_i^*)^2}{q_i}.
    \end{align}
    Evaluating the function at the optimum, $q^*$, we have
    \begin{align}
        F(q^*) =  \sum_i \frac{C_i^2}{q_i^*} = \sum_i \frac{\left( Z q_i^* \right)^2}{q_i^*} = Z^2 \sum_i q_i^* = Z^2.
    \end{align}
    To prove the uniqueness of the locally optimal proposal $q^*$, the key insight is to relate the expression for $F(q)$ to the Chi-squared divergence, which is defined as
    \begin{align}
        \chi^2 (q^* \Vert q) = \sum_i \frac{(q_i^* - q_i)^2}{q_i} = \left( \sum_i \frac{(q_i^*)^2}{q_i} \right) - 1.
    \end{align}
    Rearranging this, we see that $\sum_i \frac{(q_i^*)^2}{q_i} = \chi^2 (q^* \Vert q) + 1$. Now we can express $F(q)$ as
    \begin{align}
        F(q) =  Z^2 \sum_i \frac{(q_i^*)^2}{q_i} = Z^2 \left( \chi^2 (q^* \Vert q) + 1 \right) = Z^2 \chi^2 (q^* \Vert q) + Z^2.
    \end{align}
    Since $F(q^*) = Z^2$, we finally arrive at
    \begin{align}
        F(q) = F(q^*) + Z^2 \chi^2 (q^* \Vert q).
    \end{align}
    Since the $\chi^2$-divergence is non-negative and $\chi^2 (q^* \Vert q) = 0$ if and only if $q=q^*$, we see that the equality $F(q) = F(q^*)$ holds only when $\chi^2 (q^* \Vert q) = 0$, which requires that $q$ be identical to $q^*$. For any other distribution $q \neq q^*$, the divergence is strictly positive, meaning $F(q) > F(q^*)$. 
    Therefore, $q^*$ is the unique distribution that minimises the variance of the importance weights.
\end{proof}

\restateproplogvarobj*
\begin{proof}
    To prove the result, we first recall the basic identitie $\mathbb{E}_q [w] = \argmin_c \E_q [(w - c)^2]$ and $\V_q [w] = \E_q [(w - \E_q[w])^2]$. Applying these, we obtain
    \begin{align}
        \mathcal{L}_{\text{log-var}}(\phi)
        &= \mathbb{V}_{q_{\mathrm{ref}}} \left[ \sum_{t} \log \frac{\exp(r(x_{t-1}))}{\exp(r(x_t))}\frac{p_{\theta} (x_{t-1} | x_t)}{q_\phi(x_{t-1} | x_t)} \right] \nonumber \\
        &= \min_{F_t \in \mathbb{R}} \mathbb{E}_{q_{\mathrm{ref}}} \left[ \left| \sum_{t} \log \frac{\exp(r(x_{t-1}))}{\exp(r(x_t))}\frac{p_\theta (x_{t-1} | x_t)}{q_\phi (x_{t-1} | x_t)}  - F_t \right|^2 \right] \nonumber \\
        &= T^2 \min_{F_t \in \mathbb{R}} \mathbb{E}_{q_{\mathrm{ref}}} \left[ \left|  \sum_{t} \frac{1}{T} \log \frac{\exp(r(x_{t-1}))}{\exp(r(x_t))}\frac{p_\theta (x_{t-1} | x_t)}{q_\phi (x_{t-1} | x_t)}  - \frac{1}{T} F_t \right|^2 \right] \nonumber \\
        &\leq T^2 \min_{F_t \in \mathbb{R}} \mathbb{E}_{q_{\mathrm{ref}}} \left[ \sum_{t}  \frac{1}{T} \left|  \log \frac{\exp(r(x_{t-1}))}{\exp(r(x_t))}\frac{p_\theta (x_{t-1} | x_t)}{q_\phi (x_{t-1} | x_t)}  - F_t \right|^2 \right] \nonumber \\
        &= T^2 \min_{F_t \in \mathbb{R}} \mathbb{E}_{q_{\mathrm{ref}}, t} \left[ \left|  \log \frac{\exp(r(x_{t-1}))}{\exp(r(x_t))}\frac{p_\theta (x_{t-1} | x_t)}{q_\phi (x_{t-1} | x_t)}  - F_t \right|^2 \right]. \nonumber
    \end{align}
    This motivates defining the loss function as
    \begin{align}
        \mathcal{L}(\theta, \psi) = \E_{t, x_t \sim q_\mathrm{ref}} \left[ \left| \log \frac{\exp(r(x_{t-1}))}{\exp(r(x_t))}\frac{p_\theta (x_{t-1} | x_t)}{q_\phi (x_{t-1} | x_t)} - F_\psi(t) \right|^2 \right].
    \end{align}
    Therefore, we have the inequality 
    \begin{align}
        \mathcal{L}_{\text{log-var}}(\phi) \leq T^2 \mathcal{L}(\theta, \psi)
    \end{align}
    Consequently, if $(\theta^*, \psi^*) = \argmin_{\phi, \psi} \mathcal{L}(\theta, \psi)$, then
    \begin{align}
        \mathcal{L}(\theta^*, \psi^*) = 0 \quad \Rightarrow \quad \mathcal{L}_{\text{log-var}}(\phi^*) = 0 \quad \Rightarrow \quad \V[w] = 0.
    \end{align}
    Finally, by \cref{cor:unique-optimal}, the minimiser $\phi^*$ is unique, and the corresponding proposal takes the form $q_{\phi^*}(x_{t-1} | x_t) \propto \exp(r(x_{t-1})) p_\theta (x_{t-1} | x_t)$.
\end{proof}

\section{Extending Discrete-time SMC to Continuous-time SMC}

In this section, we extend our SMC algorithm from discrete time to continuous time. We begin by introducing the key preliminary: the continuous-time Markov chain (CTMC) \citep{norris1998markov}. We then establish connections to previous work \cite{holderrieth2025leaps,lee2025debiasing}, which develops continuous-time SMC methods for discrete diffusion models.

\subsection{Background of CTMC} 
A continuous-time Markov chain \citep{norris1998markov} at time $t$ is characterised by a time-dependent rate matrix $R_t: \mathcal{X} \times \mathcal{X} \rightarrow \mathbb{R}$, which captures the instantaneous rate of change of the transition probabilities. Specifically, the rate matrix $R_t$ is defined as
\begin{align}
    R_t (x, y) = \lim_{\Delta t \rightarrow 0} \frac{p_{t+\Delta t | t} (y | x) - \delta_{y = x}}{\Delta t}, \quad
    \delta_{y=x} = 
    \begin{cases}
        1, & y=x \\
        0, & y \neq x
    \end{cases}.
\end{align}
By definition, the rate matrix equivalently yields the transition probability
\begin{align}
    p_{t+\Delta t | t} (y | x) = \delta_{y = x} + R_t (x, y) \Delta t + \mathcal{O} (\Delta t).
\end{align}
To ensure $p_{t+\Delta t | t}$ be a valid distribution, the rate matrix $R_t$ must satisfy the following constraints:
\begin{align}
    R_t (x, y) \ge 0, \forall y \ne x, \quad R_t (x, x) = - \sum_{y \ne x} R_t (x, y).
\end{align}
The transition probability $p_{t|s}$, for $t > s$, satisfies the Kolmogrove equations \citep{oksendal2013stochastic}:
\begin{align} \label{eq:kolv-forward-eq}
    \text{Kolmogorov forward equation:}& \quad \partial_t p_{t|s} (x | \Tilde{x}) = \sum_{y} p_{t|s}(y | \Tilde{x}) R_t (y, x)
\end{align}
\begin{align} \label{eq:kolv-backward-eq}
    \text{Kolmogorov backward equation:}& \quad \partial_s p_{t|s} (x | \Tilde{x}) = - \sum_{y} R_t (\Tilde{x}, y) p_{t|s}(x | y)
\end{align}
The forward equation also induces a PDE for the marginal distribution $p_t (x)$
\begin{align} \label{eq:forward-kolv-pde}
    \partial_t p_t (x) = \sum_{y} p_t (y) R_t (y, x).
\end{align}
Using the backward equation, one can derive a Kolmogorov backward equation for expectations, also known as Dynkin’s formula \citep{oksendal2013stochastic}. In particular, we have the following lemma.
\begin{lemma} \label{eq:backward-kolv-expectation}
    Let $h$ be a test function of interest and define $u_t (x) = \E_{p_{1|t}(z | x)}[h(z)]$. Then $u_t$ satisfies the partial differential equation $\partial_t u_t (x) =  - \sum_y R_t (x, y) u_t (y)$.
\end{lemma}
\begin{proof}
    \begin{align}
    \partial_t u_t (x) 
    &= \sum_{z} h(z) \partial_t p_{1|t}(z | x)  \nonumber \\
    &= \sum_{z} h(z) - \sum_y R_t (x, y) p_{1|t} (z, y) \nonumber \\
    &= - \sum_y R_t (x, y) \sum_{z} p_{1|t} (z, y) h(z)  \nonumber \\
    &= - \sum_y R_t (x, y) u_t (y). \nonumber
\end{align}
\end{proof}

\subsection{Continuous-Time Formulation of SMC} \label{sec:appendix-conti-smc-ctmc}

\restapropcontsmc*
\begin{proof}
    Recall \cref{eq:iw_recursive}, where the importance weight is given by
    \begin{align}
        w_{t-1} (x_{t-1:T}) = \frac{\pi(x_{t-1})}{\pi(x_t)} \frac{\gamma(x_t | x_{t-1})}{q(x_{t-1} | x_t)} w_t (x_{t:T}).
    \end{align}
    We now extend it to the continuous-time setting. Let $R_t$ and $\hat{R}_t$ denote the rate matrices corresponding to the proposal $q$ and the forward noising transition $\gamma$, respectively.
    Consider a discretisation with $T$ denoising steps, indexed by time points $s = t_0 < \dots < t_i \dots < t_T = 1$, where each interval satisfies $t_{i} - t_{i-1} = \frac{1 - s}{T}$. The discrete-time importance weight at step time $s$ is then computed as
    \begin{align}
        \log w_s = \log \frac{\pi (x_s)}{\pi (x_1)} + \sum_{i=1}^T \log \frac{\gamma (x_{t_{i}} | x_{t_{i-1}})}{q (x_{t_{i-1}} | x_{t_{i}})}.
    \end{align}
    The second term in the RHS can be expanded as
    \begin{align}
        &\sum_i \log \frac{\gamma (x_{t_{i}} | x_{t_{i-1}})}{q (x_{t_{i-1}} | x_{t_{i}})}
        \!=\!\! \sum_i \log \!\left( \delta_{x_{t_{i}} \!=\! x_{t_{i\!-\!1}}} \!\!\!+\! \hat{R}_{t_i} (x_{t_{i-1}}, x_{t_{i}}) \frac{1}{T} \!\right) \!\!-\! \log \!\left( \delta_{x_{t_{i-1}} \!=\! x_{t_{i}}} \!\!\!+\! R_{t_i} (x_{t_{i}}, x_{t_{i\!-\!1}}) \frac{1}{T} \!\right) \nonumber \\
        &= \!\!\!\!\sum_{i, t_i = t_{i\!-\!1}}\!\!\!\! \log \! \left(\! 1 \!+\! \hat{R}_{t_i} (x_{t_{i}}, x_{t_{i}}) \frac{1}{T} \!\right)\! \!-\! \log \!\left(\! 1 \!+\! R_{t_i} (x_{t_{i}}, x_{t_{i}}) \frac{1}{T} \!\right)\! \!+\!\!\!\!\!\! 
        \sum_{i, t_i \ne t_{i\!-\!1}}\!\!\!\!\! \hat{R}_{t_i} (x_{t_{i\!-\!1}}, x_{t_{i}}) \!-\!  R_{t_{i}} (x_{t_{i}}, x_{t_{i\!-\!1}}) \nonumber \\
        &= \!\!\!\!\sum_{i, t_i = t_{i\!-\!1}}\!\!\!\! \hat{R}_{t_i} (x_{t_{i}}, x_{t_{i}}) \frac{1}{T} \!-\! R_{t_i} (x_{t_{i}}, x_{t_{i}}) \frac{1}{T} + \mathcal{O}(\frac{1}{T}) \!+\!\!\!\! \sum_{i, t_i \ne t_{i\!-\!1}} \!\!\!\! \hat{R}_{t_i} (x_{t_{i-1}}, x_{t_{i}}) \!-\!  R_{t_i} (x_{t_{i}}, \x_{t_{i-1}}) \nonumber 
    \end{align}
    Taking the limit $T \rightarrow +\infty$, the importance weight becomes:
    \begin{align} \label{eq:appendix-log-importance-weight}
        \!\!\!\log w_s \!=\! \log \frac{\pi (x_s)}{\pi (x_1)} \!+\!\! \int_1^s \!\! R_t (x_t, x_t) \!-\! \hat{R}_t (x_t, x_t) \dif t +\!\!\!\!\!\!\!\!\! \sum_{s\leq t, x_{t^+} \ne x_t} \!\!\!\!\!\!\!\! \log \hat{R}_t (x_t, x_{t^+}) \!-\! \log R_t (x_{t^+}, x_t).
    \end{align}
    By the fundamental theorem of calculus for piecewise differentiable functions, we have:
    \begin{align}
        \log \frac{\pi (x_s)}{\pi (x_1)} = \int_1^s -\partial_t \log \pi (x_t) + \sum_{s\leq t, x_{t^+} \ne x_t} \log \pi (x_t) - \log \pi (x_{t^+}).
    \end{align}
    If the noising process $\gamma$ is chosen such that the rate matrix satisfies $\hat{R}_t (x_t, y_t) \pi (x_t) = R_t (y_t, x_t)\pi(y_t)$, then the importance weight simplifies accordingly
    \begin{align}
        \log w_s 
        &= \int_1^s -\partial_t \log \pi (x_t) + R_t (x_t, x_t) \!-\! \hat{R}_t (x_t, x_t) \dif t \nonumber \\
        &= \int_1^s -\partial_t \log \pi (x_t) + \sum_{y_t} R_t (x_t, y_t) \frac{\pi (y_t)}{\pi (x_t)} \dif t, \nonumber 
    \end{align}
    which completes the proof.
\end{proof}
\textbf{Remark.}
\cref{prop:smc_cont} recovers the importance weights proposed in the SMC methods of \cite{holderrieth2025leaps,ou2025discrete}. In those works, the intermediate target distribution is defined as a geometric interpolation between the base and target distributions: $\pi (x_t) \propto p_{\mathrm{base}}^t (x_t) p_{\mathrm{target}}^{1-t} (x_t)$. The proposal rate matrix $R_t$ is then trained to satisfy the Kolmogorov forward equation.
In contrast, we consider a different scenario: the pretrained model is available, but the intermediate target $\pi(x_t)$ cannot be computed explicitly.
\revision{
Moreover, the importance weight in \cref{eq:appendix-log-importance-weight} can also be derived via the Radon–Nikodym derivative \citep[Appendix C.1]{campbell2024generative}, \citep[Lemma 4]{denker2025iterative}.  We instead adopt a discrete-time formulation and present a streamlined derivation to keep the exposition accessible for readers who may not be familiar with Radon–Nikodym derivative or path-measure theory.
}

We next extend the discrete SMC framework to the continuous-time setting, concentrating on the reward-tilting formulation. While this formulation has also been considered in \cite{lee2025debiasing}, our treatment proceeds from a distinct perspective. Before proceeding with the main development, we establish several auxiliary lemmas that are essential for the subsequent derivations.
\begin{lemma} \label{lemma:forwad-backward-rate-matrix-identity}
    For a continuous time Markov chain with distribution $p$ and rate matrix $R$, the rate matrix for the reverse process satisfy $\hat{R}_t(x_t, y_t) = R_t (y_t, x_t) \frac{p(y_t)}{p(x_t)}$ and $\hat{R}_t(x_t, x_t) = -\sum_{y_t \ne x_t} \hat{R}_t(x_t, y_t) = -\sum_{y_t \ne x_t} R_t (y_t, x_t) \frac{p(y_t)}{p(x_t)}$.
\end{lemma}
\begin{proof}
    See \cite[Appendix B.2]{sun2022score} for a detailed proof.
\end{proof}
\begin{lemma} \label{lemma:pde-marginal}
    For a continuous time Markov chain with distribution $p$ and rate matrix $R$, it satisfies 
    \begin{align}
        \partial_t \log p(x_t) = \sum_{y_t \ne x_t} R_t (y_t, x_t) \frac{p(y_t)}{p(x_t)} + R_t (x_t, x_t). 
    \end{align}
\end{lemma}
\begin{proof}
    By applying the forward Kolmogrov equation in \cref{eq:forward-kolv-pde}, we have
    \begin{align}
        \partial_t \log p(x_t) = \frac{\partial_t p(x_t)}{p(x_t)} = \frac{1}{p(x_t)} \sum_{y_t} R_t (y_t, x_t) p(y_t) = \sum_{y_t \ne x_t} R_t (y_t, x_t) \frac{p(y_t)}{p(x_t)} + R_t(x_t, x_t)
    \end{align}
    which completes the proof.
\end{proof}
\begin{lemma} \label{lemma:pde-expectation}
    For a continuous time Markov chain with distribution $p$ and rate matrix $R$, the function $u(x_t) = \E_{p(x_0 | x_t)}[\exp(r(x_0))]$ satisfies
    \begin{align}
        \partial_t \log u(x_t) = R_t^{\alpha=1}(x_t, x_t) - R_t (x_t, x_t), 
    \end{align}
    where $R_t^{\alpha=1}(x_t, y_t) = R_t (x_t, y_t) \frac{u(y_t)}{u(x_t)}$ and $R_t^{\alpha=1}(x_t, x_t) = -\sum_{y_t \ne x_t} R_t^{\alpha=1}(x_t, y_t)$.
\end{lemma}
\begin{proof}
    By applying \cref{eq:backward-kolv-expectation}, we have
    \begin{align}
        \partial_t \log u(x_t) 
        &= \frac{\partial_t u(x_t)}{u(x_t)} = -\frac{1}{u(x_t)} \sum_{y_t} R_t (x_t, y_t)u(y_t) \nonumber \\
        &= -\sum_{y_t \ne x_t} R_t (x_t, y_t)\frac{u(y_t)}{u(x_t)} - R_t(x_t, x_t) \nonumber \\
        &= R_t^{\alpha=1}(x_t, x_t) - R_t (x_t, x_t), \nonumber
    \end{align}
    which completes the proof.
\end{proof}
We are now ready to prove the result in \cite{lee2025debiasing}.
\begin{proposition}[Continuous-Time SMC for Reward-Tilting \citep{lee2025debiasing}] \label{prop:ct-smc-reward-tilting}
    Let $p_\theta(x_t)$ denote a pretrained diffusion model, $R_t$ the rate matrix generating the desnoising probability path, and $\hat{R}_t$ the corresponding rate matrix for the forward noising path. The intemediate target distributino is defined as $\pi (x_t) = p_\theta (x_t) u^\alpha(x_t)$, where $u(x_t) = \E_{p_\theta (x_0 | x_t)}[\exp(r(x_0))]$ is the reward-tilting functioin. Let $Q_t$ be the proposal rate matrix in SMC; the importance weight is then given by
    \begin{align}
        \log& w_s = 
        \int_1^s Q_t (x_t, x_t) - R_t (x_t, x_t) \dif t + \!\!\!\sum_{s\leq t, x_{t^+} \ne x_t}\!\!\!\! \log R_t (x_{t^+}, x_t) - \log Q_t (x_{t^+}, x_t) \nonumber \\
        &\quad \!+\! \int_1^s \alpha \left( R_t (x_t, x_t) \!-\! R_t^{\alpha=1} (x_t, x_t) \right) \dif t \!+\! \!\!\!\!\!\sum_{s\leq t, x_{t^+} \ne x_t}\!\!\!\!\!\!\! \alpha \left ( \log R_t^{\alpha=1} (x_{t^+}, x_t) \!-\! \log R_t (x_{t^+}, x_{t}) \right), \nonumber
    \end{align}
    where $R_t^{\alpha=1} (x_t, y_t) = R_t(x_t, y_t) \frac{u(y_t)}{u(x_y)}$ and $R_t^{\alpha=1}(x_t, x_t) = -\sum_{y_t \ne x_t} R_t(x_t, y_t) \frac{u(y_t)}{u(x_y)}$.
\end{proposition}
\begin{proof}
    By the derivation of \cref{prop:smc_cont}, it gives that the importance weight takes the form
    \begin{align}
        \log w_s 
        &= \underbrace{\int_1^s -\partial_t \log \pi (x_t) + Q_t (x_t, x_t) - \hat{R}_t (x_t, x_t) \dif t }_{\textcircled{1}} \nonumber \\
        &\qquad + \!\!\underbrace{\sum_{s\leq t, x_{t^+} \ne x_t}\!\! \log \pi (x_t) - \log \pi (x_{t^+}) + \log \hat{R}_t (x_t, x_{t^+}) - \log Q_t (x_{t^+}, x_t)}_{\textcircled{2}}.
    \end{align}
    By applying \cref{lemma:forwad-backward-rate-matrix-identity,lemma:pde-marginal,lemma:pde-expectation}, we can expand \textcircled{1} as
    \begin{align}
        \textcircled{1} 
        &= \!\int_1^s \!\!\!-\!\left(\! \alpha \partial_t \log u(x_t) \!+\!\!\! \sum_{y_t \ne x_t}\!\! R_t (y_t, x_t) \frac{p_\theta (y_t)}{p_\theta (x_t)} \!+\! R_t (x_t, x_t) \!\right)\! \!+\! Q_t (x_t, x_t) \!+\!\!\! \sum_{y_t \ne x_t}\!\! R_t(y_t, x_t) \frac{p_\theta (y_t)}{p_\theta (x_t)} \dif t \nonumber \\
        &= \!\int_1^s - \alpha \partial_t \log u(x_t) - R_t (x_t, x_t) + Q_t (x_t, x_t) \dif t \nonumber \\
        &= \!\int_1^s \alpha \left( R_t (x_t, x_t) - R_t^{\alpha=1} (x_t, x_t) \right) - R_t (x_t, x_t) + Q_t (x_t, x_t) \dif t. \nonumber
    \end{align}
    Similarly, by applying \cref{lemma:forwad-backward-rate-matrix-identity}, $\textcircled{2} $ follows
    \begin{align}
        \textcircled{2} 
        &= \sum_{s\leq t, x_{t^+} \ne x_t} \alpha \left ( \log u (x_t) - \log u (x_{t^+}) \right) + \log R_t (x_{t^+}, x_t) - \log Q_t (x_{t^+}, x_t) \nonumber \\
        &= \sum_{s\leq t, x_{t^+} \ne x_t} \alpha \left ( \log R_t^{\alpha=1} (x_{t^+}, x_t) - \log R_t (x_{t^+}, x_{t}) \right) + \log R_t (x_{t^+}, x_t) - \log Q_t (x_{t^+}, x_t). \nonumber
    \end{align}
    where the second equation follows from the identity
    \begin{align}
        \log u(y_t) - \log u(x_t) = \log R_t^{\alpha=1}(x_t, y_t) - \log R_t (x_t, y_t),
    \end{align}
    which is followed by the definition of $R_t (x_t, y_t)$.
    Combining $\textcircled{1}$ and $\textcircled{2}$, the full expression for the importance weight becomes
    \begin{align}
        \log& w_s = 
        \int_1^s Q_t (x_t, x_t) - R_t (x_t, x_t) \dif t + \!\!\!\sum_{s\leq t, x_{t^+} \ne x_t}\!\!\!\! \log R_t (x_{t^+}, x_t) - \log Q_t (x_{t^+}, x_t) \nonumber \\
        &\quad \!+\! \int_1^s \alpha \left( R_t (x_t, x_t) \!-\! R_t^{\alpha=1} (x_t, x_t) \right) \dif t \!+\! \!\!\!\!\!\sum_{s\leq t, x_{t^+} \ne x_t}\!\!\!\!\!\!\! \alpha \left ( \log R_t^{\alpha=1} (x_{t^+}, x_t) \!-\! \log R_t (x_{t^+}, x_{t}) \right), \nonumber
    \end{align}
    which completes the proof.
\end{proof}

\section{Implementation Details of Computing Importance Weight }

In masked diffusion models, although ancestor sampling \citep{austin2021structured,sahoo2024simple,shi2024simplified} is the de facto method for inference, low-confidence sampling \citep{chang2022maskgit} is more widely used in practice due to its stronger empirical performance. However, this approach makes it challenging to explicitly compute the importance weights. In this section, we first provide a brief recap of the main sampling schemes used in masked diffusion models, and then present a method to address the difficulty of computing importance weights under low-confidence sampling.

\subsection{Sampling Schemes in Masked Diffusion Models} \label{sec:appendix-diff-sampling-methods}

\textbf{MDM Sampling \citep{sahoo2024simple}.} 
MDM sampling is the de facto method for inference in masked diffusion models. Given a trained denoiser $\mu_\theta$, which predicts the clean data $x_0$, MDM sampling performs ancestor sampling to generate samples according to
\begin{equation}
    p_\theta (x_{t-1} | x_t) =
    \begin{cases}
        \cat(x_{t-1}; x_t) & x_t \neq \mask\\
        \cat\left(x_{t-1}; \frac{(1-\alpha_{t-1})\mask + (\alpha_{t-1} - \alpha_t) \mu_\theta(x_t)}{1 - \alpha_t}\right) & x_t = \mask
    \end{cases}
\end{equation}
While theoretically sound, a major limitation of MDM sampling is that once a latent variable $x_t$ is assigned a non-mask category during the unmasking process, it becomes immutable. Consequently, any errors made during unmasking are irreversible and persist in the final generated samples.

\textbf{ReMDM Sampling \citep{wang2025remasking}.}
ReMDM sampling is a modification of MDM that allows previously unmasked tokens to be remasked during the unmasking process. The posterior is constructed so that the forward marginal $p(x_t|x_0)$ remains identical to that of masked diffusion in \cref{eq:diffusion-forward}:
\begin{equation}
    p_\sigma (x_{t-1} | x_t, x_0) =
    \begin{cases}
        \cat(x_{t-1}; (1 - \sigma_t) x_t + \sigma_t \mask) & x_t \neq \mask\\
        \cat\left(x_{t-1}; \frac{\alpha_{t-1} - (1 - \sigma_t)\alpha_{t}}{1 - \alpha_t} x_0 + \frac{1 -\alpha_{t-1} - \sigma_t \alpha_{t}}{1 - \alpha_t} \mask \right) & x_t = \mask
    \end{cases}.
\end{equation}
Here $\sigma_t$ is the remasking schedule. To ensure the posterior remains valid, it must satisfy the constraint:
\begin{align}
    0 \le \sigma_t \le \min \left\{  1, \frac{1-\alpha_{t-1}}{\alpha_t} \right\}.
\end{align}
The reverse unmasking process is then parameterised as
\begin{align} \label{eq:remdmd-sampling}
    p_\theta (x_{t-1} | x_t) = p_\sigma (x_{t-1} | x_t, \mu_\theta (x_t)).
\end{align}
Notably, the ReMDM training objective is a reweighted version of the standard masked diffusion loss in \cref{eq:mdm-loss}.
Thus, we can take a pretrained masked diffusion model, and use the ReMDM sampling in \cref{eq:remdmd-sampling} for inference.

\textbf{Low-Confidence Sampling \citep{chang2022maskgit}.}
Low-confidence sampling is the most commonly used method in discrete diffusion. In brief, at each denoising step, the denoiser $\mu_\theta$ predicts the clean data $x_0$, and tokens with low confidence, which are measured as the maximum logit of $\mu_\theta$ at each position, are selectively remasked for further refinement.
Formally, the reverse unmasking process can be parametrised as
\begin{align} \label{eq:low-conf-sampling}
    p_\theta (x_{t-1} | x_t) = \sum_{x_0} p_\theta (x_0 | x_t) \mathbf{1}_{x_{t-1}[l] = x_{0}[l]}, \quad l = \argmax_{l \in \{1, \dots, L\} } \max (\mu_\theta (x_t)[l]) \land x_t[l]=\mask ,
\end{align}
where $x[l]$ denotes the $l$-th token of $x$ of $L$ length, and $\max (v)$ returns the maximum value of the vector $v$.
For clarity, here we only consider unmasking a single token at each step. In practice, multiple tokens can be unmasked simultaneously by using the same strategy.

\subsection{First-Order Approximation with Gumbel-Softmax Relaxation} \label{sec:appendix-gumbel-softmax}

To apply the first-order approximately optimal proposal, we need to compute $\nabla_{x_t} \hat{r}(x_t)$, where $\hat{r}(x_t) = \frac{1}{M} \sum_{m=1}^M r(x_0^{(m)}), \quad x_0^{(m)} \sim p_\theta (x_0 | x_t)$ as defined in \cref{eq:reward-est}.
However, because $x_0^{(m)}$ is drawn via categorical sampling, $\hat{r}(x_t)$ is not differentiable with respect to $x_t$.
To address this, we use the Gumbel–Softmax reparameterization trick to obtain a differentiable surrogate.
Concretely, we break the computation of $\hat{r}(x_t)$ into three steps:
\begin{itemize}
    \item[1.] compute the denoising logits: $p = \mu_\theta (x_t)$, where $\mu_\theta$ is the denoising model;
    \item[2.] sample $x_0$: $x_0^{(m)} \sim \mathrm{Cat}(x; p)$;
    \item[3.] evaluate the reward: $\hat{r}(x_t) = \frac{1}{M} \sum_{m=1}^M r(x_0^{(m)})$.
\end{itemize}
Following \cite{grathwohl2021oops}, we treat both $r$ and $\mu_\theta$ as functions that accept continuous inputs so that their gradients are well defined (steps 1 and 3). For step 2, we replace the categorical draw with its Gumbel–Softmax relaxation \citep{jang2016categorical}, making the sample $x_0^{(m)}$ differentiable with respect to $p$.
Using these relaxations, the gradient can be obtained by the chain rule:
\begin{align} \label{eq:gumbel-softmax-grad-approx}
    \nabla_{x_t} \hat{r}(x_t) \approx \frac{1}{M} \sum_{m=1}^M \frac{\partial r(x_0^{(m)})}{\partial x_0^{(m)}} \frac{\partial x_0^{(m)}}{\partial p} \frac{\partial p}{\partial x_t}, \quad x_0^{(m)} \sim p_\theta (x_0 | x_t).
\end{align}
The first and last factors are provided by the differentiability of $r$ and $\mu_\theta$, while the middle term is approximated using the Gumbel–Softmax relaxation.

\textbf{Remark.}
The gradient approximation in \cref{eq:gumbel-softmax-grad-approx} requires the reward $r(x_0)$ to be differentiable w.r.t. $x_0$.
In the setting of non-differentiable rewards, one can apply the REINFORNCE gradient estimator \citep{williams1992simple}. Specifically, we have
\begin{align}
    \nabla_{x_t} \hat{r}(x_t) 
    &= \nabla_{x_t} \E_{p_\theta (x_0 | x_t)} [r(x_0)] \nonumber \\
    &=  \sum_{x_0} p_\theta (x_0 | x_t) \nabla_{x_t} \log p_\theta (x_0 | x_t) r(x_0) \nonumber \\
    &= \E_{p_\theta (x_0 | x_t)} [\nabla_{x_t} \log p_\theta (x_0 | x_t) r(x_0)]. \nonumber
\end{align}
Thus, the gradient can be approximated via
\begin{align}
    \nabla_{x_t} \hat{r}(x_t) \approx \frac{1}{M} \sum_{m=1}^M  r(x_0^{(m)}) \nabla_{x_t} \log p_\theta (x_0^{(m)} | x_t), \quad x_0^{(m)} \sim p_\theta (x_0 | x_t).
\end{align}

\subsection{Computing Importance Weight with Low-Confidence Sampling} \label{sec:appendix-ratio-low-confidence}
To apply SMC to masked diffusion models, one must compute the log-ratio in the importance weight, as in \cref{eq:important-weight-tilting}:
\begin{align}
    \log p_\theta (x_{t-1} | x_t) - \log q_\phi (x_{t-1} | x_t).
\end{align}
While this computation is straightforward for MDM and ReMDM sampling, it becomes tricky for low-confidence sampling.
The difficulty arises because $p$ and $q$ rely on different denoisers, denoted as $\mu_\theta$ and $\mu_\phi$, respectively. 
If one strictly follows the rule in \cref{eq:low-conf-sampling}, the log-ratio often collapses to zero whenever
\begin{align}
    l^p \ne l^q, \quad \text{where}\ l^q = \argmax_{l} \max (\mu_\phi (x_t)[l]) \land x_t[l]=\mask.
\end{align}
As a result, both SMC and the training objective in \cref{eq:log-var-upper-bound} become ineffective in practice.
To address this issue, we adopt a strategy in which both $p_\theta$ and $q_\phi$ use the same logit $\mu_\phi(x_t)$ to determine the remasked position $l$:
\begin{equation}
\begin{aligned}
    p_\theta (x_{t-1} | x_t) &= \sum\nolimits_{x_0} p_\theta (x_0 | x_t) \mathbf{1}_{x_{t-1}[l] = x_{0}[l]}\\
    q_\phi (x_{t-1} | x_t) &= \sum\nolimits_{x_0} q_\phi (x_0 | x_t) \mathbf{1}_{x_{t-1}[l] = x_{0}[l]}
\end{aligned}
\quad l = \argmax_{l \in \{1, \dots, L\} } \max (\mu_\phi (x_t)[l]) \land x_t[l]=\mask ,
\end{equation}
Under this formulation, the log-ratio can be computed as
\begin{align}
     \log \frac{p_\theta (x_{t-1} | x_t)}{q_\phi (x_{t-1} | x_t)} = \sum_{l \in \{1,\dots,L | x_{t-1}[l] \ne \mask, x_{t}[l] = \mask \}} \log p_\theta (x_{t-1}[l]\ \mid \ x_t[l]) - \log q_\phi (x_{t-1}[l]\ \mid \ x_t[l])
\end{align}
This modification ensures that the importance weights remain well-defined under low-confidence sampling, enabling SMC to be applied effectively.

\begin{figure}[!t]
    \centering
    \begin{minipage}{\linewidth}
        \centering
        \begin{minipage}{0.24\linewidth}
            \centering
            \includegraphics[width=.99\linewidth]{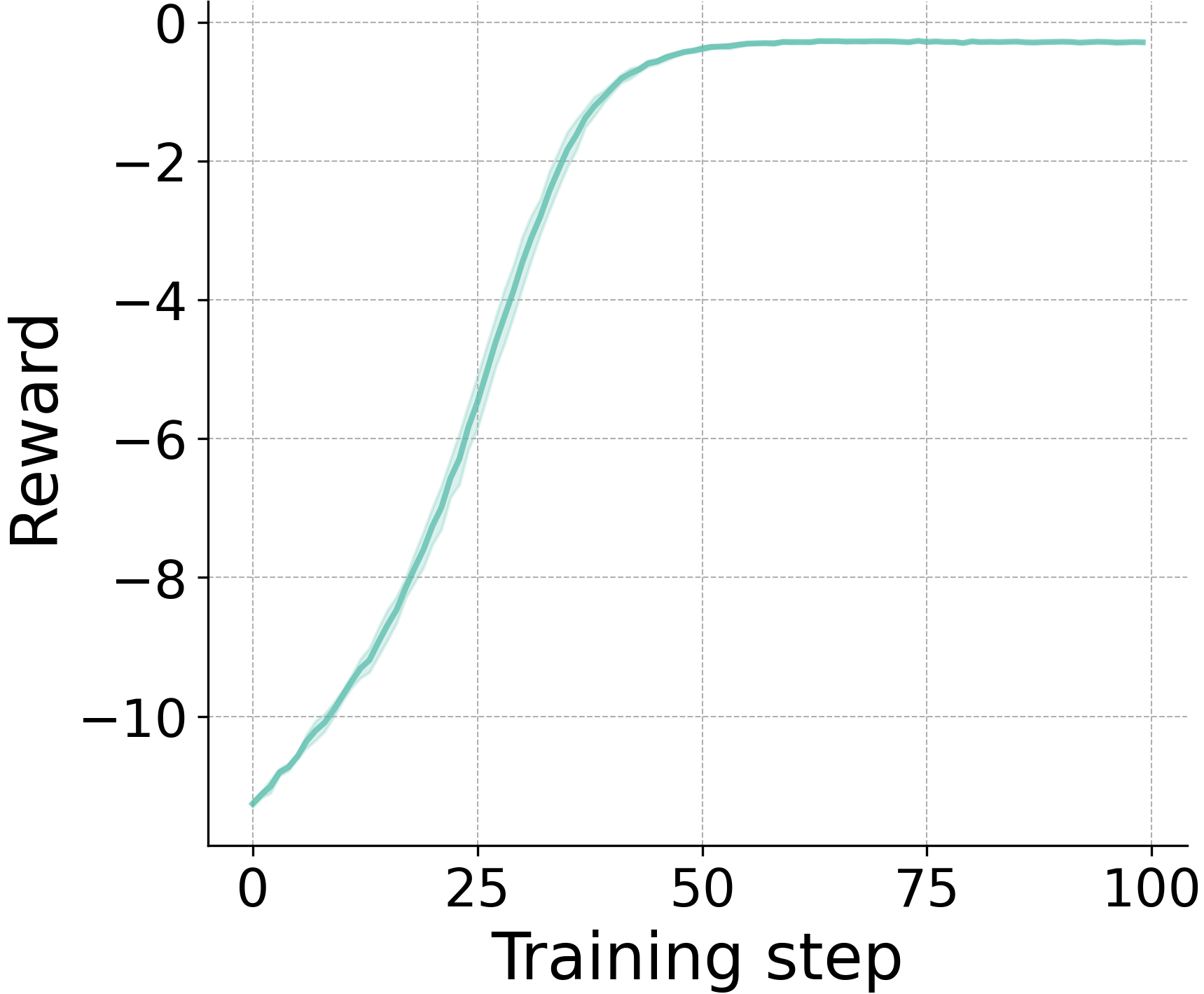}
            \subcaption{MoG}
            \label{fig:gmm_rewards_training}
        \end{minipage}
        \begin{minipage}{0.24\linewidth}
            \centering
            \includegraphics[width=.99\linewidth]{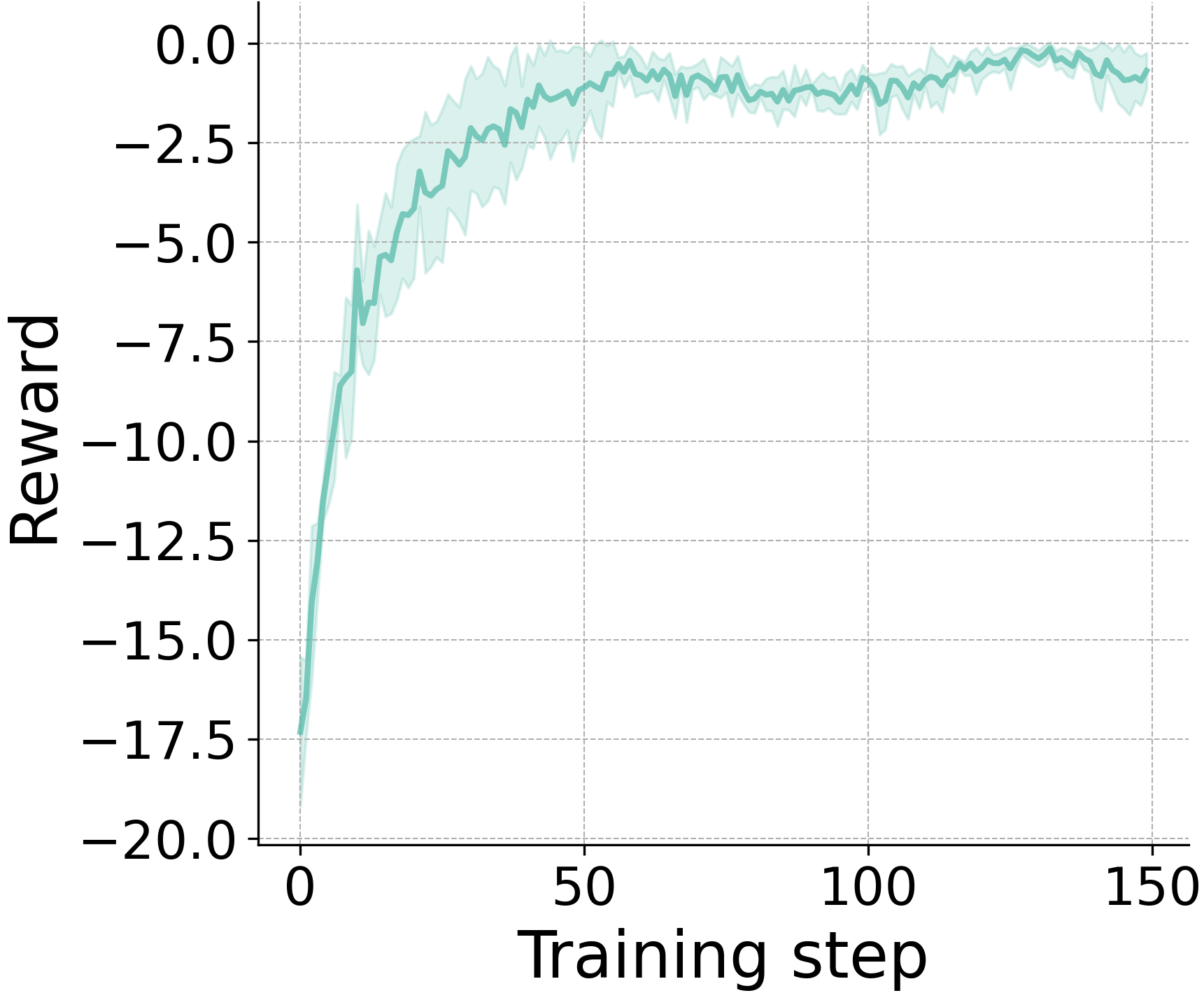}
            \subcaption{MNIST}
            \label{fig:mnist_rewards_training}
        \end{minipage}
        \begin{minipage}{0.24\linewidth}
            \centering
            \includegraphics[width=.99\linewidth]{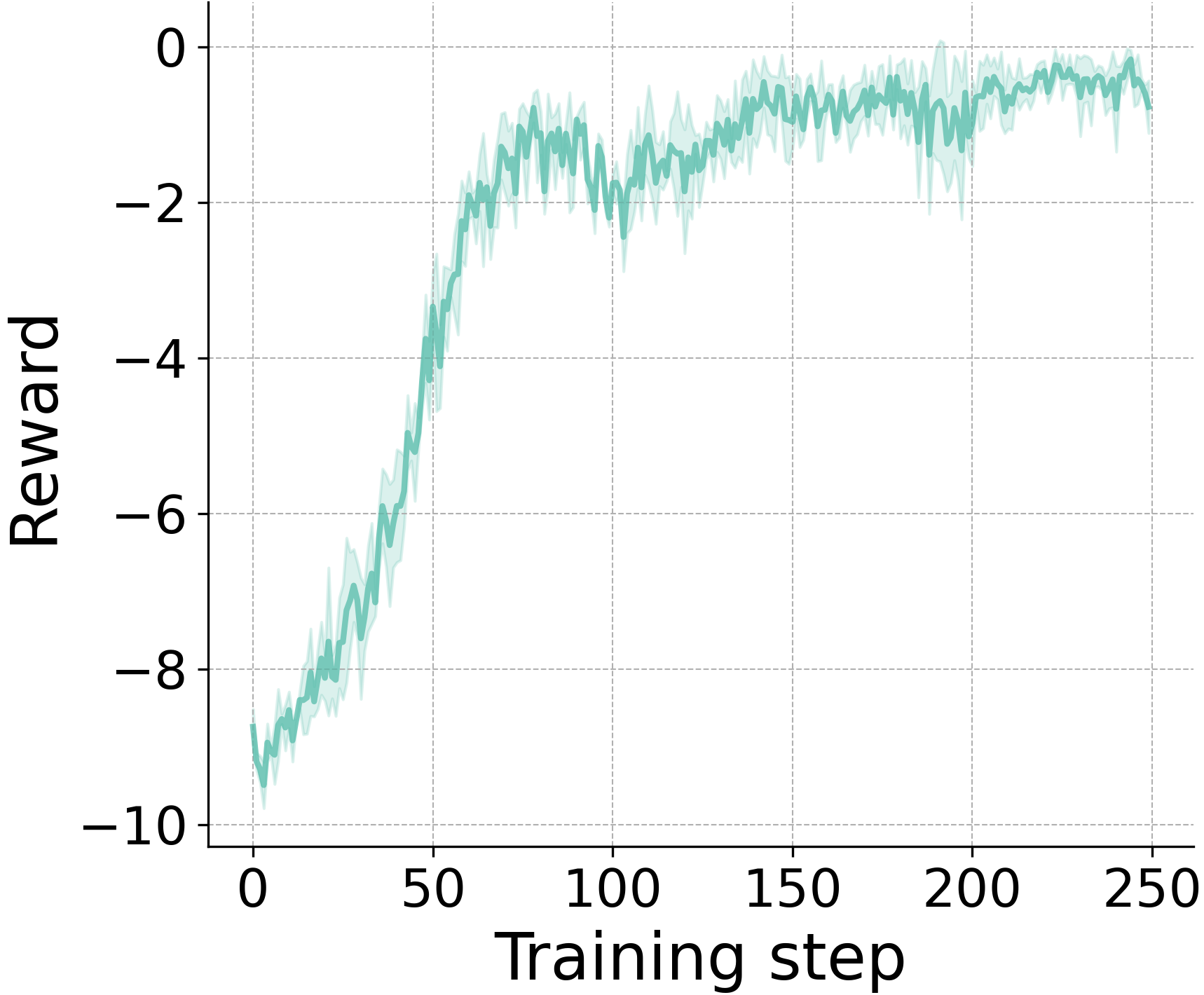}
            \subcaption{Language Modelling}
            \label{fig:toxicity_rewards_training}
        \end{minipage}
        \begin{minipage}{0.24\linewidth}
            \centering
            \includegraphics[width=.99\linewidth]{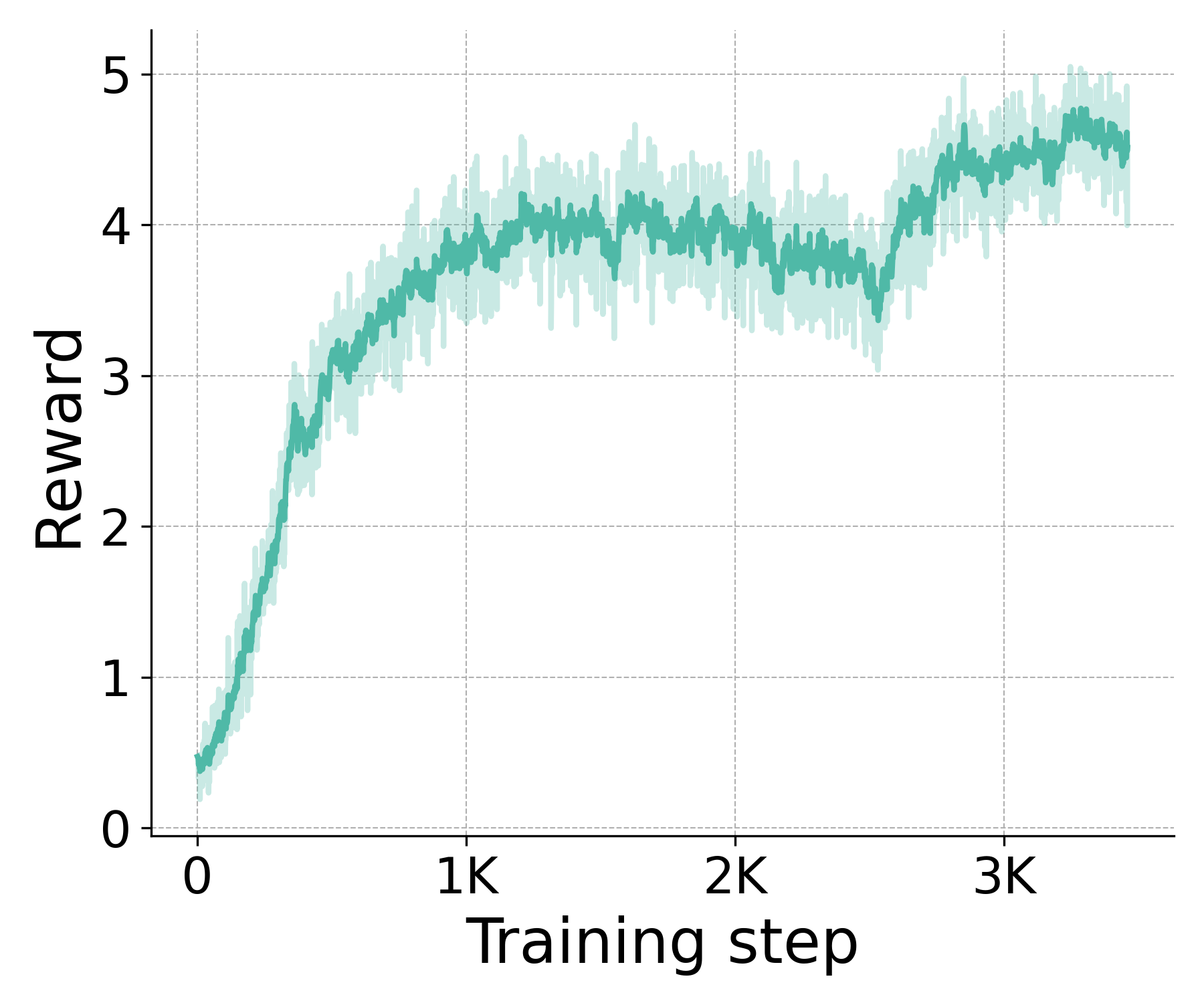}
            \subcaption{Biology Design}
            \label{fig:dna_rewards_training}
        \end{minipage}
    \end{minipage}
    \begin{minipage}{\linewidth}
        \centering
        \begin{minipage}{0.24\linewidth}
            \centering
            \includegraphics[width=.99\linewidth]{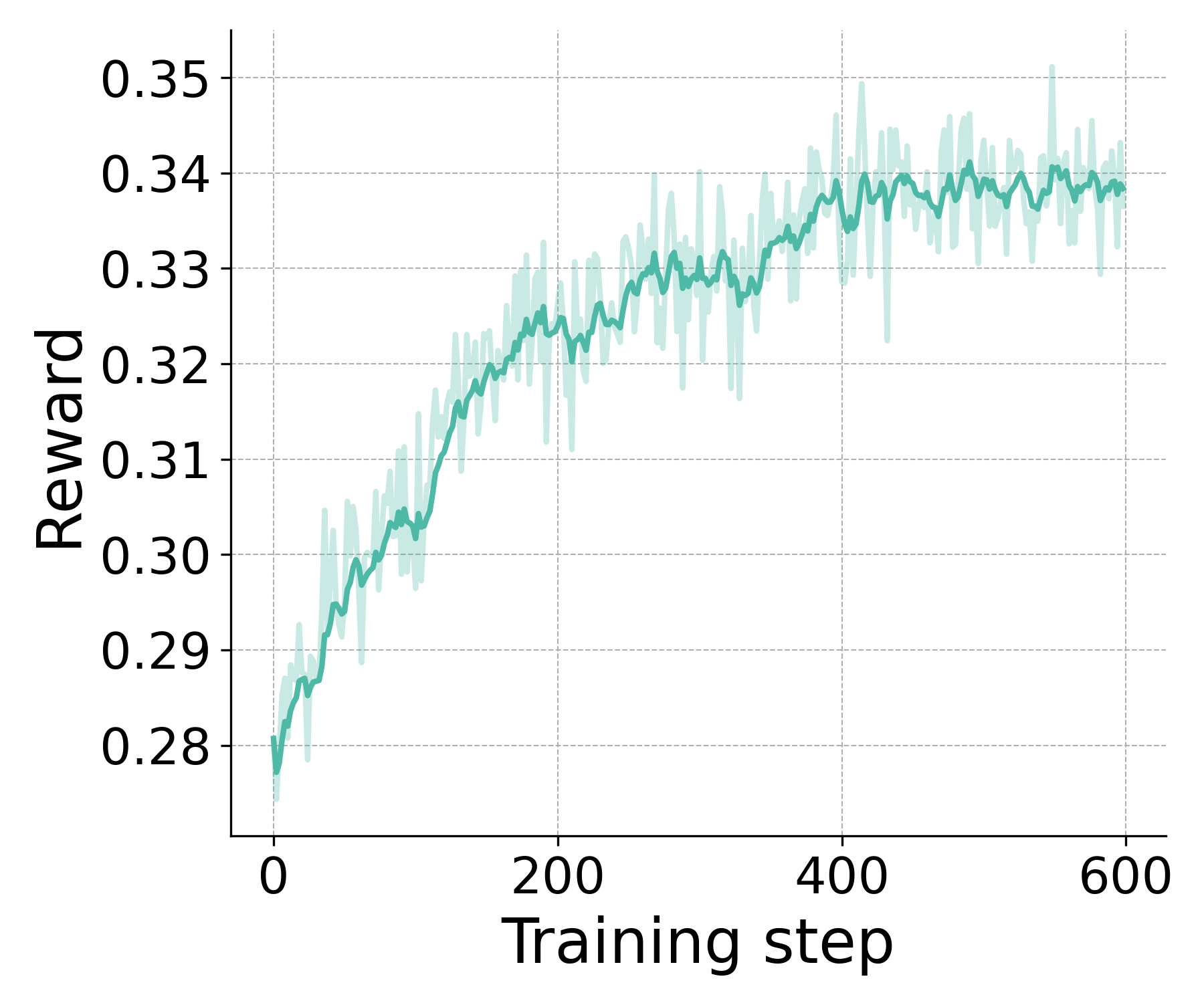}
            \subcaption{HPSv2}
            \label{fig:hpsv2_rewards_training}
        \end{minipage}
        \begin{minipage}{0.24\linewidth}
            \centering
            \includegraphics[width=.99\linewidth]{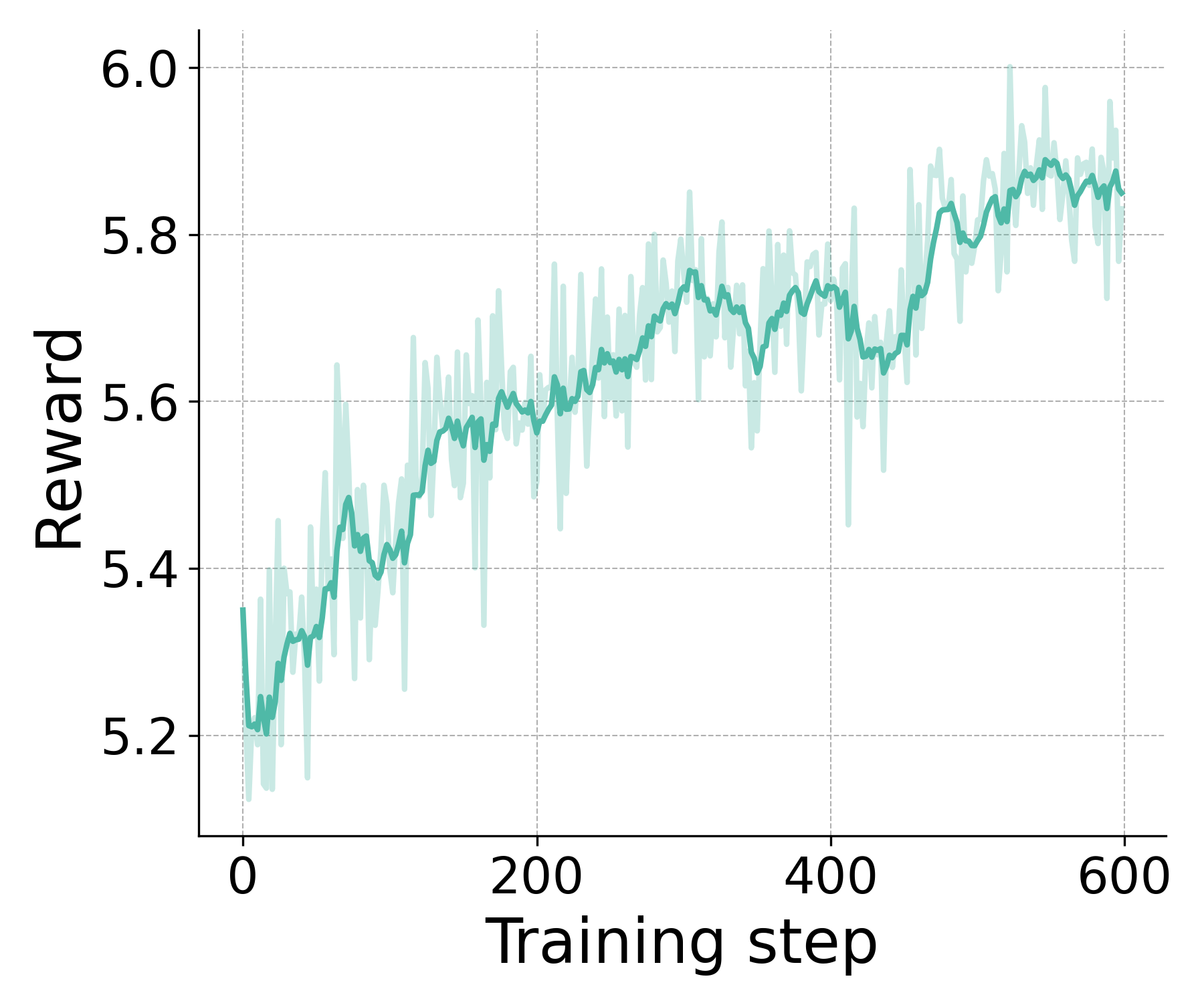}
            \subcaption{Aesthetic Score}
            \label{fig:aesthetic_rewards_training}
        \end{minipage}
        \begin{minipage}{0.24\linewidth}
            \centering
            \includegraphics[width=.99\linewidth]{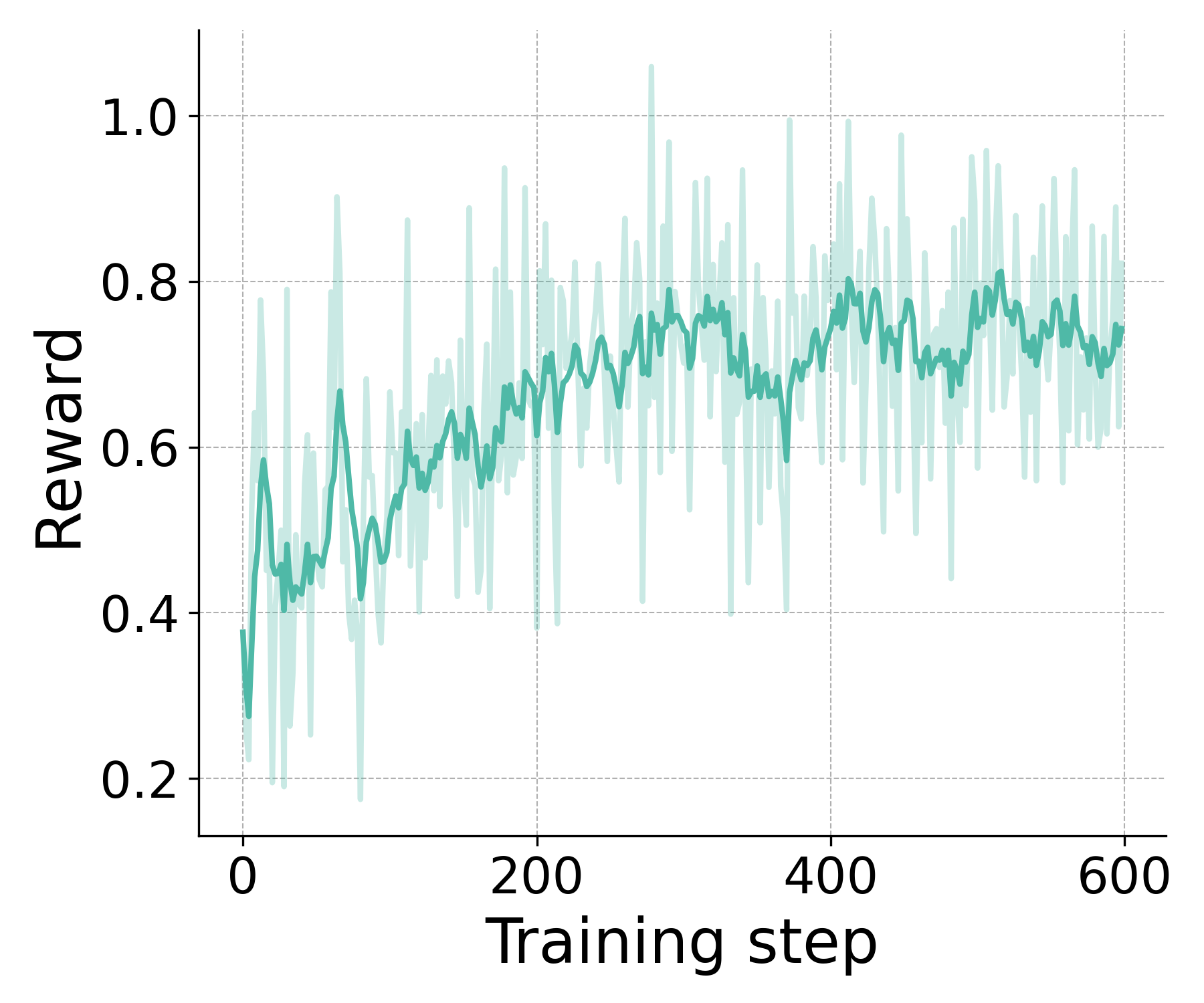}
            \subcaption{ImageReward}
            \label{fig:imagereward_rewards_training}
        \end{minipage}
    \end{minipage}
    \vspace{-2mm}
    \caption{Reward convergence curves for different experiments throughout the finetuning process.}
    \label{fig:all_training_rewards}
    \vspace{-2mm}
\end{figure}

\section{Experimental Setting and Additional Results} \label{sec:appendix_exp}

In this section, we provide the details of experimental settings and addtional exeperiemental resuts.

\subsection{Details of Experimental Setting} \label{sec:appendix-exp-setting-details}

We first describe the hyperparameters used in the SMC variants, and then discuss the training details of the learnable amortized proposal.

\subsubsection{Choice of Hyperparameters in SMC}

\begin{wraptable}{r}{0.52\linewidth}
    \small
    \centering
    \caption{Hyperparameters used in the SMC methods.} 
    \label{tab:hyper-param-smc}
    \vspace{-2mm}
    \begin{threeparttable}
    \resizebox{\linewidth}{!}{
        \begin{tabular}{lcccc}
            \toprule
            & $\alpha$ & $\lambda_t$ & $M$ & $T$ \\
            \midrule
            MoG & $1$ & $1-\frac{t}{T}$ & $10$ & $100$ \\
            MNIST & $1$ & $\min (1.05^{T-t} - 1, 1)$ & $10$ & $100$ \\
            Language Modelling & $0.2$ & $1-\frac{t}{T}$ & $4$ & $100$ \\
            Biology Design & $0.1$ & $1-\frac{t}{T}$ & $4$ & $128$ \\
            Text-to-Image Generation & $0.01$ & $ 1 - \frac{t}{T}$ & $1$ & $48$ \\
            \bottomrule
        \end{tabular}}
        \vspace{-4mm}
    \end{threeparttable}
\end{wraptable}
As described in \cref{sec:smc-recipe}, there are four key hyperparameters in our proposed SMC framework for the reward-tilting target: (i) the KL-regularization coefficient $\alpha$, (ii) the reward-twisted schedule $\lambda_t$, (iii) the number of Monte Carlo samples $M$, and (iv) the number of denoising steps $T$.
\cref{tab:hyper-param-smc} summarises the values of these hyperparameters used in our experiments.
In practice, instead of using the mean to estimate the intermediate reward in \cref{eq:reward-est}, we employ the log-sum-exp operation for improved stability, following \cite{singhal2025general}:
\begin{align}
    \hat{r}(x_t) = \log \left(\frac{1}{M} \sum_{m=1}^M \exp(r(x_0^{(m)})) \right), \quad x_0^{(m)} \sim p_\theta (x_0 | x_t).
\end{align}
Additionally, we provide ablation studies in \cref{sec:appendix-ablation-smc-hyperparam} to investigate the effects of $\lambda_t$ and $M$.

\subsubsection{Training Details of the Amortised Proposal} 

\textbf{Synthetic Experiments.}
In this experiment, we take the MDLM \citep{sahoo2024simple} as the pretrained diffusion model. 
Finetuning is performed on a single NVIDIA A6000 GPU with a batch size of $32$ for MNIST and $128$ for MoG.
The model is trained for $30$ epochs on MNIST and $20$ epochs on MoG, with $5$ optimisation steps per epoch.
To avoid out-of-memory issues, we compute the loss over $10$ randomly sampled time steps $t$ instead of using gradient accumulation, and choose $M=10$ to estimate the reward. The Adam optimiser \citep{adam2014method} is applied to train both the model and $F_\psi$, with a learning rate of $0.001$ for MoG and $0.0001$ for MNIST.

\textbf{Language Modelling.}
This experiment closely follows \cite{singhal2025general}. The pretrained language model used is MDLM\footnote{\url{https://huggingface.co/kuleshov-group/mdlm-owt}}
 \citep{sahoo2024simple}, which is trained on the OpenWebText dataset.
We perform full-parameter finetuning on a single NVIDIA A6000 GPU with a batch size of $32$. Training is conducted for $50$ epochs, with $5$ optimisation steps per epoch.
To avoid the memory issue, at each optimisation step we compute the loss using one randomly selected time step $t$, together with a fixed $t=0$. During training, rewards are scaled by a factor of $20$, and estimated with $M=20$ Monte Carlo samples.
Both the model and $F_\psi$ are optimised using Adam \citep{adam2014method} with a learning rate of $0.0001$.

\textbf{Biology Design.}  This experiment focuses on regulatory DNA sequence generation. We use the pretrained masked discrete diffusion model from \citet{wang2024fine} which has been trained on a dataset of $\sim700$k DNA sequences \citep{gosai2023machine}. We perform full-parameter finetuning on a single NVIDIA RTX 3090 GPU. Training is conducted for 350 epochs, with 10 optimisation steps per epoch. We use a batch size of 64 and use a sampling mix of $0.9:0.1$ of on-policy from $q_\phi$ and off-policy samples from $p_\theta$. To manage memory usage, we do a gradient accumulation of the loss at each timestep before taking an optimisation step. Additionally, we only consider the final 50 of the 128 timesteps for loss calculation, following \cite{wang2024fine}. We also add a negative entropy term of the form $ \sum_{x_{t-1}} q_\phi (x_{t-1} | x_t) \log q_\phi (x_{t-1} | x_t)$ to the loss with a coefficient of $2.5$\footnote{We empirically observe that the best value for the entropy coefficient varies proportionally with the reward scaling factor maintaining a ratio of $0.002-0.003$.}. We observe empirically that it helps in preventing mode collapse during training. For rewards, we use a scaling factor of $1000$ and
estimate it using just $M=1$ Monte Carlo sample. Both the model and $F_\psi$ are optimized using the AdamW optimizer \citep{loshchilov2017decoupled} with a learning rate of $1 \times 10^{-5}$.

\textbf{Text-to-Image Generation.}
To finetune the Meissonic\footnote{\url{https://huggingface.co/MeissonFlow/Meissonic}} model \citep{bai2024meissonic}, we adopt low-rank adaptation (LoRA) \citep{hu2022lora} for parameter-efficient training.
For training hyperparameters, we largely follow the DDPO \citep{black2023training} implementation\footnote{\url{https://github.com/kvablack/ddpo-pytorch}}, with details provided here for completeness.
All experiments are run on 8$\times$NVIDIA H100 GPUs with a per-GPU batch size of $8$. With $4$-step gradient accumulation, this yields an effective batch size of $256$.
We train for $300$ epochs, where each epoch consists of sampling $512$ trajectories from the reference distribution $q_{\mathrm{ref}}$ and performing $4$ optimisation steps.
The learning rate is fixed at $3 \times 10^{-4}$ for both the diffusion model and $F_\psi$ without further tuning.
We employ the AdamW optimiser \citep{loshchilov2017decoupled} with gradient clipping at a norm of $1$.

During training, we adopt classifier-free guidance \citep{ho2022classifier} with a guidance scale of $5$, using the negative prompt “worst quality, low quality, low res, blurry, distortion, watermark, logo, signature, text, jpeg artifacts, sketch, duplicate, ugly, identifying mark”, following the inference script provided by Meissonic.
Reward rescaling proves to be critical for stable optimisation \citep{liu2024efficient}. Specifically, we multiply the reward by a coefficient $\beta$, setting $\beta = 100$ for both the Aesthetic Score and ImageReward, and $\beta = 10,000$ for HPSv2. The coefficient is linearly annealed from $0$ to its maximum value over the first $25$ epochs.
For ImageReward and HPSv2, no KL regularisation between the fine-tuned and pretrained models is applied. However, for the Aesthetic Score, we observe that incorporating a KL term of the form $\mathbb{KL} (q_\phi (x_{t-1} | x_t) || p_\theta (x_{t-1} | x_t))$ with a coefficient of $0.01$ enhances training stability, consistent with prior observations \citep{fan2023dpok}.

\subsection{Additional Experimental Results}

\subsubsection{Ablation Study of SMC Hyperparameters.} \label{sec:appendix-ablation-smc-hyperparam}

\begin{figure}[!t]
    \centering
    \begin{minipage}{0.35\linewidth}
        \centering
        \includegraphics[width=.9\linewidth]{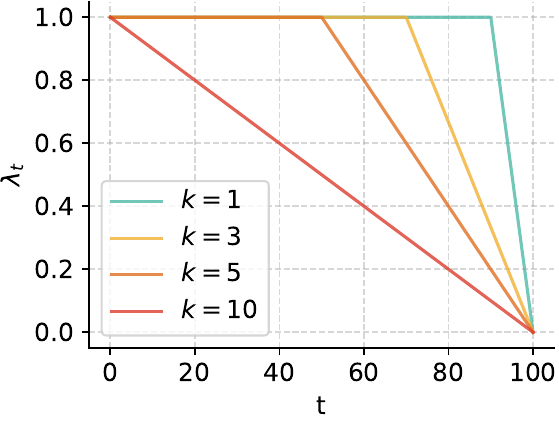}
        \vspace{-4mm}
        \caption{Plot of $\lambda_t$ schedules of the family, $\lambda_t(k) = \min \Big(1, \frac{10}{k}(1-\frac{t}{T})\Big)$, for different values of $k$.}
        \label{fig:lambda_t_plot}
        \vspace{-2mm}
    \end{minipage}
    \hfill
    \begin{minipage}{0.63\linewidth}
        \centering
        \includegraphics[width=.9\linewidth]{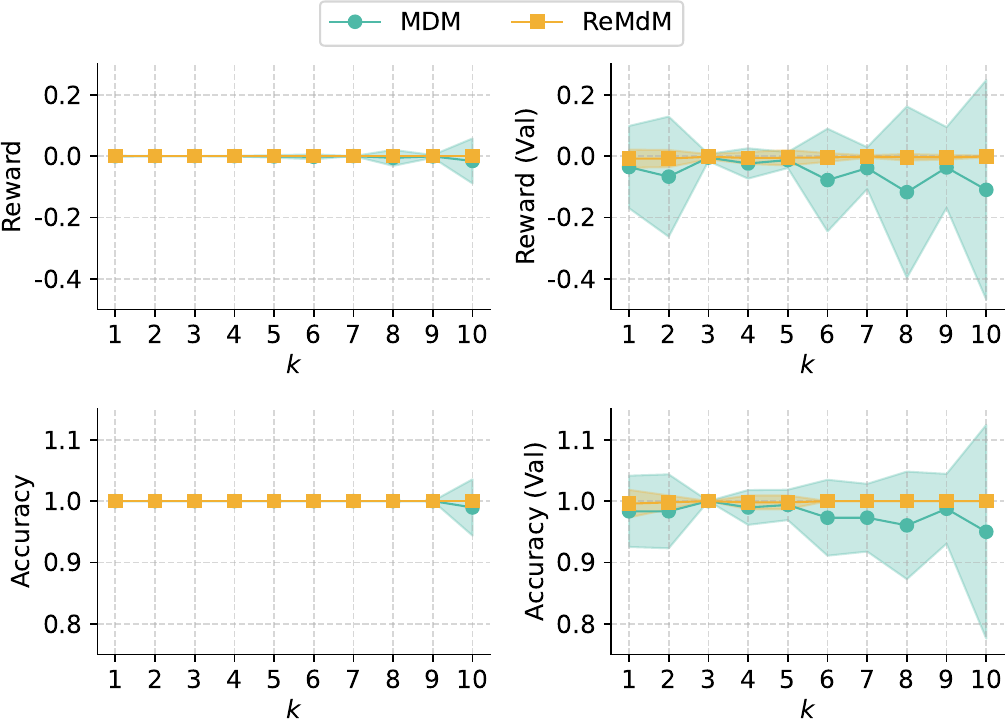}
        \vspace{-4mm}
        \caption{Comparing \oursgrad ($N=16$) with different $\lambda_t$ schedules on reward-tiled binary MNIST.}
        \label{fig:smc_binarized_mnist_lambda_t_ablation}
        \vspace{-2mm}
    \end{minipage}
\end{figure}

\begin{figure}[!t]
    \centering
    \includegraphics[width=.9\linewidth]{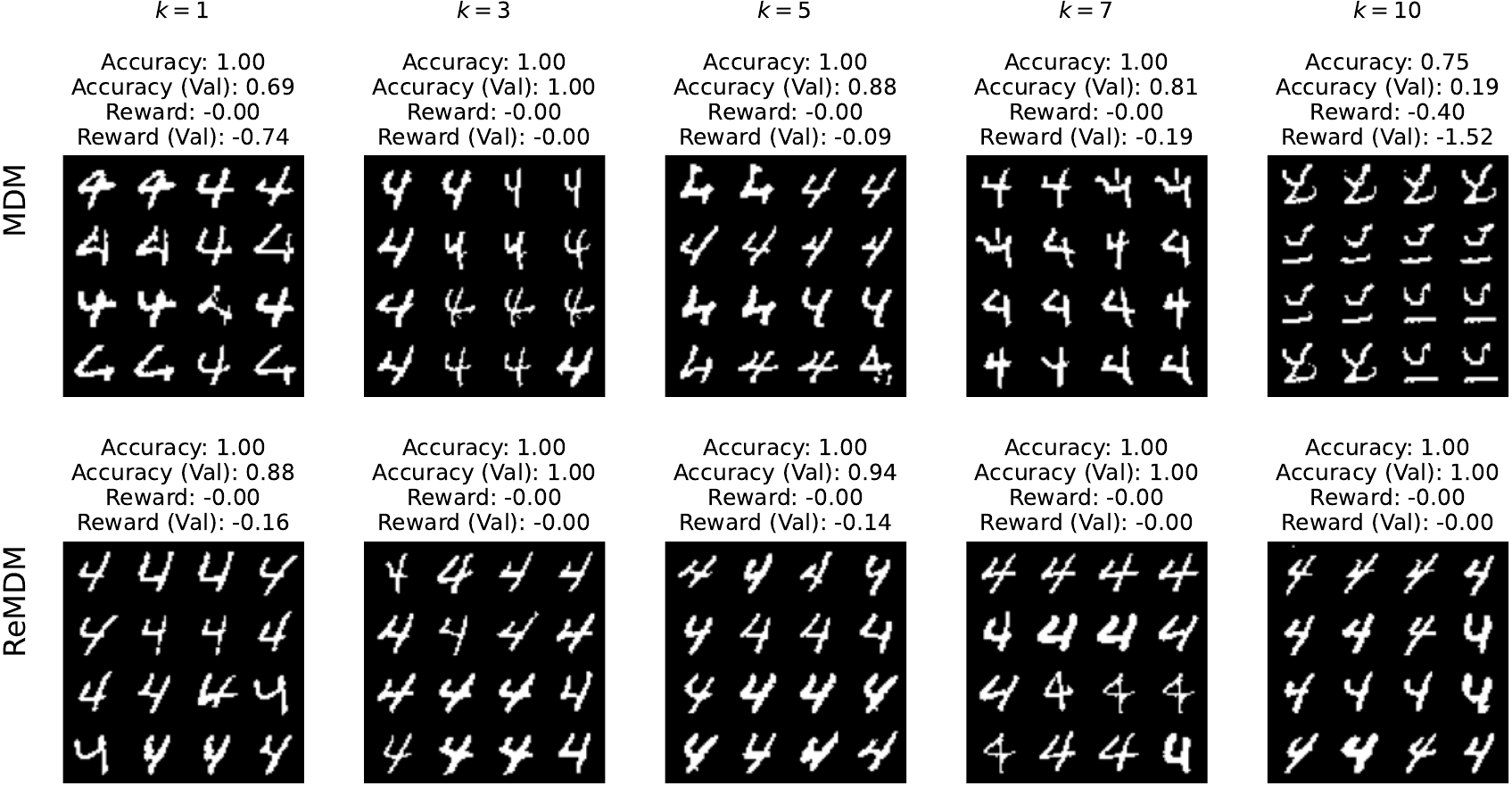}
    \caption{Samples from \oursgrad ($N=16$) for different $\lambda_t$ schedules; the run is selected based on lowest validation accuracy.}
    \label{fig:lambda_t_ablation_mnist_samples}
\end{figure}

\textbf{Additional Results with Different $\lambda_t$ Schedules.}
To investigate the effect of the $\lambda_t$ schedule on the performance of SMC, we define a family of linear schedules with different slopes (see \cref{fig:lambda_t_plot}) parametrised with $k$,
\begin{align*}
    \lambda_t(k) = \min \left(1, \frac{10}{k}\Big(1-\frac{t}{T}\Big)\right).
\end{align*}
In \cref{fig:smc_binarized_mnist_lambda_t_ablation}, we compare the results of \oursgrad ($N=16$) with different $\lambda_t$ schedules. The reward is given by $r(x) = \log p_{\text{clf}}(y=4|x)$ where $p_{\text{clf}}(y|x)$ is a classifier trained on the clean MNIST data. The accuracy is given as the fraction of final SMC samples (out of $N$) which are classified as the digit $4$ i.e., $p_{\text{clf}}(y=4|x) > 0.5$. For the validation reward and accuracy, we train another classifier with a slightly different neural architecture. The means and standard deviations are calculated using 30 independent SMC runs for each $k$. When using MDM sampling \citep{sahoo2024simple}, we observe the highest validation accuracy and the lowest variance at $k=3$; the average validation accuracy drops slightly, and variance increases for both larger and smaller values of $k$. We show the samples from the runs with the lowest validation accuracies for selected values of $k$ in \cref{fig:lambda_t_ablation_mnist_samples}. If $\lambda_t$ increases too slowly (large $k$), early unmasked pixels may resemble incorrect digits which cannot be corrected in MDM, leading to corrupted final samples despite high reward value. Conversely, increasing $\lambda_t$ too quickly (small $k$) there is a risk of weighting particles using a high variance reward estimate in early steps when most of the image is still masked. Finally, we observe that ReMDM sampling \citep{wang2025remasking} is much more resilient to different $\lambda_t$ schedules as can be seen from both \cref{fig:smc_binarized_mnist_lambda_t_ablation,fig:lambda_t_ablation_mnist_samples}.

\begin{wrapfigure}{r}{0.6\linewidth}
    \centering
    \vspace{-3mm}
    \includegraphics[width=\linewidth]{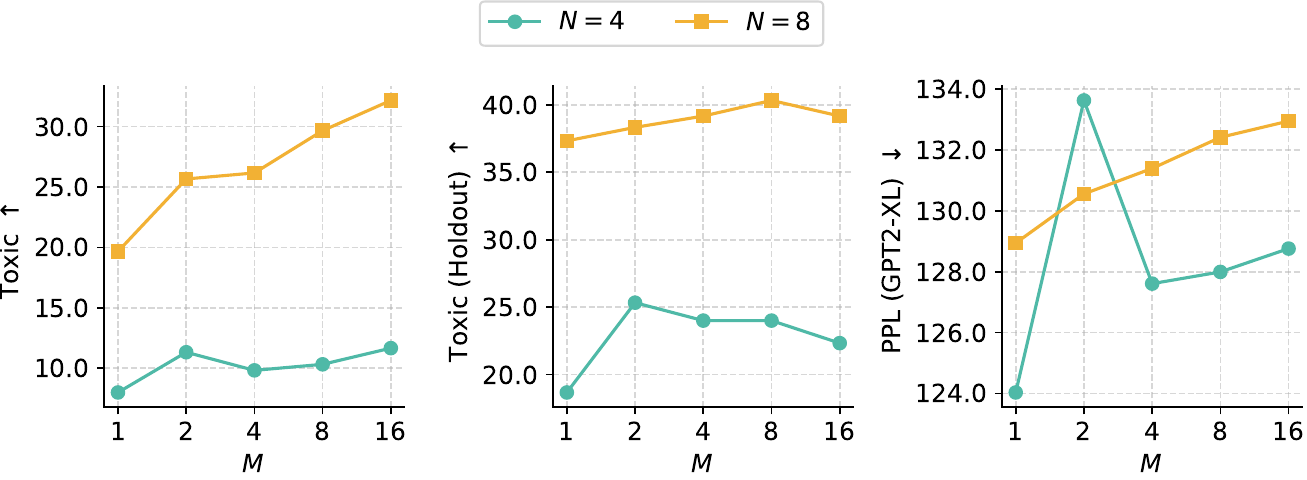}
    \vspace{-6mm}
    \caption{Comparing \oursbase with different values of $M$ for toxic text generation}
    \label{fig:smc-toxicity-phi-ablation}
    \vspace{-4mm}
\end{wrapfigure}

\textbf{Additional Results with Different $M$.}
In \cref{sec:smc-recipe}, we use $M$ samples to estimate the reward. In \cref{fig:smc-toxicity-phi-ablation}, we compare the results of \oursbase for toxic text generation with different values of $M$. We observe a clear increase in the toxicity metrics when $M$ is increased from $1$ to $2$. However, the performance gain from increasing $M$ sometimes saturates at higher values. This is expected, as the variance of the Monte Carlo reward estimator decreases rapidly at first but slows down as $M$ grows.

\begin{figure}[!t]
    \centering
    \begin{minipage}{0.55\linewidth}
        \centering
        \includegraphics[width=.9\linewidth]{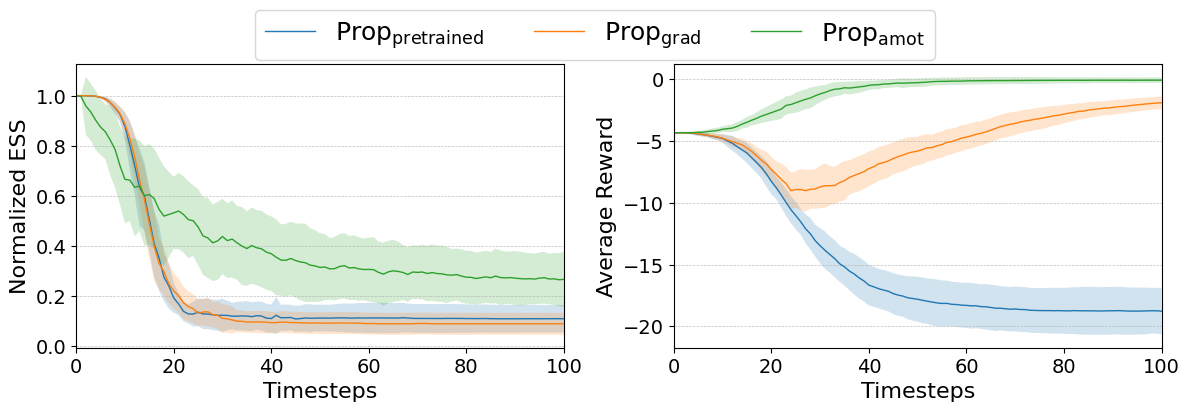}
        \vspace{-2mm}
        \caption{Normalised effective sample size (ESS) and reward of the particles across timesteps using different proposals without resampling.}
        \label{fig:ess_reward_diffprop}
        \vspace{-2mm}
    \end{minipage}
    \hfill
    \begin{minipage}{0.43\linewidth}
        \centering
        \vspace{-4mm}
        \includegraphics[width=1.\linewidth]{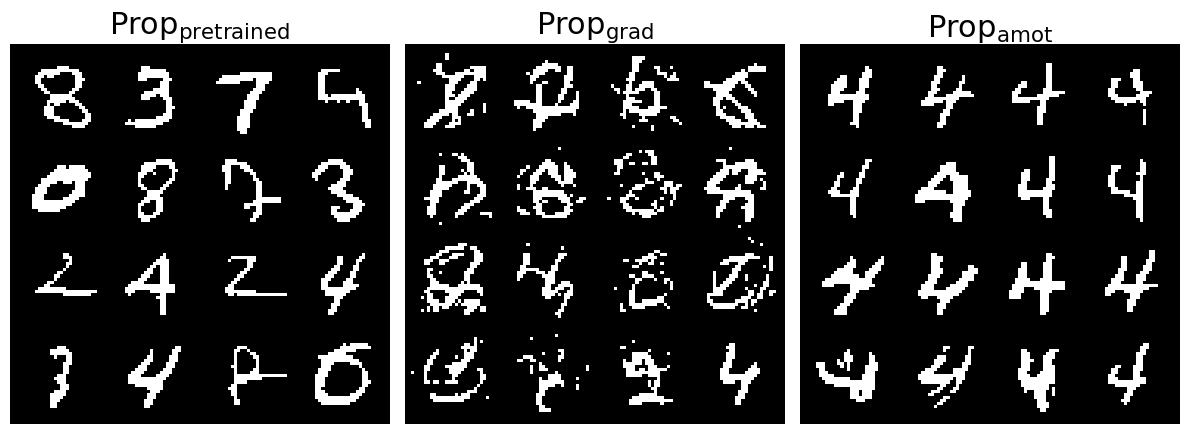}
        \vspace{-3.3mm}
        \caption{Generated samples using different proposals without resampling.}
        \label{fig:samples_mnist_diffprop}
        \vspace{-2mm}
    \end{minipage}
\end{figure}

\subsubsection{ESS and Reward Traces for Different Proposals}
Monitoring the effective sample size (ESS) provides a useful diagnostic of particle diversity over the SMC algorithm. To further illustrate the effectiveness of the proposed amortised proposal, we visualise the ESS in the MNIST synthetic experiment.

In this experiment, we evaluate three different proposals: \propbase, \propgrad, and \propamot, corresponding to the pretrained diffusion proposal, the first-order approximated proposal, and the learned amortised proposal, respectively. Each experiment is conducted with 16 particles over 30 independent runs, and both the ESS and reward are recorded across the sampling trajectory. Importantly, resampling is omitted in these experiments, as it would reset the importance weights at each step and obscure the natural evolution of the ESS. Omitting resampling allows ESS to serve as a clearer measure of each proposal’s intrinsic ability to maintain particle diversity.

The ESS and reward trajectories are shown in \cref{fig:ess_reward_diffprop}, alongside the generated samples illustrated in \cref{fig:samples_mnist_diffprop}. The results indicate that \propamot consistently achieves the highest ESS and reward during sampling, demonstrating the effectiveness of the log-variance minimisation objective in learning an approximately optimal proposal. In contrast, while \propgrad achieves higher reward values than \propbase, it exhibits lower ESS, which is expected given that the first-order approximated proposal introduces bias relative to the optimal proposal and is therefore more prone to reward hacking.
Regarding the generated samples, \propgrad fails to produce high-quality images, whereas \propamot consistently generates visually coherent and realistic samples. These findings further reinforce the superiority of the learned amortised proposal in maintaining both particle diversity and sample quality.

\subsubsection{Additional Results on Language Modelling}
We provide additional comparisons in \cref{tab:toxicity-results-expanded}, which extends the results in \cref{tab:toxicity-results}. The expanded table shows that increasing the number of particles consistently improves the performance of all SMC methods with respect to the toxicity metrics. Furthermore, employing a proposal distribution that more closely approximates the optimal proposal leads to further performance gains, highlighting the critical role of the proposal distribution in SMC.

We further compare our methods to SVDD \citep{li2024derivative}, which performs importance sampling at every unmasked step while aggressively maintaining only a single particle and using a pretrained diffusion model as its proposal distribution.
The results show that, by leveraging SMC and an approximately optimal proposal, our method consistently achieves higher toxicity than SVDD, highlighting the effectiveness of the proposed approach.

\begin{table}[!t]
\centering
\small
\caption{The results of toxic text generation (the expanded version of \cref{tab:toxicity-results}). }
\label{tab:toxicity-results-expanded}
\vspace{-2mm}
\resizebox{1.0\linewidth}{!}{
\begin{tabular}{l l c c c c}
\toprule
{\# Particles} & {Method} & {Toxic \(\uparrow\)} & {Toxic (Holdout) \(\uparrow\)} & {PPL (GPT2-XL) \(\downarrow\)} & {Dist-1/2/3 \(\uparrow\)} \\
\midrule

\multirow{3}{*}{N = 1} & Pretrained
&  0.8\% &  5.2\% & {121.1} & 56/92/96 \\
& \propgrad
& 58.0\% & 58.3\% & 216.7 & 58/93/96 \\
& \propamot
& 63.7\% & 75.7\% & 131.9 & 53/89/94 \\
\midrule
\multirow{5}{*}{N = 2} & BoN
&  1.7\% & 9.3\% & 129.1 & 57/92/96 \\
&\revision{SVDD}
& \revision{14.6\%} & \revision{27.0\%} & \revision{129.0}  & \revision{56/91/95} \\
&\oursbase
& 1.0\% & 5.3\% & 133.6  & 58/92/96 \\
&\oursgrad
& 74.3\% & 68.7\% & 199.9 & 58/92/96 \\
&\oursamot
& 84.7\% & 90.3\% & 140.6 & 51/88/94 \\
\midrule
\multirow{5}{*}{N = 4} & BoN
&  2.8\% & 13.3\% & 121.5 & 57/92/96 \\
&\revision{SVDD}
& \revision{65.7\%} & \revision{67.0\%} & \revision{129.1} & \revision{58/91/94} \\
&\oursbase
& 10.3\% & 26.3\% & 125.6 & 56/92/96 \\
&\oursgrad
& 85.0\% & 76.3\% & 137.8 & 57/92/96 \\
&\oursamot
& 98.3\% & 99.0\% & 127.0 & 43/81/91 \\
\midrule
\multirow{5}{*}{N = 8} & BoN
&  6.3\% & 16.7\% & 127.4 & 56/91/96 \\
&\revision{SVDD}
& \revision{92.2\%} & \revision{82.2\%} & \revision{121.9} & \revision{59/90/93} \\
&\oursbase
& 26.7\% & 40.0\% & 132.3 & 57/92/96 \\
&\oursgrad
& {95.0\%} & {86.3\%} & 132.1 & 57/92/96 \\
&\oursamot
& 100.0\% & 100.0\% & 127.0 & 43/81/91 \\
\midrule
\multirow{5}{*}{N = 16} & BoN
&  9.7\% & 24.3\% & 118.8 & 57/92/96 \\
&\revision{SVDD}
& \revision{97.5\%} & \revision{91.0\%} & \revision{127.7}  & \revision{58/89/93} \\
&\oursbase
& 52.3\% & 54.7\% & 117.0  & 57/92/95 \\
&\oursgrad
& 98.7\% & 88.0\% & 121.7 & 56/91/95 \\
&\oursamot
& 100.0\% & 100.0\% & 114.2 & 40/79/90 \\
\bottomrule
\end{tabular}
}
\end{table}

\subsubsection{Additional Results on Biology Design} \label{sec:appendix-comp-baselines}
We provide a comparison of our methods against baselines in \cref{table:dna_comp_baseline}. Compared to the pretrained model, \oursamot with a single particle achieves superior performance across all metrics, demonstrating the effectiveness of the learnable amortised proposal. 
Although \oursamot (N=1) underperforms DRAKES \citep{wang2024fine}, we find that increasing the number of particles substantially improves results: \oursamot attains better performance on \textit{Pred-Activity} and \textit{ATAC-Acc}, while achieving comparable performance on the remaining three metrics.
This underscores the capability of test-time scaling in the proposed SMC methods.

\revision{
We also compare with two additional baselines, SGDD \citep{chu2025split} and SVDD \citep{li2024derivative}, to better contextualise the behaviour of our SMC methods.
It is noteworthy that SGDD is a sampler restricted to uniform noising processes, while other methods in \cref{table:dna_comp_baseline} use masked diffusions.
Although SGDD attains higher predicted activity and ATAC accuracy, its substantially weaker performance on correlation indicates mode-collapse behaviour, suggesting that it overfits to a narrow region of sequence space and struggles to generate diverse samples.
SVDD shows a complementary pattern: while it attains strong predicted activity, its lower ATAC accuracy points to reward-hacking tendencies. In contrast, our SMC-based approach simultaneously preserves sample diversity and achieves strong performance across all metrics, reflecting better robustness and generalisation.
}

\subsubsection{Additional Results on Image Generation}
\cref{tab:fid_is_diff_particles_cfg1.5,tab:fid_is_diff_particles_cfg1.75} demonstrate additional results on enhancing CFG with the proposed SMC methods. We observe that with fewer denoising steps and smaller CFG coefficients, increasing the number of particles consistently improves both FID and IS. In contrast, when using more denoising steps and larger CFG coefficients, adding particles leads to higher IS but worse FID.
This behavior aligns with our expectations. Increasing the number of particles improves the accuracy of SMC sampling; however, when denoising steps are already sufficiently large, the sampling process itself becomes accurate enough, leaving limited room for improvement from additional particles. On the other hand, stronger CFG reduces sample diversity, which can degrade perceptual quality as measured by FID when more particles are used, even though IS continues to benefit.

We further compare different sampling schemes (see \cref{sec:appendix-diff-sampling-methods} for details) in \cref{tab:fid_is_diff_sampling_method_cfg1.5,tab:fid_is_diff_sampling_method_cfg1.75,tab:fid_is_diff_sampling_method_cfg1.25}.
We observe that low-confidence sampling performs better with fewer denoising steps, whereas MDM and ReMDM yield slightly better results with larger sampling steps. This provides evidence that the original low-confidence sampling in MaskGit \citep{chang2022maskgit} can be safely replaced by ReMDM, which additionally enables tractable importance weights for SMC.

\begin{table}[!t]
    \centering
    \caption{Model performance on DNA sequence design. We report the mean across 3 random seeds, with standard deviations in parentheses. The results of baselines are from \cite{wang2024fine}.}
    \label{table:dna_comp_baseline}
     \vspace{-2mm}
    \resizebox{1.0\linewidth}{!}{
        \begin{tabular}{l ccccc}
            \toprule
            Method & Pred-Activity (median)\,$\uparrow$ & ATAC-Acc\,$\uparrow$ (\%) & 3-mer Corr\,$\uparrow$ & JASPAR Corr\,$\uparrow$ & {App-Log-Lik} (median)\,$\uparrow$ \\
            \midrule
            Pretrained & 0.17(0.04) & 1.5(0.2) & -0.061(0.034) & 0.249(0.015) & -261(0.6) \\
            CG & 3.30(0.00) & 0.0(0.0) & -0.065(0.001) & 0.212(0.035) & -266(0.6) \\
            CFG & 5.04(0.06) & 92.1(0.9) & 0.746(0.001) & 0.864(0.011) & -265(0.6) \\
            $\text{DRAKES}_{\text{w/o KL}}$ & {6.44(0.04)} & 82.5(2.8) & 0.307(0.001) & 0.557(0.015) & -281(0.6) \\
            DRAKES & 5.61(0.07) & {92.5(0.6)} & {0.887(0.002)} & {0.911(0.002)} & -264(0.6) \\
            \revision{SGDD ($\beta$ = 30)} & \revision{8.85(0.07)} & \revision{90.9(0.00)} & \revision{0.470(0.014)} & \revision{0.466(0.015)} & \revision{-263(1.6)} \\
            \revision{SGDD ($\beta$ = 50)} & \revision{9.32(0.04)} & \revision{96.4(0.01)} & \revision{0.370(0.010)} & \revision{0.398(0.001)} & \revision{-269(0.1)} \\
            \revision{SVDD (N = 8)} & \revision{6.57(0.01)} & \revision{67.4(0.01)} & \revision{0.813(0.009)} & \revision{0.753(0.011)} & \revision{-258(0.2)} \\
            \revision{SVDD (N = 16)} & \revision{6.89(0.04)} & \revision{84.3(0.01)} & \revision{0.891(0.009)} & \revision{0.834(0.011)} & \revision{-260(0.2)} \\
            \midrule
            \oursamot (N = 1) & 5.40(0.02) & 82.1(0.01) & 0.653(0.001) & 0.778(0.005) & -259(0.1) \\
            \oursamot (N = 8) & 6.35(0.01) & 95.8(0.01) & 0.736(0.003) & 0.845(0.005) & -261(0.2) \\
            \oursamot (N = 16) & 6.68(0.02) & 97.6(0.01) & 0.796(0.005) & 0.886(0.002) & -261(0.4) \\
            \bottomrule
        \end{tabular}
    }
\end{table}

\begin{figure}
    \begin{minipage}{\linewidth}
        \centering
        \begin{minipage}{0.24\linewidth}
            \centering
            \includegraphics[width=.99\linewidth]{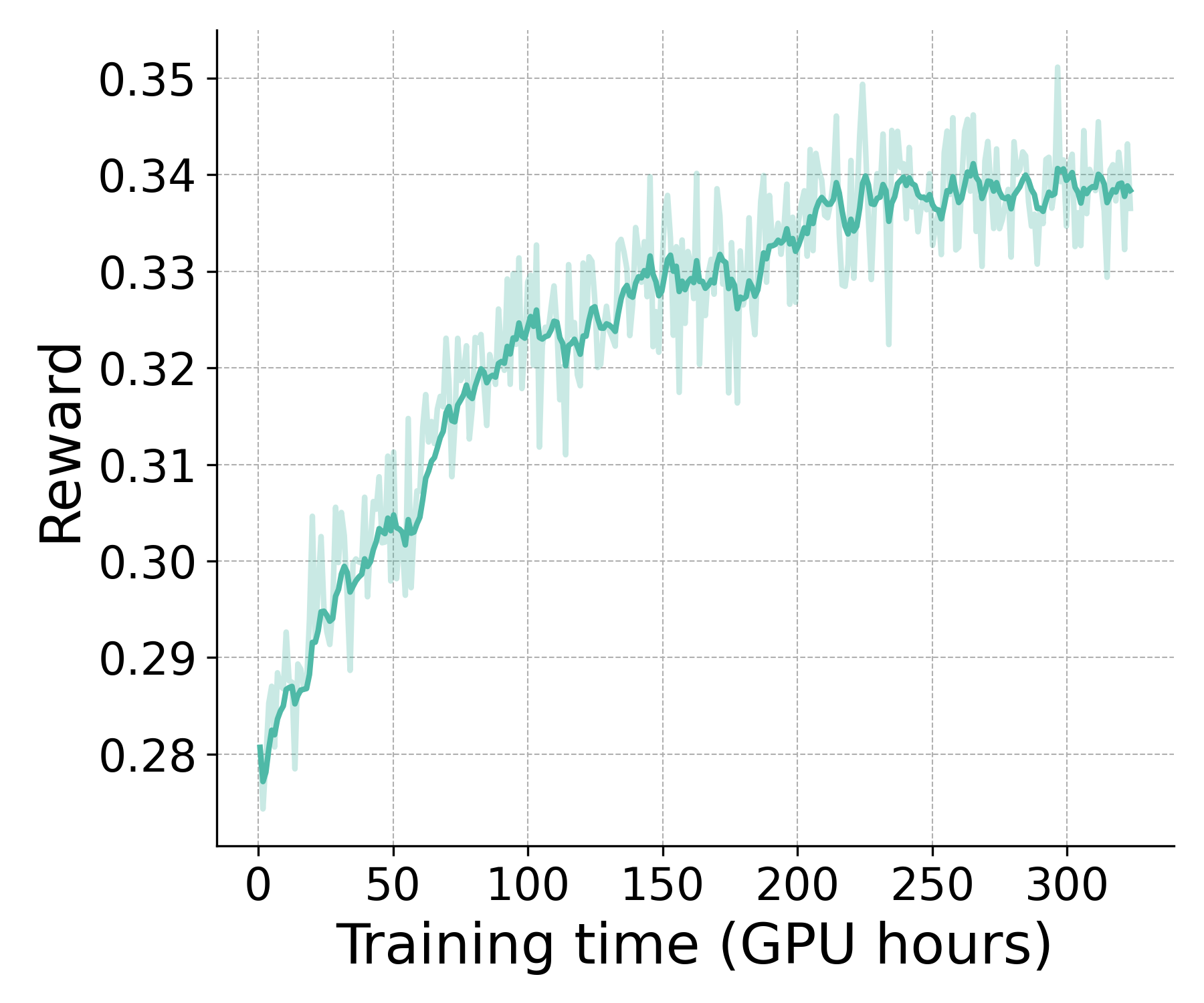}
            \subcaption{HPSv2}
        \end{minipage}
        \begin{minipage}{0.24\linewidth}
            \centering
            \includegraphics[width=.99\linewidth]{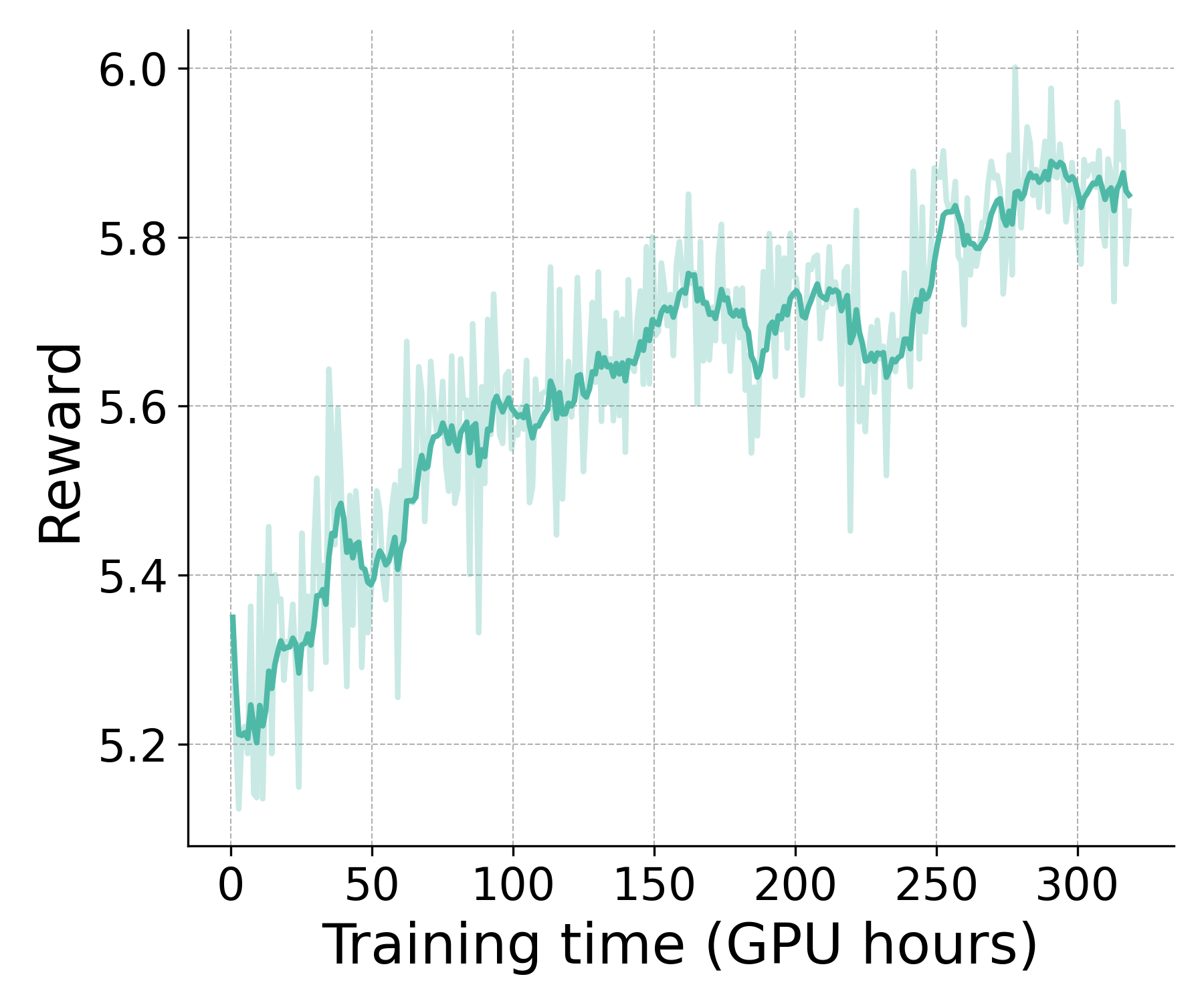}
            \subcaption{Aesthetic Score}
        \end{minipage}
        \begin{minipage}{0.24\linewidth}
            \centering
            \includegraphics[width=.99\linewidth]{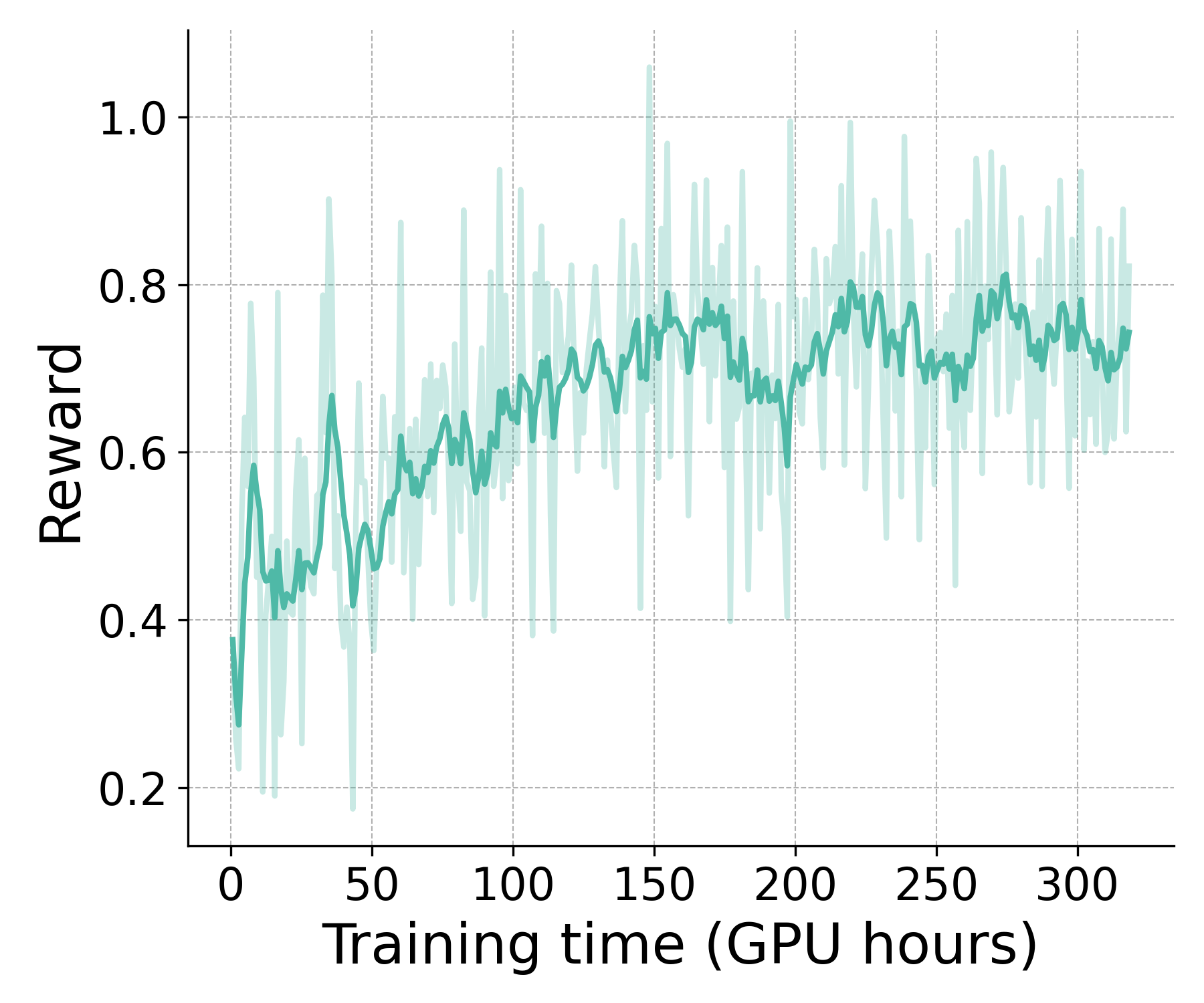}
            \subcaption{ImageReward}
        \end{minipage}
    \end{minipage}
    \vspace{-2mm}
    \caption{\revision{Illustration of the training cost: training time (GPU hours) against the reward.}}
    \label{fig:training_time_rewards}
    \vspace{-2mm}
\end{figure}

\subsubsection{\revision{Computational Cost of Text-to-Image Generation}}

\begin{wraptable}{r}{0.5\linewidth}
    \small
    \centering
    \caption{\revision{Comparisons of inference time cost on the text-to-image generation.}}
    \label{tab:inference_time_cost_t2v}
    \vspace{-2.5mm}
    \resizebox{1.0\linewidth}{!}{
    \revision{
    \begin{tabular}{lrrrrr}
        \toprule
        \# particles & 1 & 2 & 4 & 8 & 16 \\
        \midrule
        BoN (s) & 3.91 & 7.10 & 13.13 & 24.77 & 48.47 \\
        $\text{SMC}_\text{base/amot}$ (s) & - & 12.20 & 21.69 & 41.44 & 80.61 \\
        \oursgrad (s) & 16.19 & 26.70 & 47.98 & 95.70 & 181.26 \\
    \bottomrule
    \end{tabular}
    }}
    \vspace{-3mm}
\end{wraptable}
\revision{
To provide a clear picture of the computational cost of the text-to-image experiment, we plot the training time (GPU hours) against reward in \cref{fig:training_time_rewards}.
It can be seen that fine-tuning the Meissonic model requires approximately 300 GPU hours using 8 GPUs, which corresponds to roughly 1.5 days of wall-clock time.
For the inference cost, we summarise the wall-clock time in \cref{tab:inference_time_cost_t2v}, where we measure the time required to sample an image from a single prompt using different numbers of particles. The results show that \oursbase and \oursamot methods are more expensive than BoN. This is expected since SMC must evaluate the reward function repeatedly along the sampling trajectory, whereas BoN evaluates it only once. Moreover, \oursgrad is the most computationally costly, as it requires computing the gradient of the reward at each step.
}

\begin{table}[!t]
\centering
    \begin{minipage}[t]{0.49\linewidth}
        \caption{Comparisons of different numbers of particles with CFG=1.5 on ImageNet256.}
        \label{tab:fid_is_diff_particles_cfg1.5}
        \centering
        \setlength{\tabcolsep}{4.65pt}
        \vspace{-2.5mm}
        \resizebox{1.0\linewidth}{!}{
        \begin{tabular}{lrrrrrr}
            \toprule
            & \multicolumn{3}{c}{FiD $\downarrow$} & \multicolumn{3}{c}{IS $\uparrow$} \\
            \cmidrule(lr){2-4} \cmidrule(lr){5-7} 
            \# steps & 8 & 16 & 32 & 8 & 16 & 32 \\
            \midrule
            N = 1 & 15.67 & 9.67 & 8.57 & 97.6 & 135.2 & 155.5 \\
            N = 2 & 13.01 & 8.57 & 8.20 & 116.6 & 160.1 & 181.4 \\
            N = 4 & 11.00 & 8.35 & 8.75 & 138.5 & 186.6 & 207.3 \\
            N = 8 & 9.98 & 8.51 & 9.13 & 152.7 & 202.6 & 222.0 \\
            N = 16 & 9.74 & 8.86 & 9.70 & 166.3 & 216.5 & 233.8 \\
        \bottomrule
        \end{tabular}
        }
    \end{minipage}
    \begin{minipage}[t]{0.49\linewidth}
        \caption{Comparisons of different numbers of particles with CFG=1.75 on ImageNet256.}
        \label{tab:fid_is_diff_particles_cfg1.75}
        \centering
        \setlength{\tabcolsep}{4.pt}
        \vspace{-2.5mm}
        \resizebox{1.0\linewidth}{!}{
        \begin{tabular}{lrrrrrr}
            \toprule
            & \multicolumn{3}{c}{FiD $\downarrow$} & \multicolumn{3}{c}{IS $\uparrow$} \\
            \cmidrule(lr){2-4} \cmidrule(lr){5-7} 
            \# steps & 8 & 16 & 32 & 8 & 16 & 32 \\
            \midrule
            N = 1 & 11.04 & 7.94 & 8.12 & 133.4 & 178.2 & 194.5 \\
            N = 2 & 9.36 & 7.88 & 8.40 & 159.9 & 204.0 & 225.4 \\
            N = 4 & 9.05 & 8.81 & 9.88 & 180.0 & 229.4 & 247.7 \\
            N = 8 & 8.84 & 9.66 & 10.88 & 197.3 & 243.0 & 260.1 \\
            N = 16 & 9.26 & 10.30 & 11.59 & 206.6 & 254.1 & 271.6 \\
        \bottomrule
        \end{tabular}
        }
    \end{minipage}
    \begin{minipage}[t]{0.48\linewidth}
        \vspace{1mm}
        \caption{Comparisons of different sampling methods with CFG=1.5 on ImageNet256.}
        \label{tab:fid_is_diff_sampling_method_cfg1.5}
        \centering
        \vspace{-2.5mm}
        \resizebox{1.0\linewidth}{!}{
        \begin{tabular}{lrrrrrr}
            \toprule
            & \multicolumn{3}{c}{FiD $\downarrow$} & \multicolumn{3}{c}{IS $\uparrow$} \\
            \cmidrule(lr){2-4} \cmidrule(lr){5-7} 
            \# steps & 8 & 16 & 32 & 8 & 16 & 32 \\
            \midrule
            Confident & 12.87 & 9.47 & 10.48 & 110.0 & 147.6 & 153.0 \\
            MDM & 15.58 & 9.98 & 9.07 & 97.6 & 130.5 & 146.7 \\
            ReMDM & 15.67 & 9.67 & 8.57 & 97.6 & 135.2 & 155.5 \\
        \bottomrule
        \end{tabular}
        }
    \end{minipage}
    \begin{minipage}[t]{0.48\linewidth}
        \vspace{1mm}
        \caption{Comparisons of different sampling methods with CFG=1.75 on ImageNet256.}
        \label{tab:fid_is_diff_sampling_method_cfg1.75}
        \centering
        \vspace{-2.5mm}
        \resizebox{1.0\linewidth}{!}{
        \begin{tabular}{lrrrrrr}
            \toprule
            & \multicolumn{3}{c}{FiD $\downarrow$} & \multicolumn{3}{c}{IS $\uparrow$} \\
            \cmidrule(lr){2-4} \cmidrule(lr){5-7} 
            \# steps & 8 & 16 & 32 & 8 & 16 & 32 \\
            \midrule
            Confident & 9.74 & 8.85 & 10.48 & 146.2 & 185.3 & 153.0 \\
            MDM & 10.98 & 8.05 & 8.13 & 134.4 & 171.9 & 188.2 \\
            ReMDM & 11.04 & 7.94 & 8.12 & 133.4 & 178.2 & 194.5 \\
        \bottomrule
        \end{tabular}
        }
    \end{minipage}
    \begin{minipage}[t]{0.48\linewidth}
        \vspace{1mm}
        \caption{Comparisons of different sampling methods with CFG=1.25 on ImageNet256.}
        \label{tab:fid_is_diff_sampling_method_cfg1.25}
        \centering
        \vspace{-2.5mm}
        \resizebox{1.0\linewidth}{!}{
        \begin{tabular}{lrrrrrr}
            \toprule
            & \multicolumn{3}{c}{FiD $\downarrow$} & \multicolumn{3}{c}{IS $\uparrow$} \\
            \cmidrule(lr){2-4} \cmidrule(lr){5-7} 
            \# steps & 8 & 16 & 32 & 8 & 16 & 32 \\
            \midrule
            Confident & 19.50 & 12.59 & 12.95 & 74.5 & 104.1 & 108.4 \\
            MDM & 24.05 & 15.11 & 12.68 & 63.64 & 88.7 & 100.9 \\
            ReMDM & 24.64 & 14.94 & 12.02 & 62.8 & 90.7 & 107.5 \\
        \bottomrule
        \end{tabular}
        }
    \end{minipage}
\vspace{-3mm}
\end{table}

\subsubsection{More Qualitative Results with Generated Samples} \label{sec:appendix-illustrate-samples}
\begin{figure}[!t]
    \begin{minipage}{1.\linewidth}
        \centering
        \begin{minipage}{.16\linewidth}\centering Pretrained \end{minipage}
        \begin{minipage}{.16\linewidth}\centering $\text{Prop}_{\text{grad}}$ \end{minipage}
        \begin{minipage}{.16\linewidth}\centering $\text{Prop}_{\text{amot}}$ \end{minipage}
        \begin{minipage}{.16\linewidth}\centering \oursbase \end{minipage}
        \begin{minipage}{.16\linewidth}\centering \oursgrad \end{minipage}
        \begin{minipage}{.16\linewidth}\centering \oursamot \end{minipage}
    \end{minipage}
    \begin{minipage}{1.\linewidth}
        \centering
        \vspace{1.mm}
        {\scriptsize A cat in the style of Van Gogh's Starry Night.}
        \includegraphics[width=1.\linewidth]{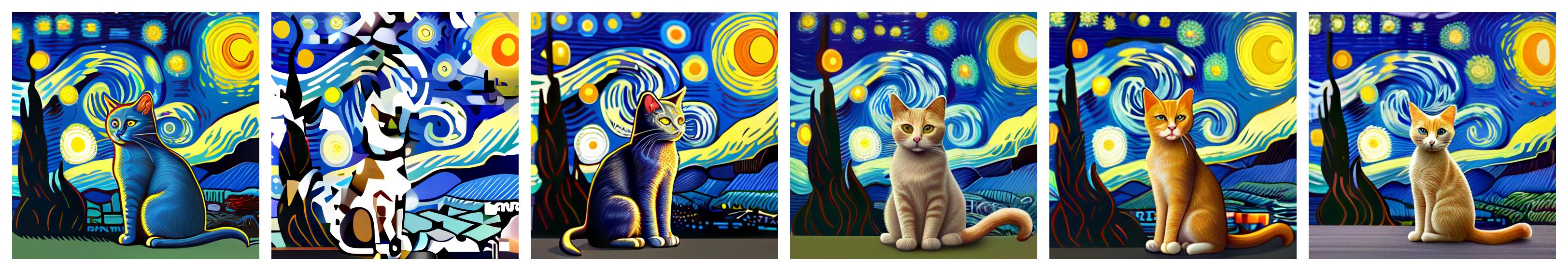}
    \end{minipage}
    \begin{minipage}{1.\linewidth}
        \centering
        {\scriptsize A photo of a brown knife and a blue donut.}
        \includegraphics[width=1.\linewidth]{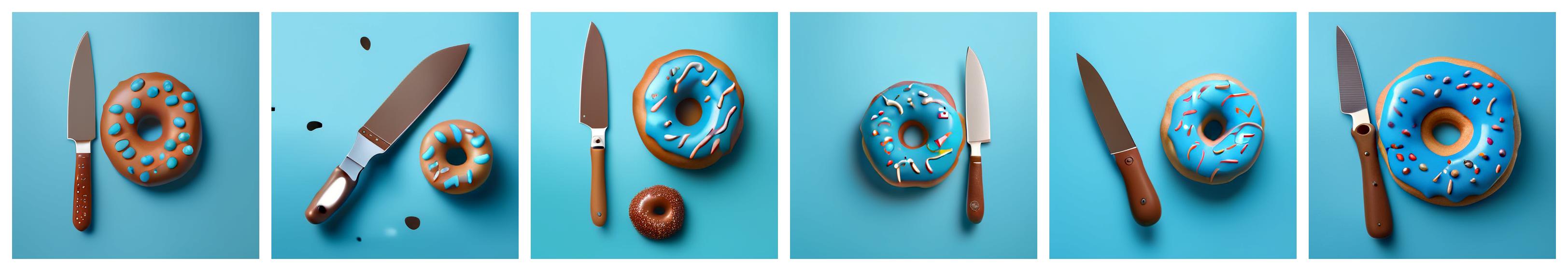}
    \end{minipage}
    \begin{minipage}{1.\linewidth}
        \centering
        {\scriptsize A photo of a yellow bird and a black motorcycle.}
        \includegraphics[width=1.\linewidth]{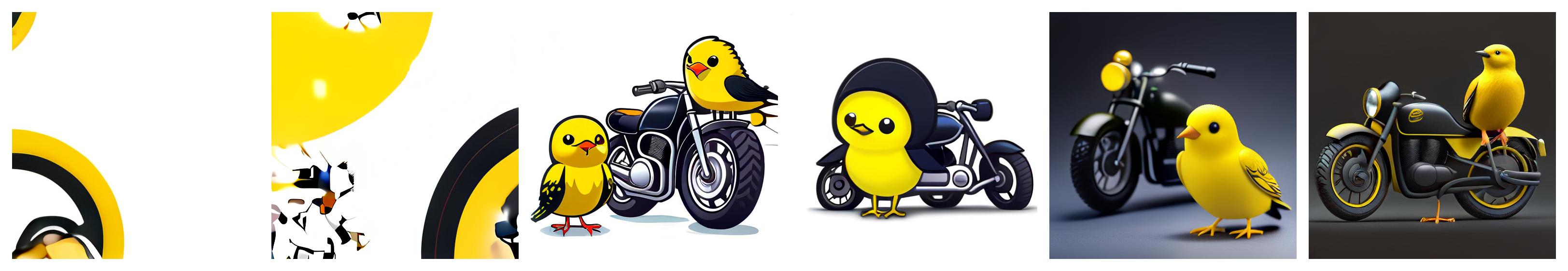}
    \end{minipage}
    \begin{minipage}{1.\linewidth}
        \centering
        {\scriptsize A cat and a dog.}
        \includegraphics[width=1.\linewidth]{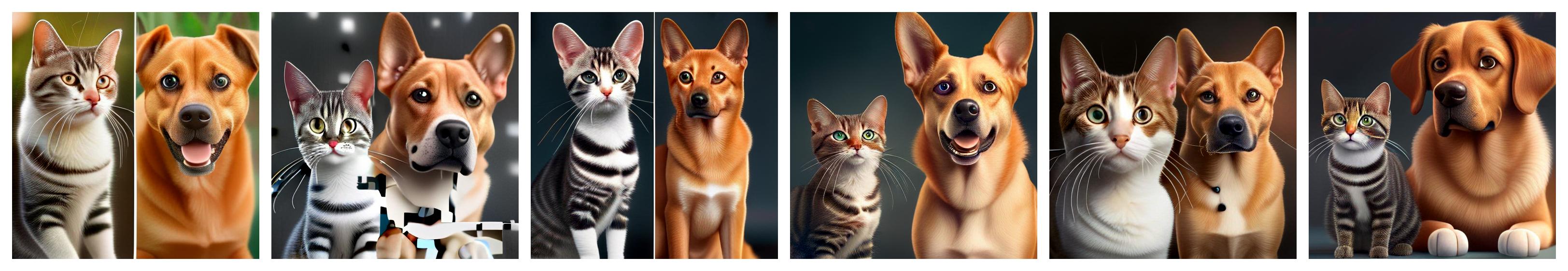}
    \end{minipage}
    \vspace{-2mm}
    \caption{Visualised comparison of different methods on ImageReward.}
    \label{fig:t2i-imagereward-compsmc}
\end{figure}

\begin{figure}[!t]
    \centering
    \begin{minipage}{.95\linewidth}
        \centering
        \begin{minipage}{.32\linewidth}\centering Pretrained \end{minipage}
        \hfill
        \begin{minipage}{.32\linewidth}\centering $\text{Prop}_{\text{amot}}$ \end{minipage}
        \hfill
        \begin{minipage}{.32\linewidth}\centering \oursamot \end{minipage}
    \end{minipage}
    \begin{minipage}{0.95\linewidth}
        \centering
        \begin{minipage}{1.\linewidth}
            \centering
            {\scriptsize A broken videogame console with a colorful and compelling painting.}
        \end{minipage}
        \begin{minipage}{0.32\linewidth}
            \centering
            \includegraphics[width=1.\linewidth]{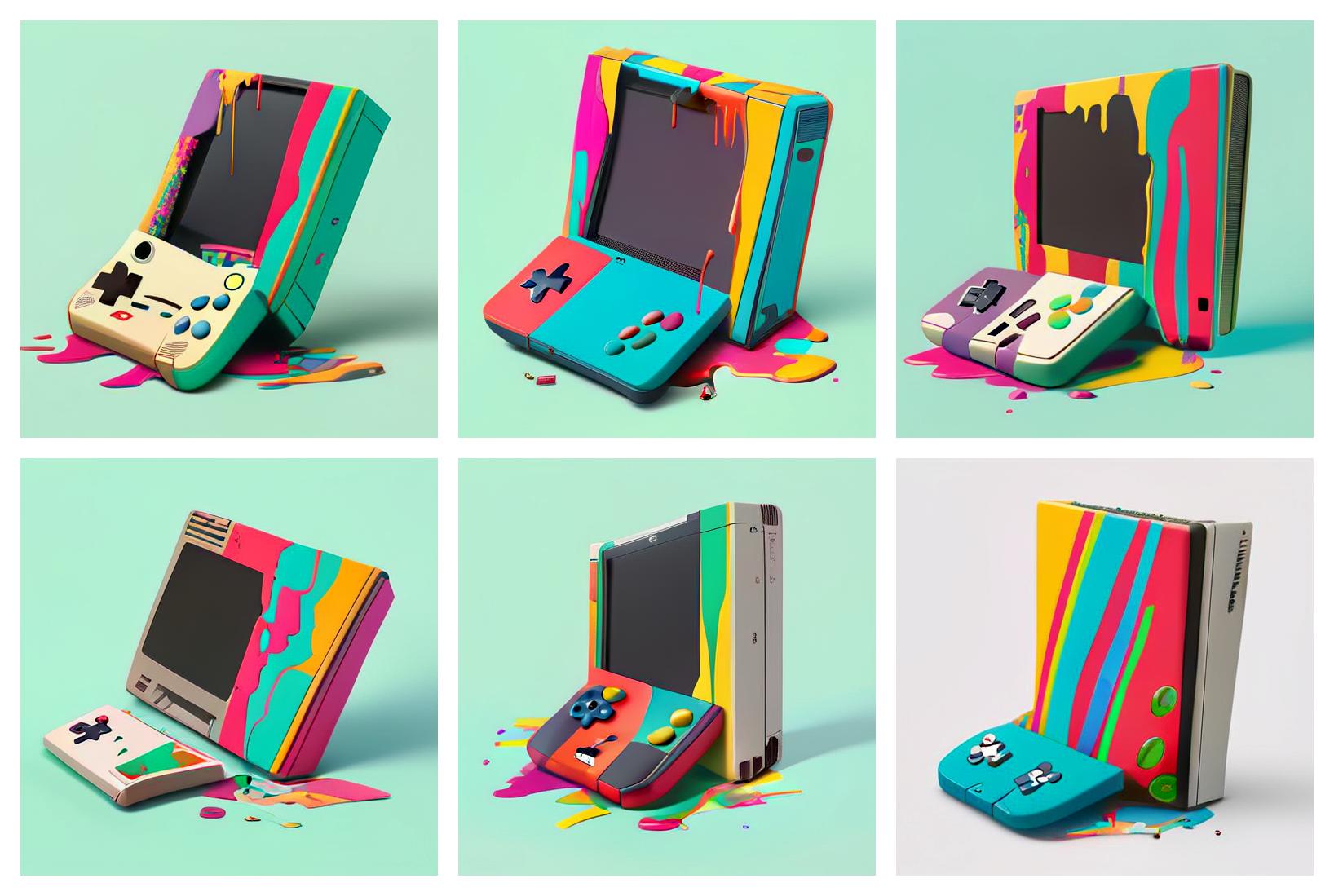}
        \end{minipage}
        \hfill
        \begin{minipage}{0.32\linewidth}
            \centering
            \includegraphics[width=1.\linewidth]{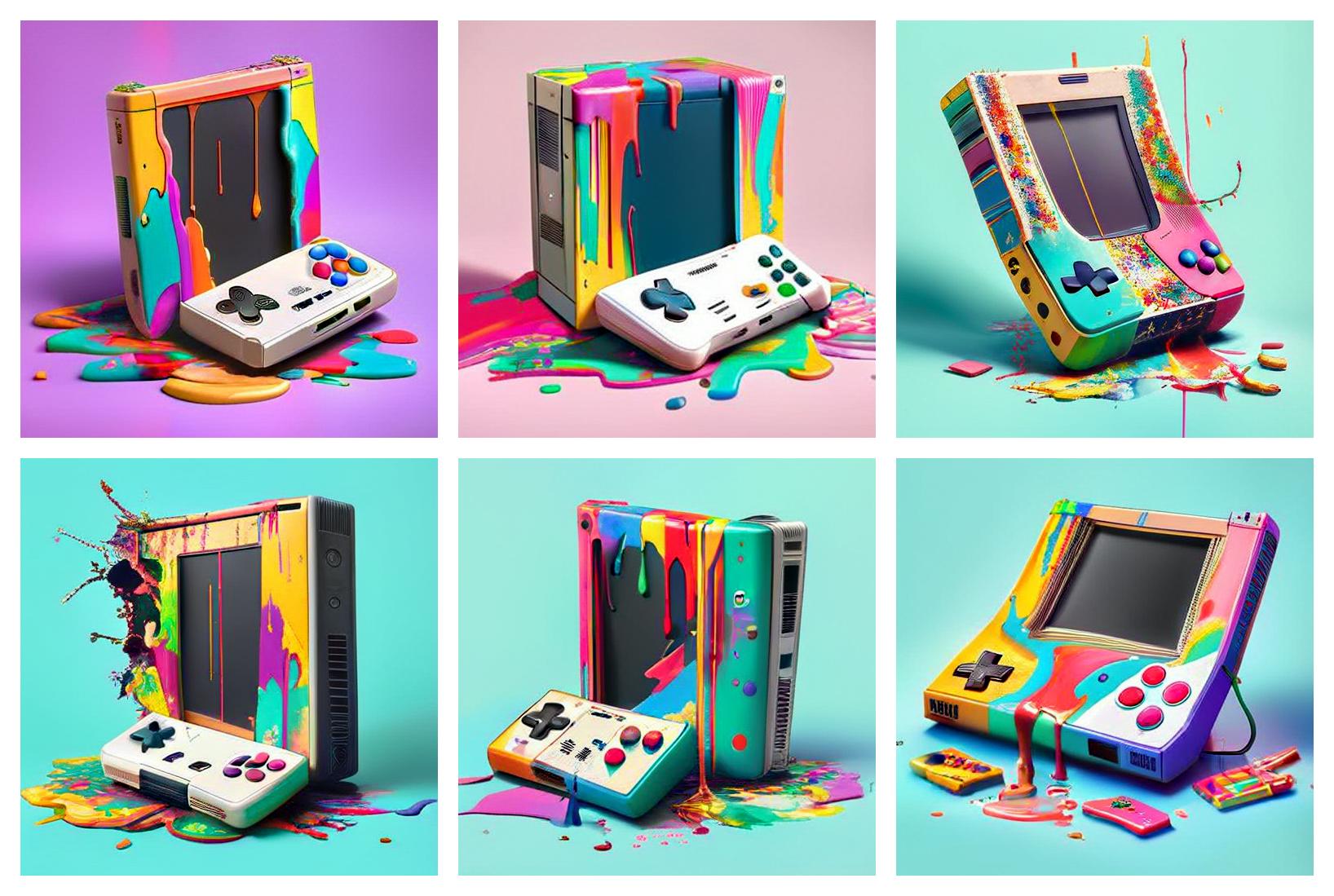}
        \end{minipage}
        \hfill
        \begin{minipage}{0.32\linewidth}
            \centering
            \includegraphics[width=1.\linewidth]{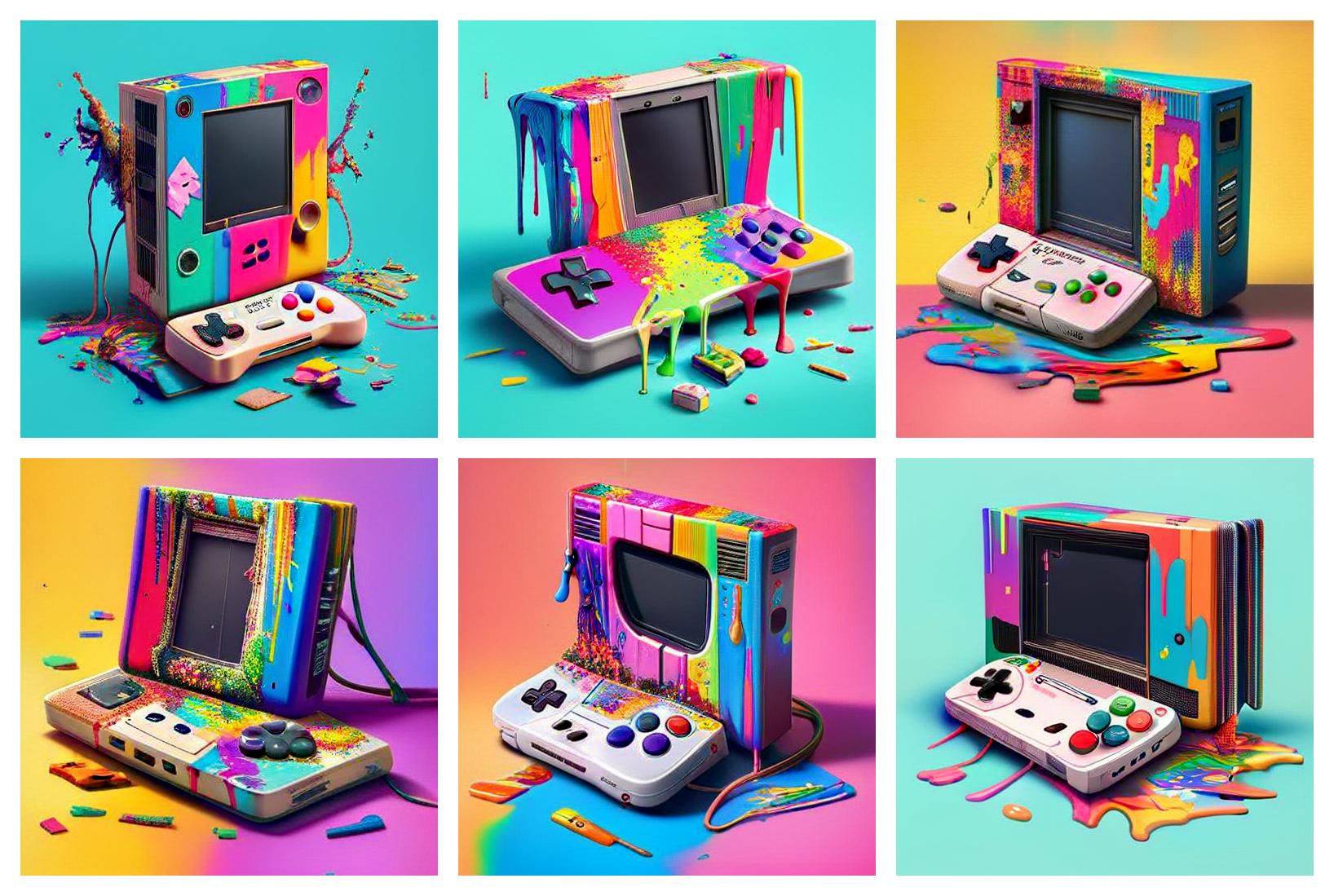}
        \end{minipage}
    \end{minipage}
    \begin{minipage}{0.95\linewidth}
        \centering
        \begin{minipage}{1.\linewidth}
            \centering
            {\scriptsize Anthropomorphic Virginia opossum playing guitar.}
        \end{minipage}
        \begin{minipage}{0.32\linewidth}
            \centering
            \includegraphics[width=1.\linewidth]{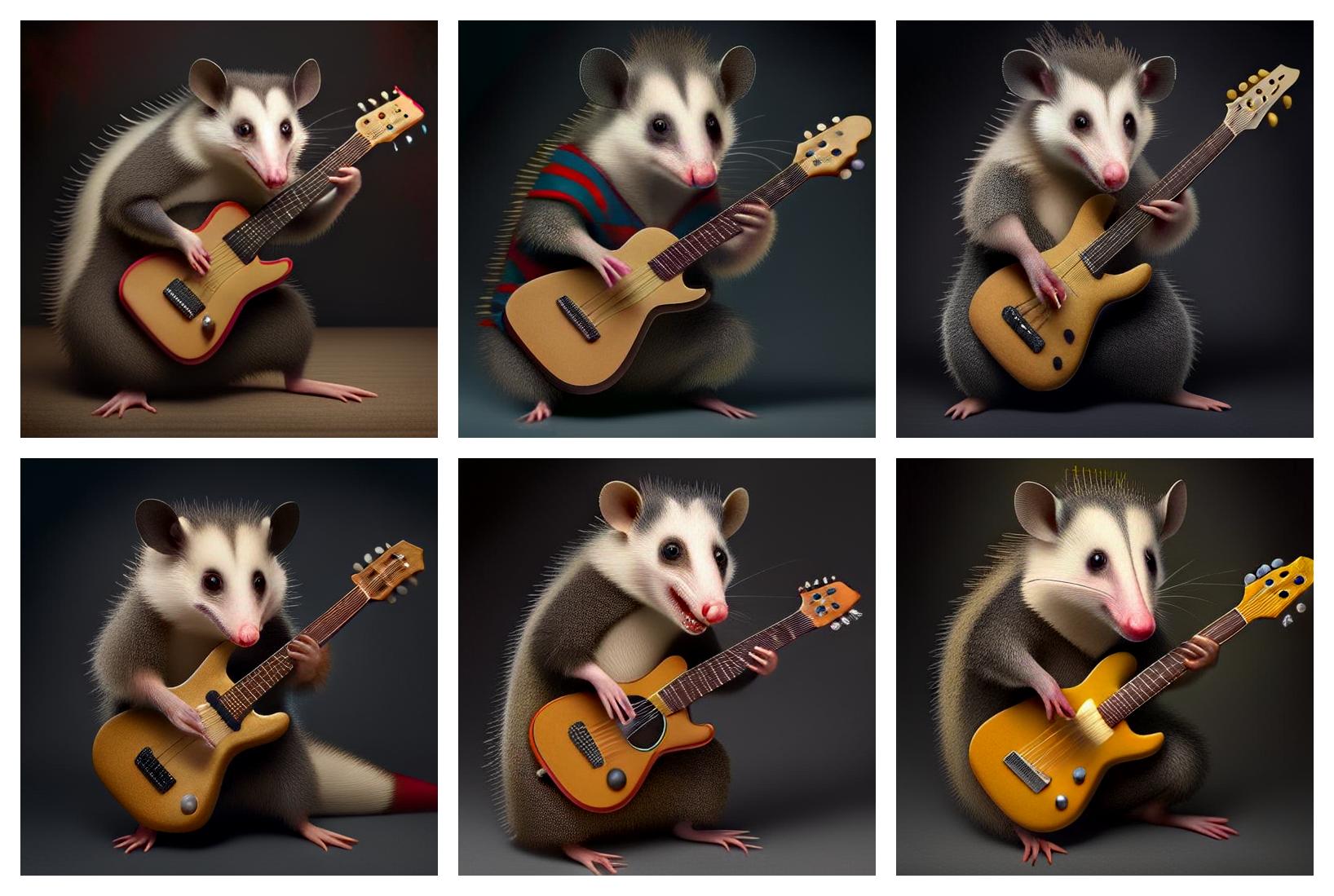}
        \end{minipage}
        \hfill
        \begin{minipage}{0.32\linewidth}
            \centering
            \includegraphics[width=1.\linewidth]{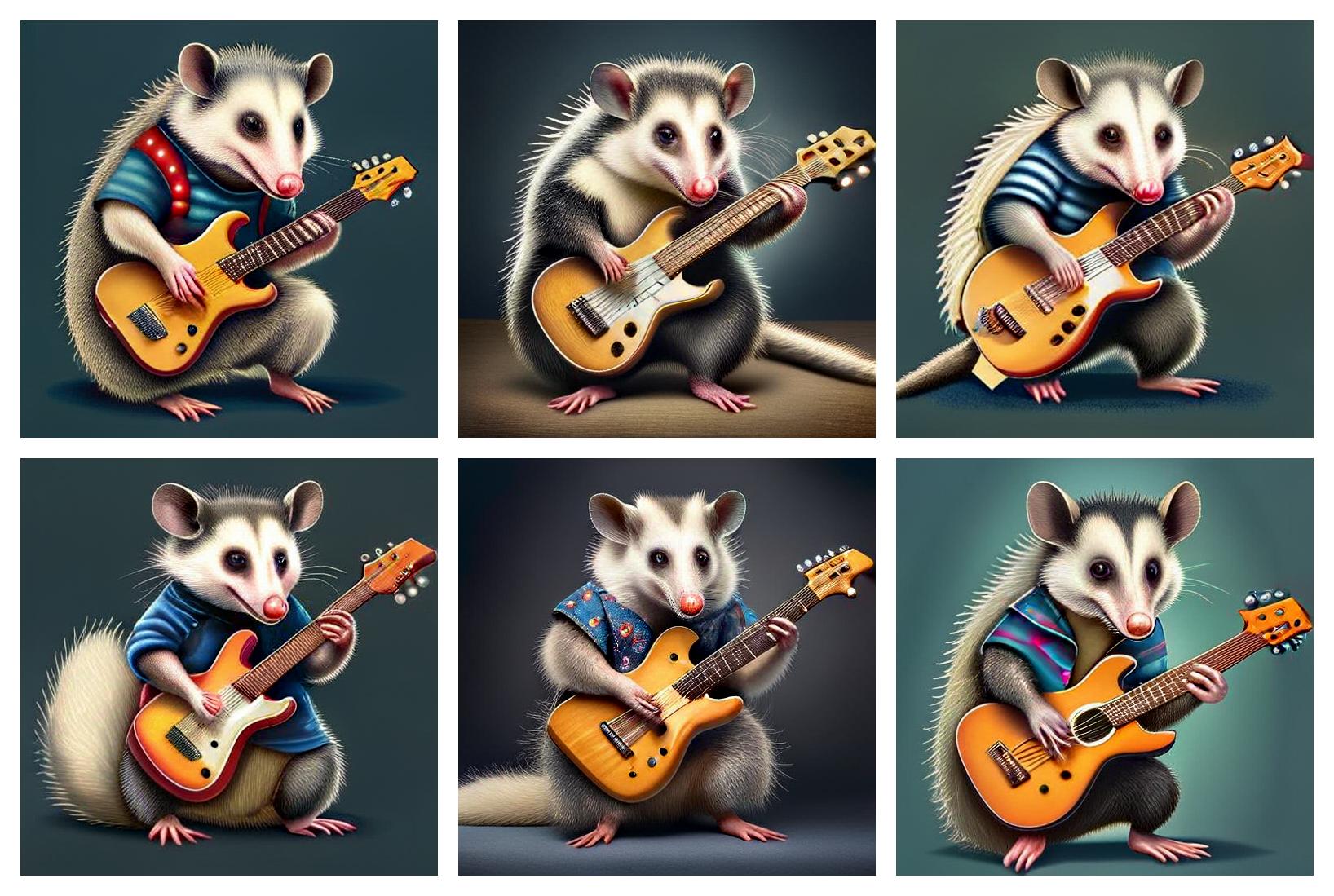}
        \end{minipage}
        \hfill
        \begin{minipage}{0.32\linewidth}
            \centering
            \includegraphics[width=1.\linewidth]{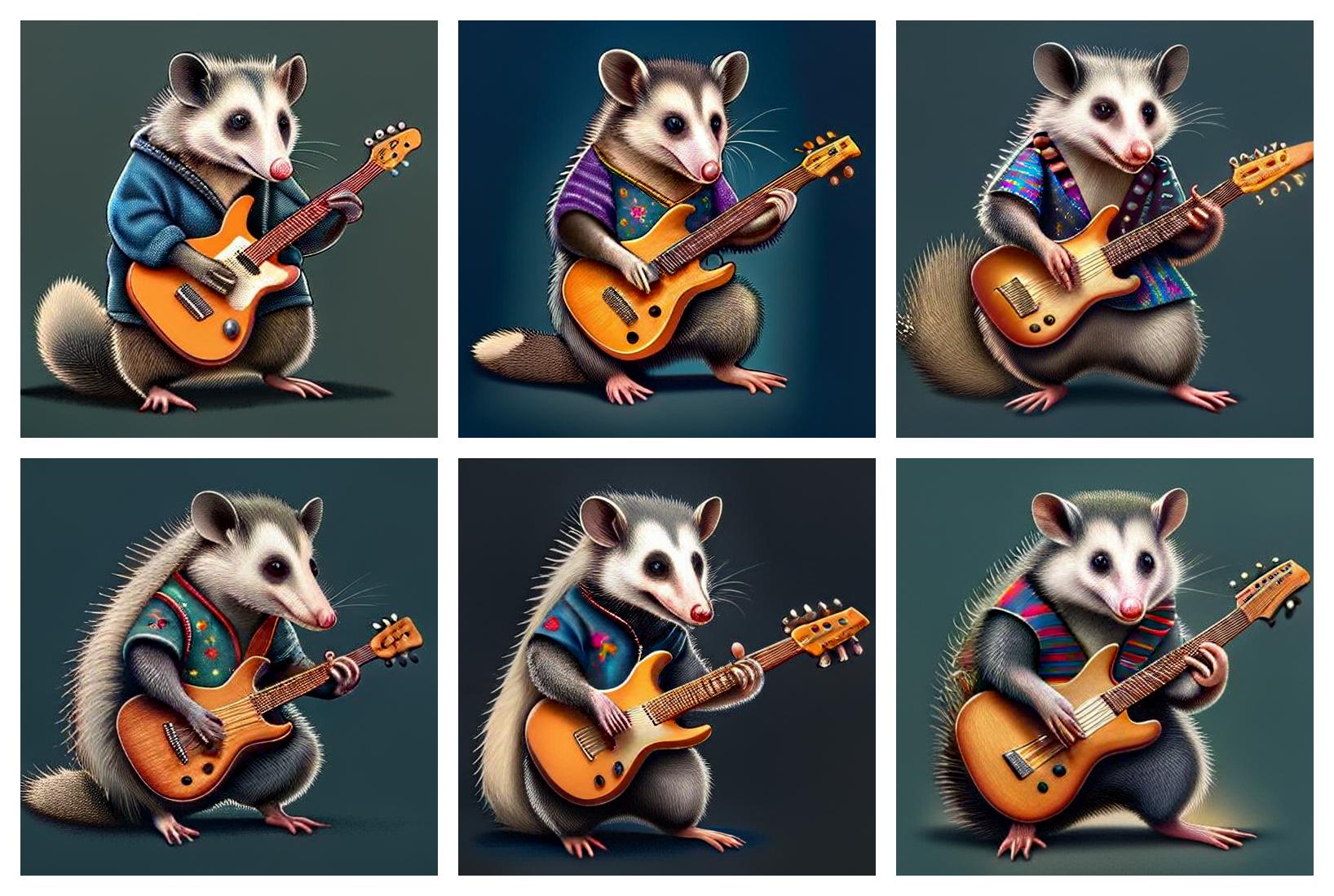}
        \end{minipage}
    \end{minipage}
    \begin{minipage}{0.95\linewidth}
        \centering
        \begin{minipage}{1.\linewidth}
            \centering
            {\scriptsize A painting of a Bladerunner interior room in Africa with detailed artwork.}
        \end{minipage}
        \begin{minipage}{0.32\linewidth}
            \centering
            \includegraphics[width=1.\linewidth]{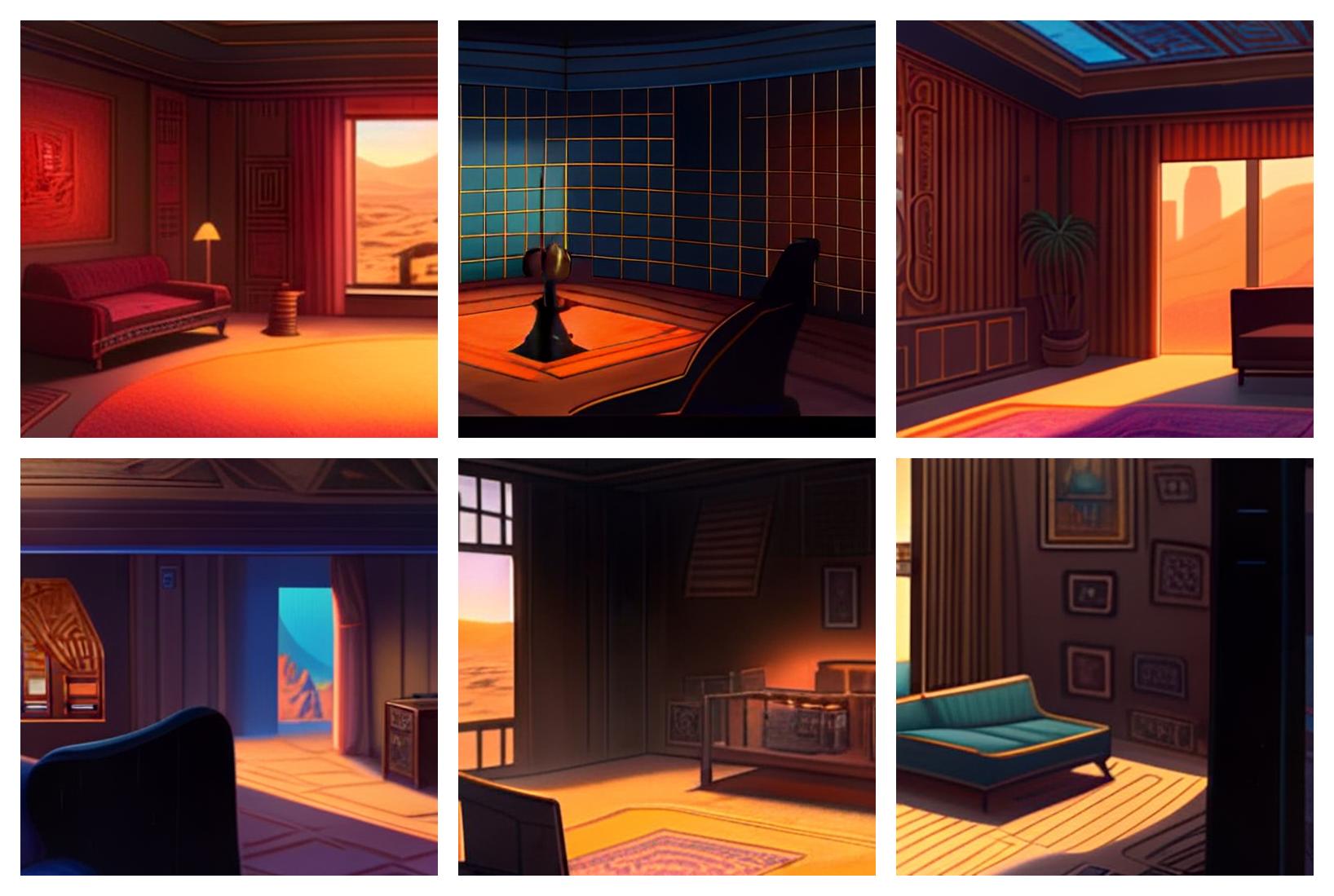}
        \end{minipage}
        \hfill
        \begin{minipage}{0.32\linewidth}
            \centering
            \includegraphics[width=1.\linewidth]{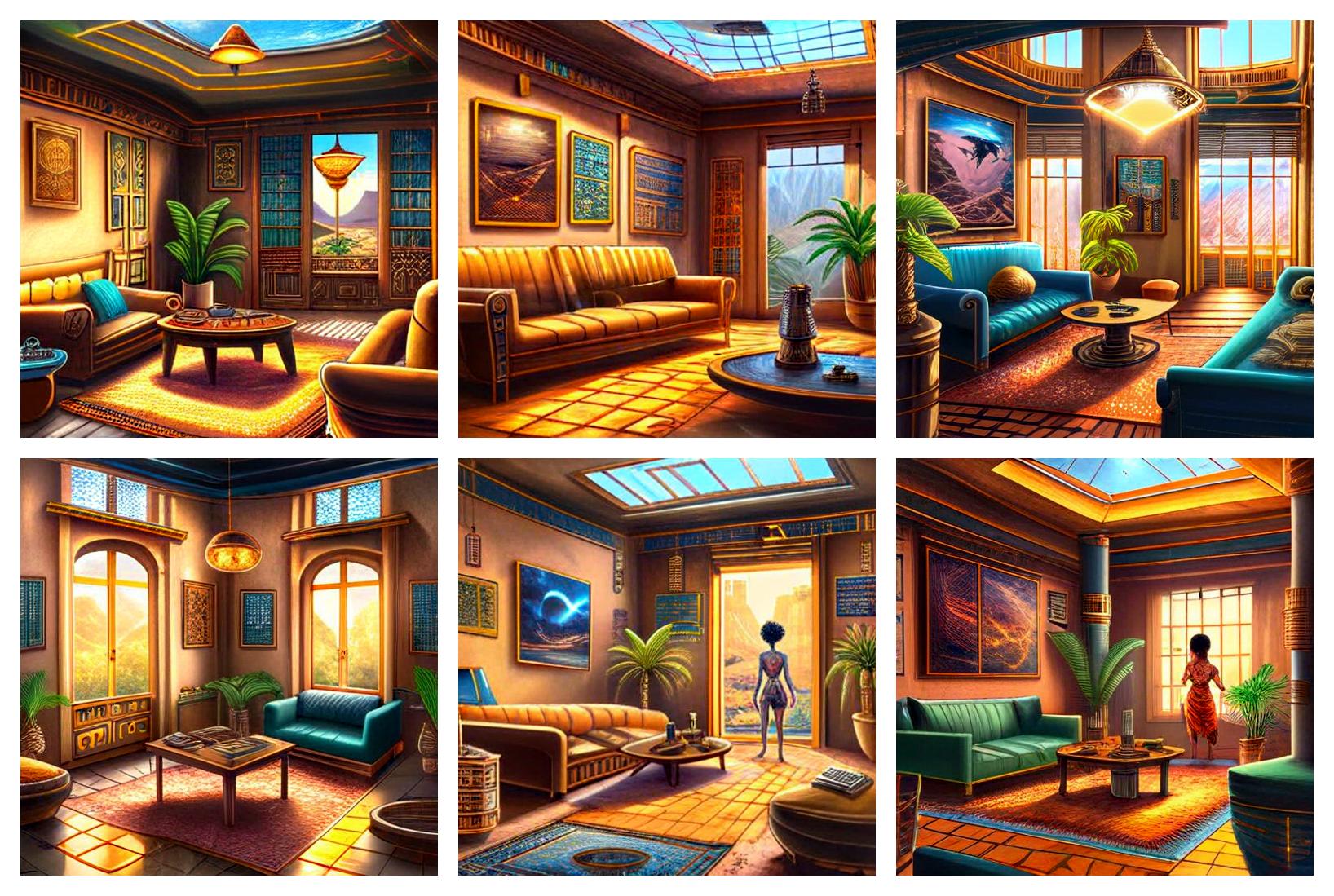}
        \end{minipage}
        \hfill
        \begin{minipage}{0.32\linewidth}
            \centering
            \includegraphics[width=1.\linewidth]{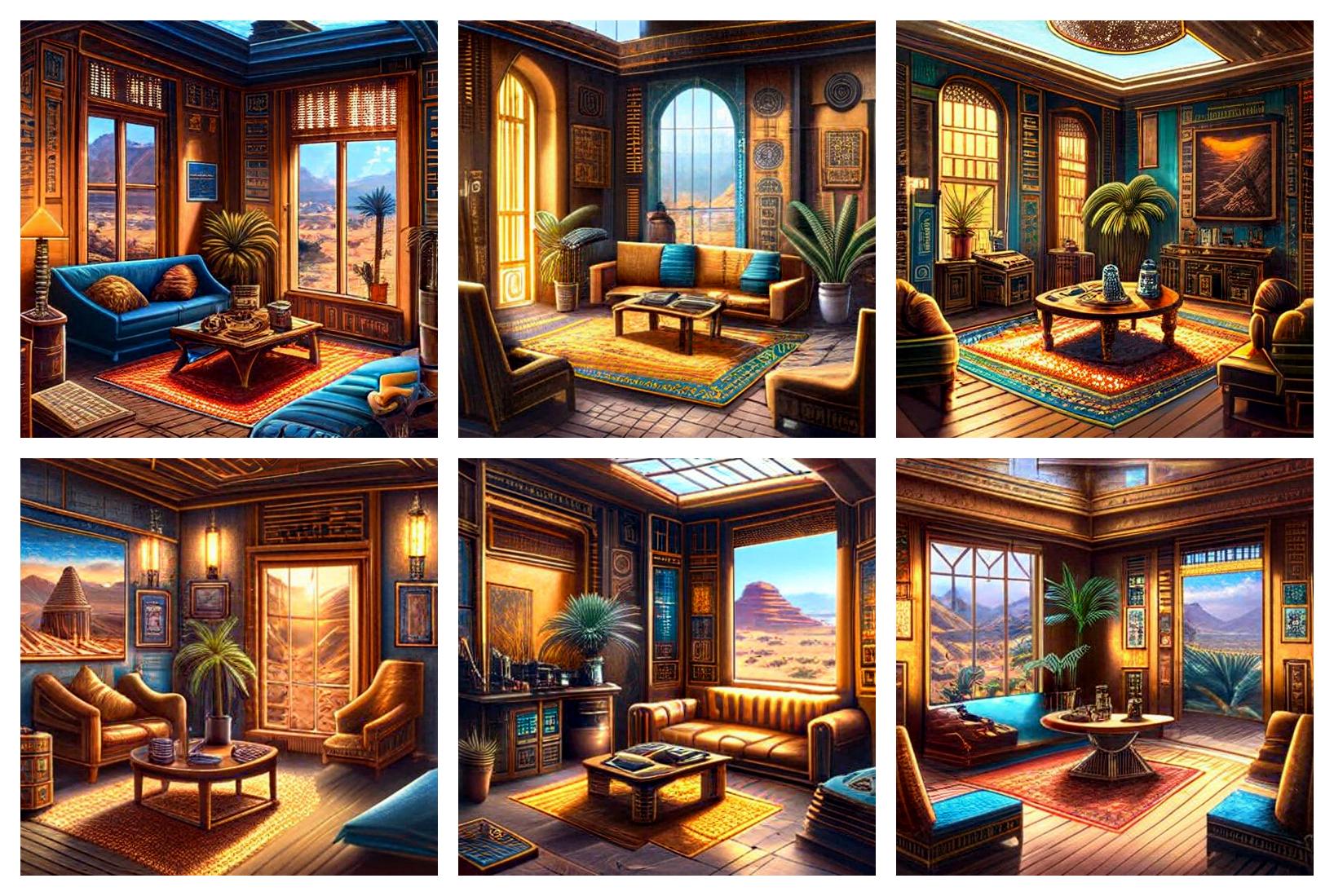}
        \end{minipage}
    \end{minipage}
    \begin{minipage}{0.95\linewidth}
        \centering
        \begin{minipage}{1.\linewidth}
            \centering
            {\scriptsize A female archer elf leads a group of adventurers through a forest of crystal trees in a fantasy matte painting.}
        \end{minipage}
        \begin{minipage}{0.32\linewidth}
            \centering
            \includegraphics[width=1.\linewidth]{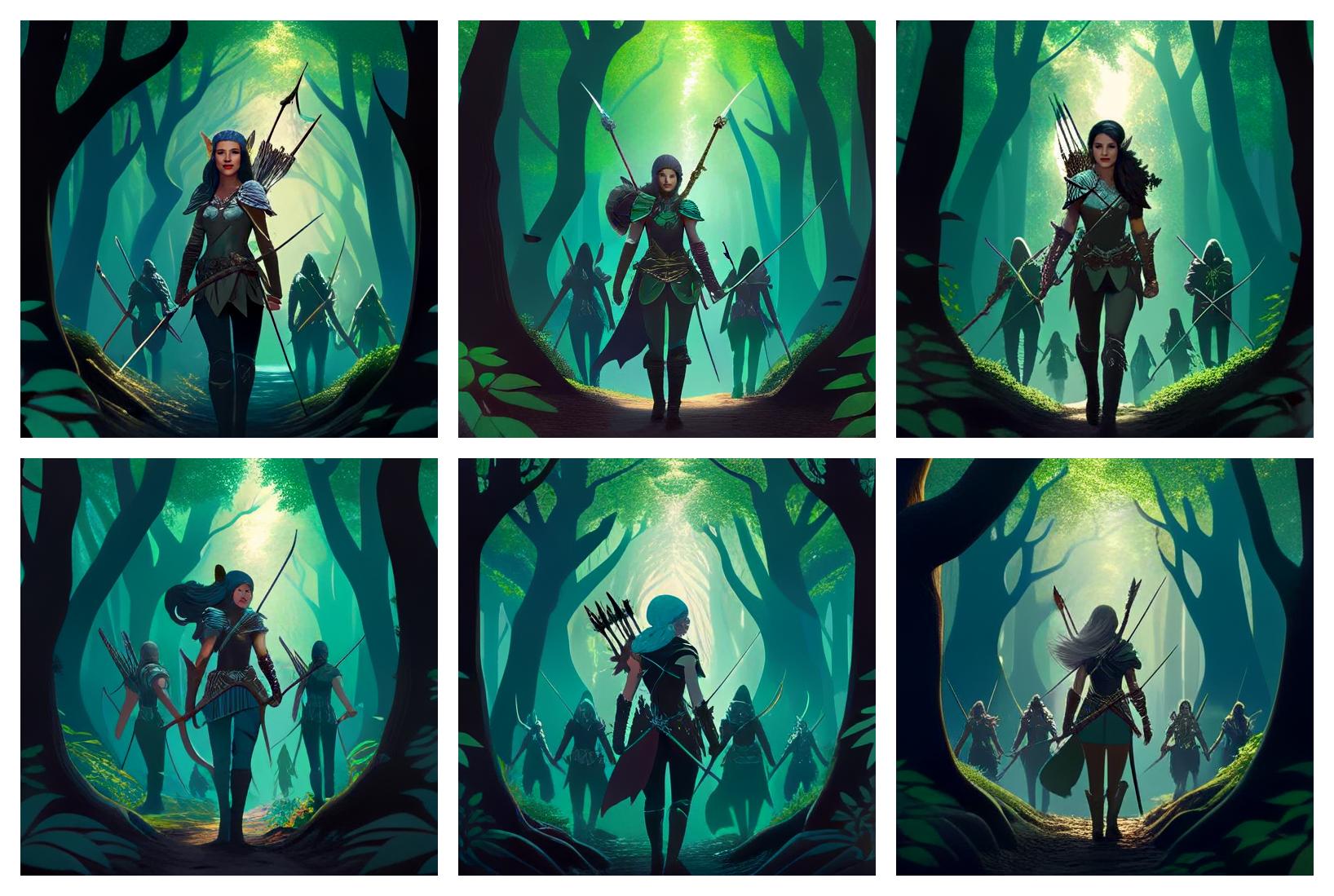}
        \end{minipage}
        \hfill
        \begin{minipage}{0.32\linewidth}
            \centering
            \includegraphics[width=1.\linewidth]{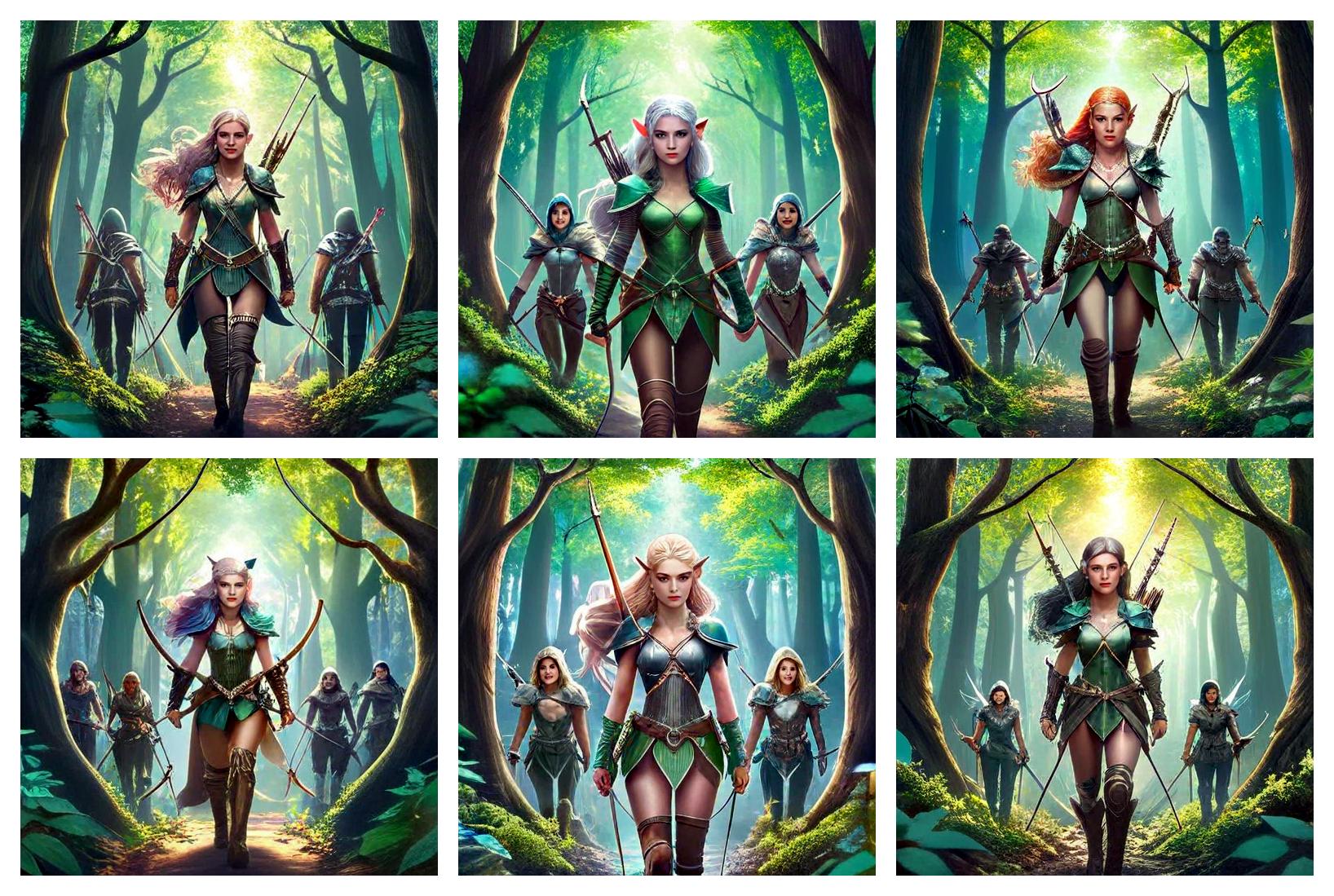}
        \end{minipage}
        \hfill
        \begin{minipage}{0.32\linewidth}
            \centering
            \includegraphics[width=1.\linewidth]{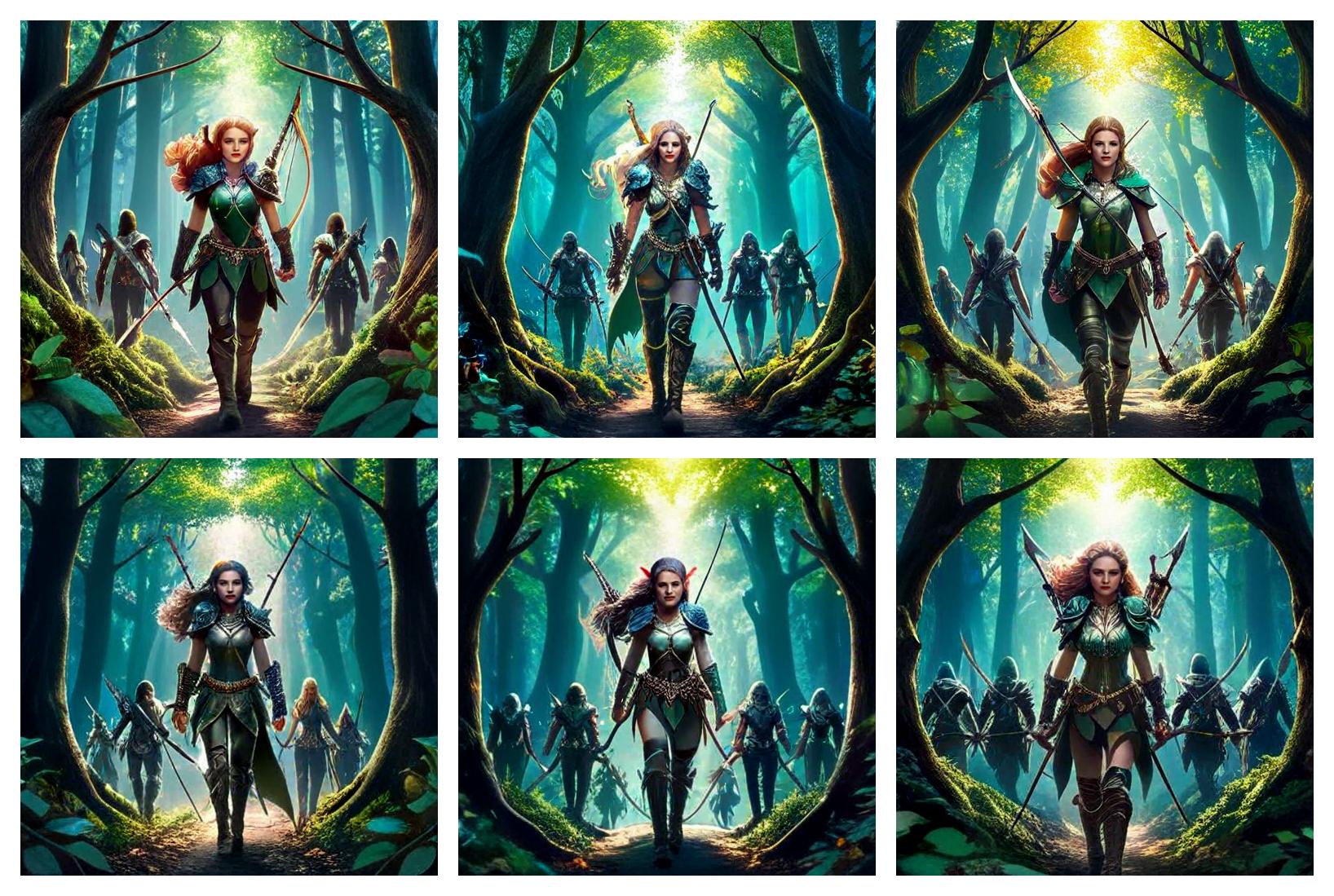}
        \end{minipage}
    \end{minipage}
    \begin{minipage}{0.95\linewidth}
        \centering
        \begin{minipage}{1.\linewidth}
            \centering
            {\scriptsize A pink bicycle leaning against a fence near a river.}
        \end{minipage}
        \begin{minipage}{0.32\linewidth}
            \centering
            \includegraphics[width=1.\linewidth]{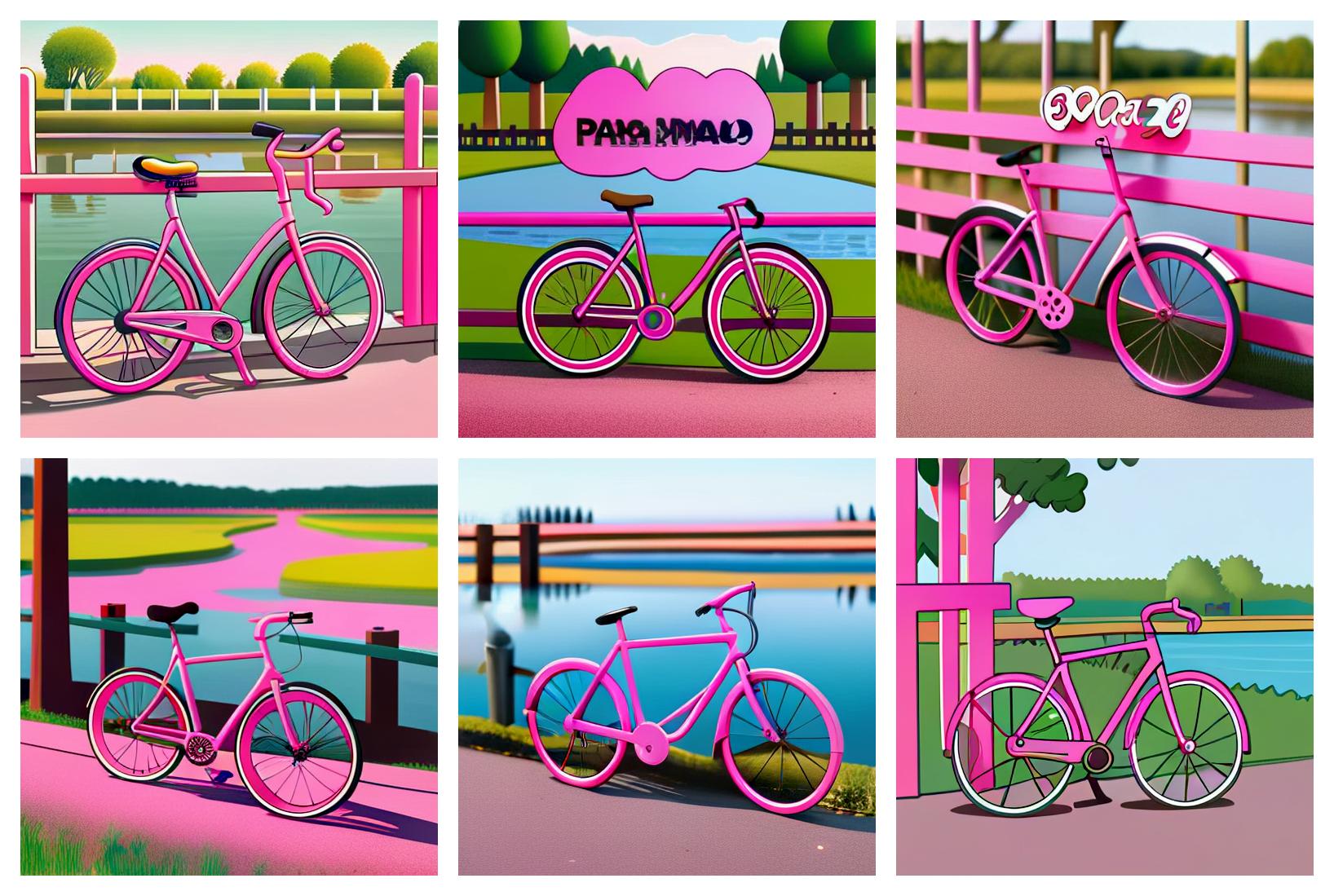}
        \end{minipage}
        \hfill
        \begin{minipage}{0.32\linewidth}
            \centering
            \includegraphics[width=1.\linewidth]{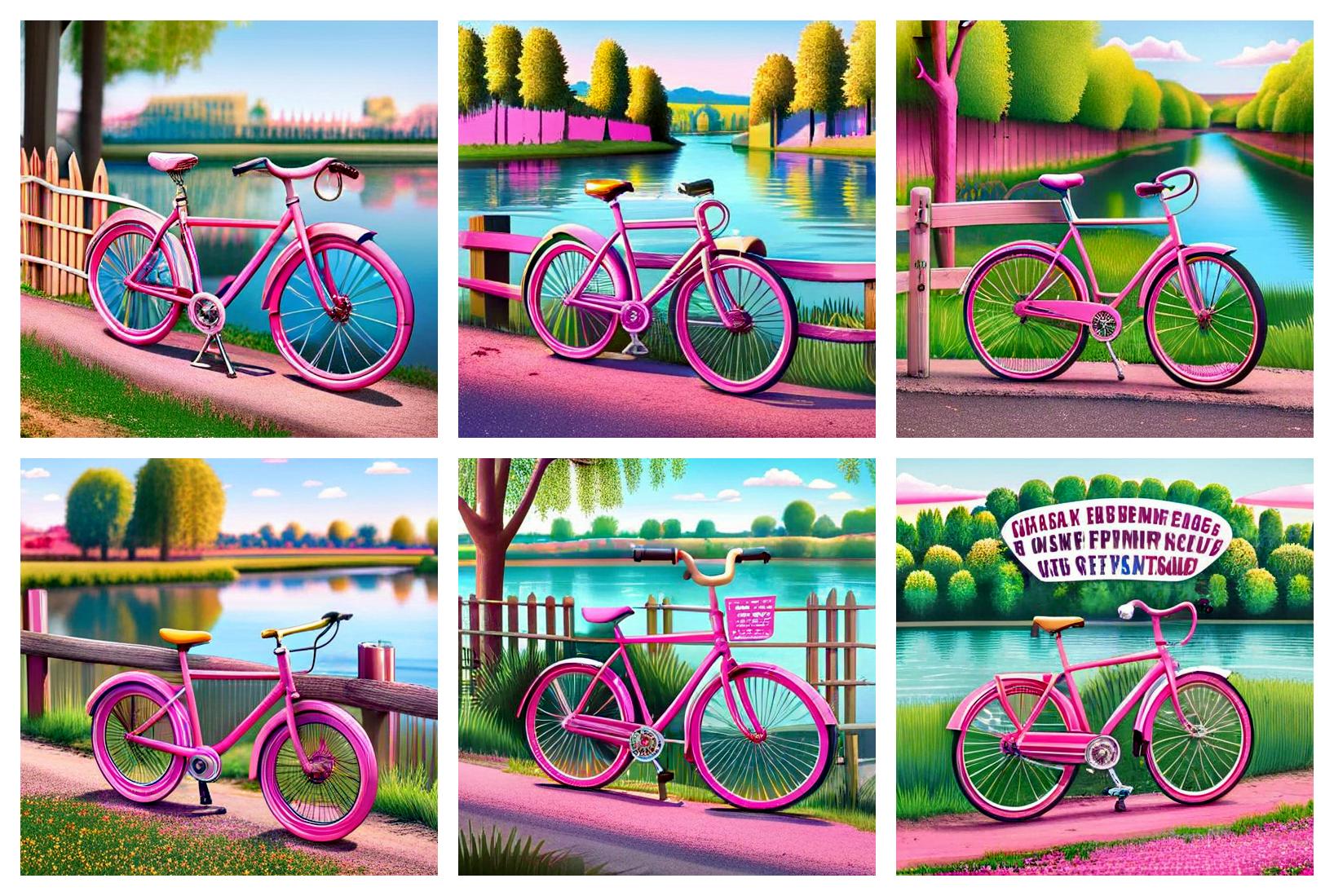}
        \end{minipage}
        \hfill
        \begin{minipage}{0.32\linewidth}
            \centering
            \includegraphics[width=1.\linewidth]{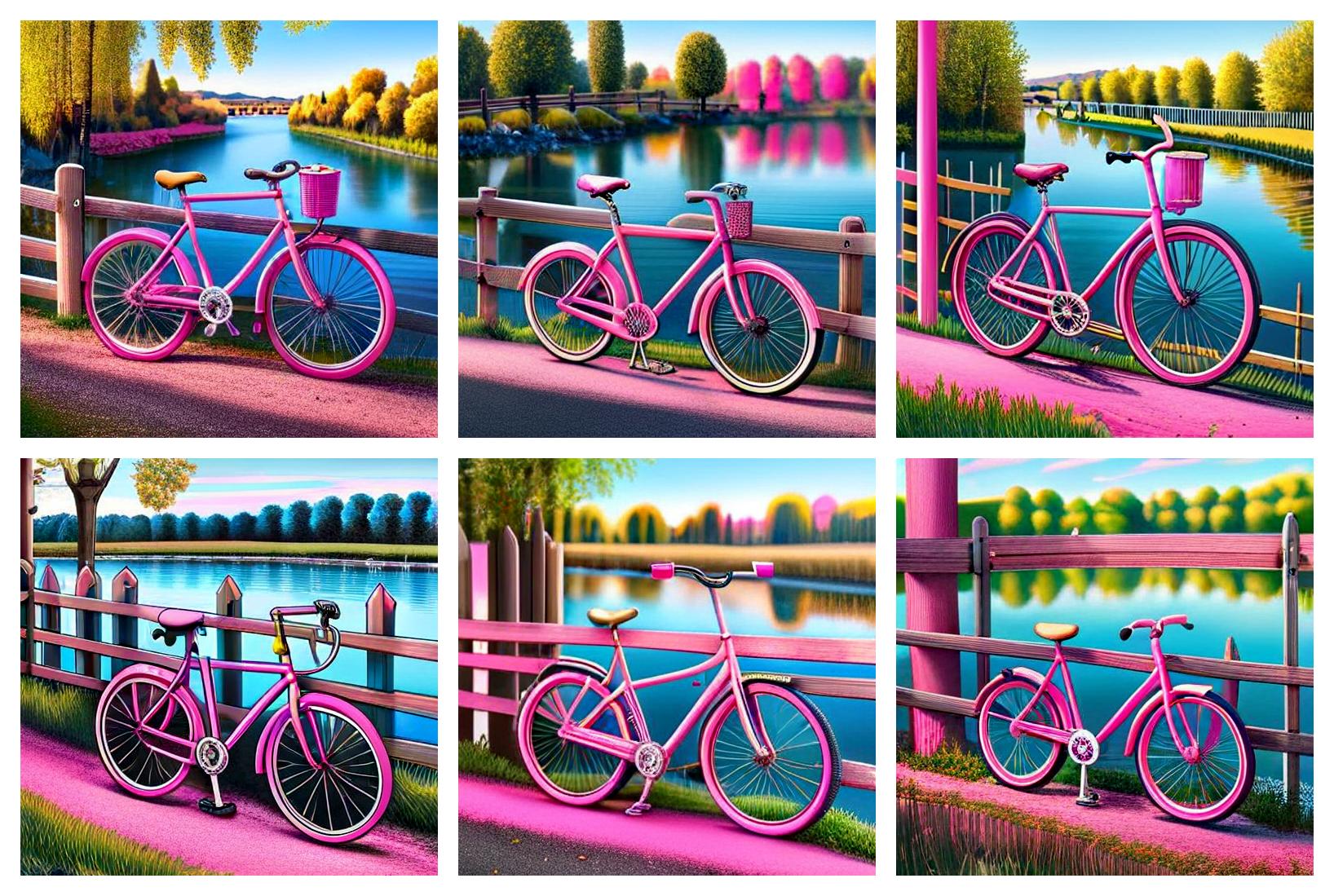}
        \end{minipage}
    \end{minipage}
    \begin{minipage}{0.95\linewidth}
        \centering
        \begin{minipage}{1.\linewidth}
            \centering
            {\scriptsize A surreal cat with a smile and intricate details.}
        \end{minipage}
        \begin{minipage}{0.32\linewidth}
            \centering
            \includegraphics[width=1.\linewidth]{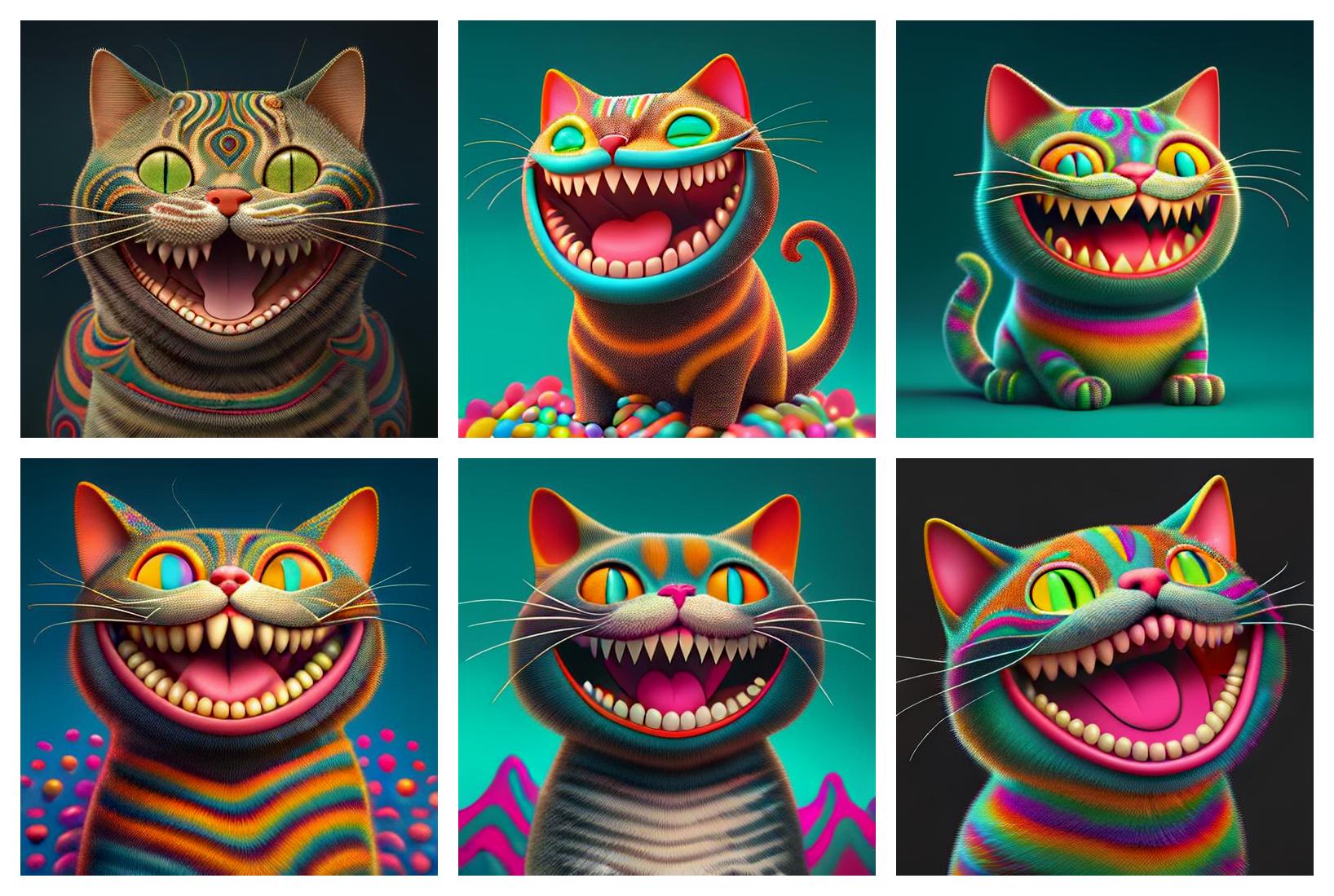}
        \end{minipage}
        \hfill
        \begin{minipage}{0.32\linewidth}
            \centering
            \includegraphics[width=1.\linewidth]{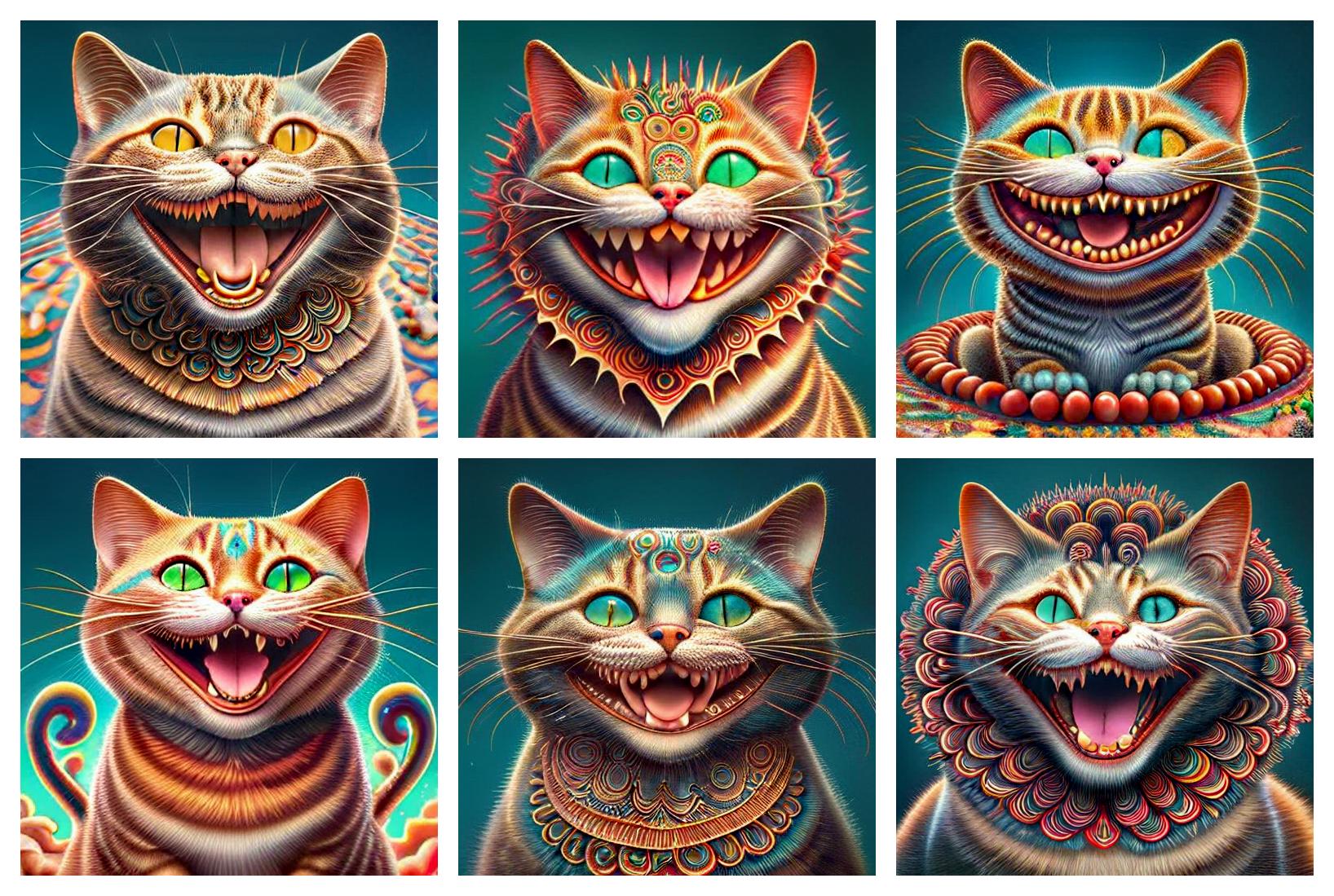}
        \end{minipage}
        \hfill
        \begin{minipage}{0.32\linewidth}
            \centering
            \includegraphics[width=1.\linewidth]{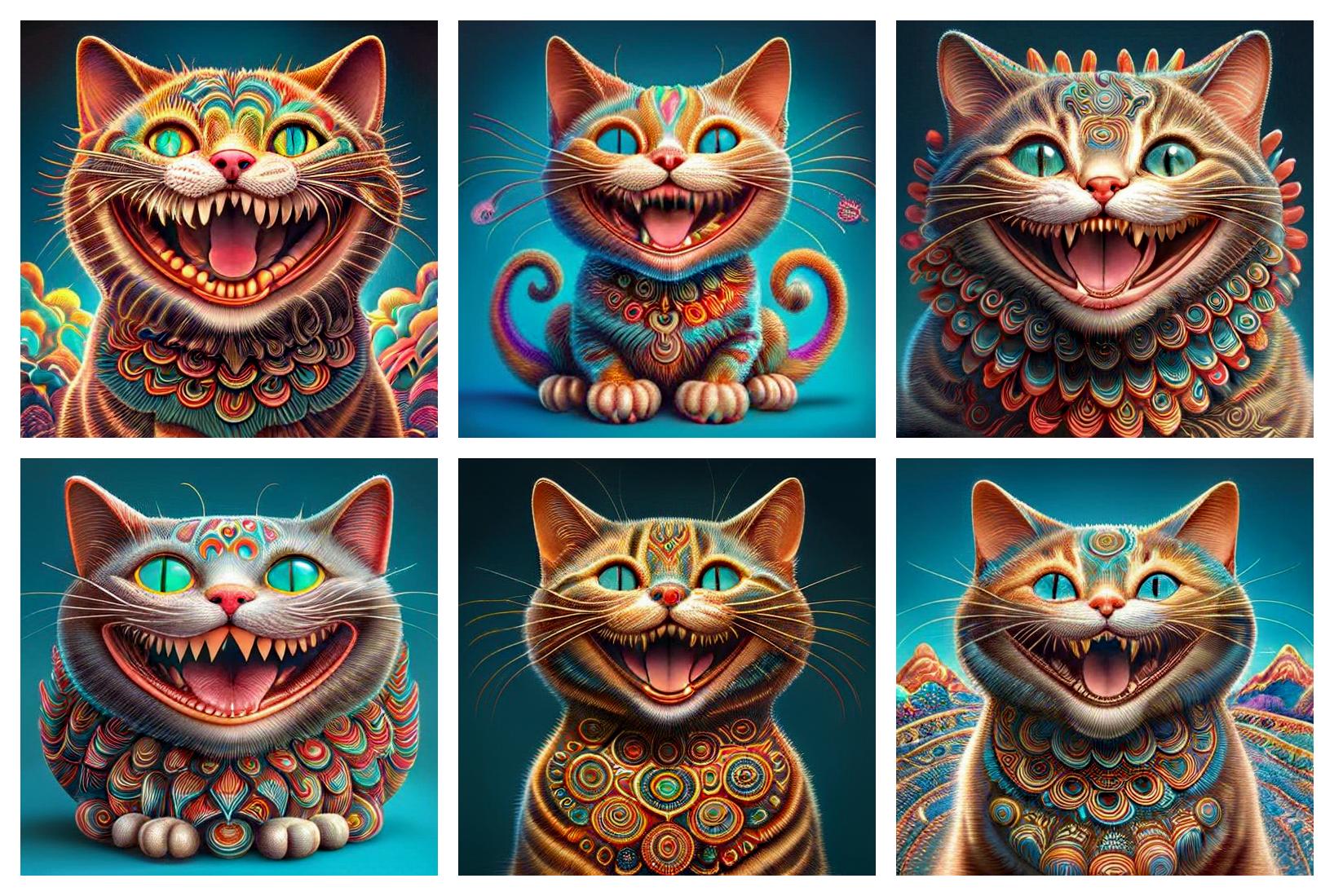}
        \end{minipage}
    \end{minipage}
    \caption{Illustration of the generated samples on HPSV2.}
    \label{fig:t2i-hpsv2-gen-samples-comp}
\end{figure}

\begin{figure}[!t]
    \centering
    \begin{minipage}{.95\linewidth}
        \centering
        \begin{minipage}{.32\linewidth}\centering Pretrained \end{minipage}
        \hfill
        \begin{minipage}{.32\linewidth}\centering $\text{Prop}_{\text{amot}}$ \end{minipage}
        \hfill
        \begin{minipage}{.32\linewidth}\centering \oursamot \end{minipage}
    \end{minipage}
    \begin{minipage}{0.95\linewidth}
        \centering
        \begin{minipage}{1.\linewidth}
            \centering
            {\scriptsize bird}
        \end{minipage}
        \begin{minipage}{0.32\linewidth}
            \centering
            \includegraphics[width=1.\linewidth]{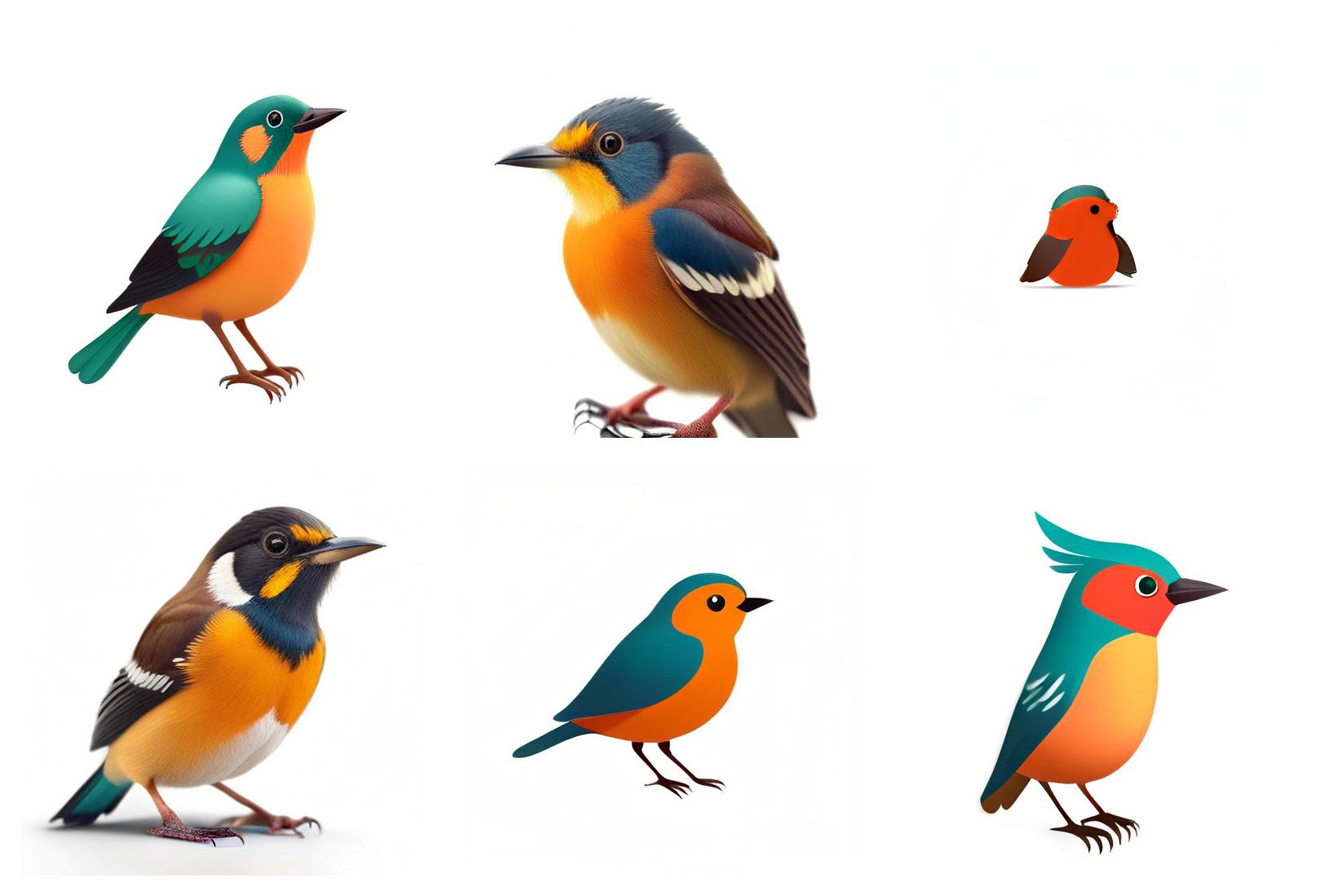}
        \end{minipage}
        \hfill
        \begin{minipage}{0.32\linewidth}
            \centering
            \includegraphics[width=1.\linewidth]{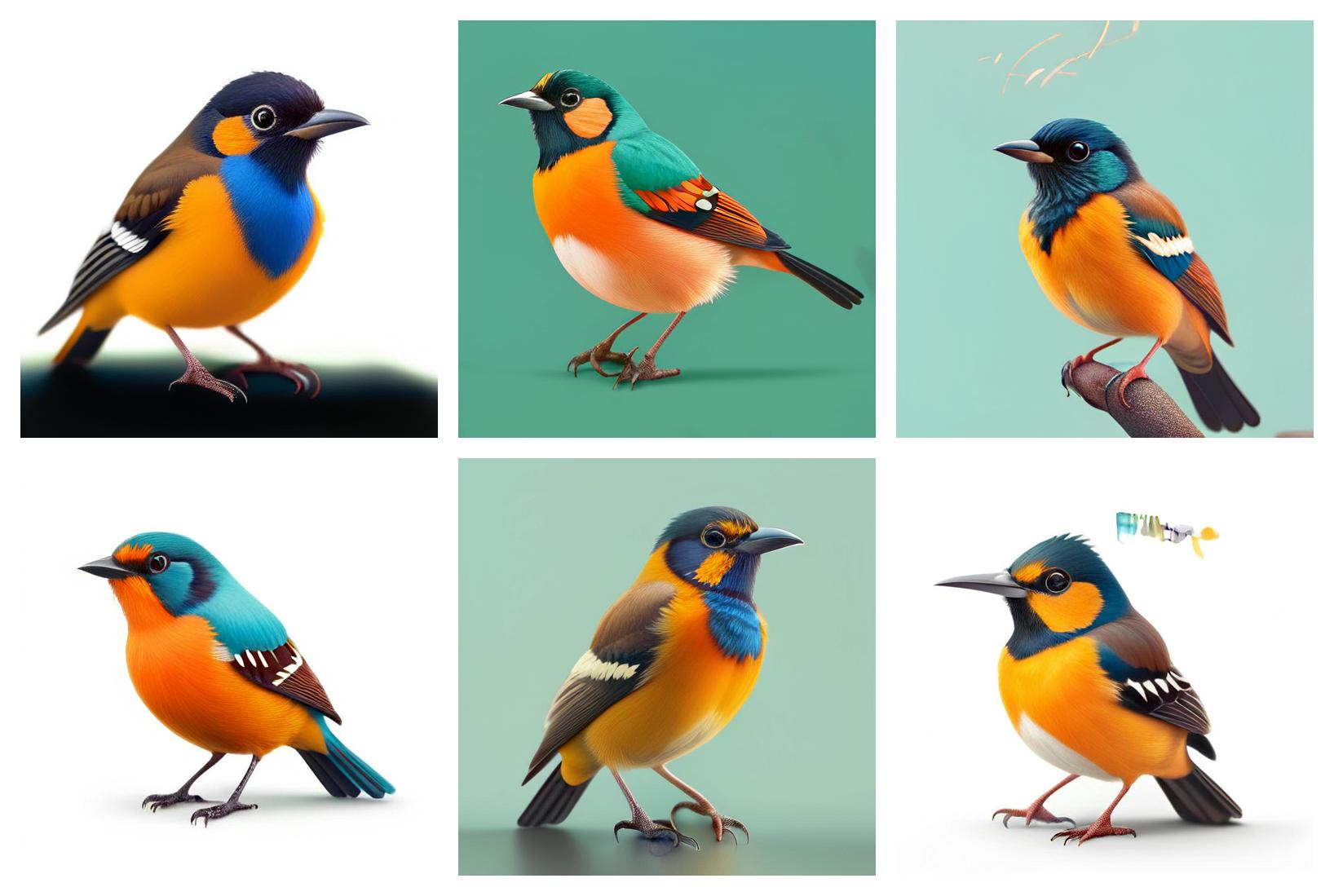}
        \end{minipage}
        \hfill
        \begin{minipage}{0.32\linewidth}
            \centering
            \includegraphics[width=1.\linewidth]{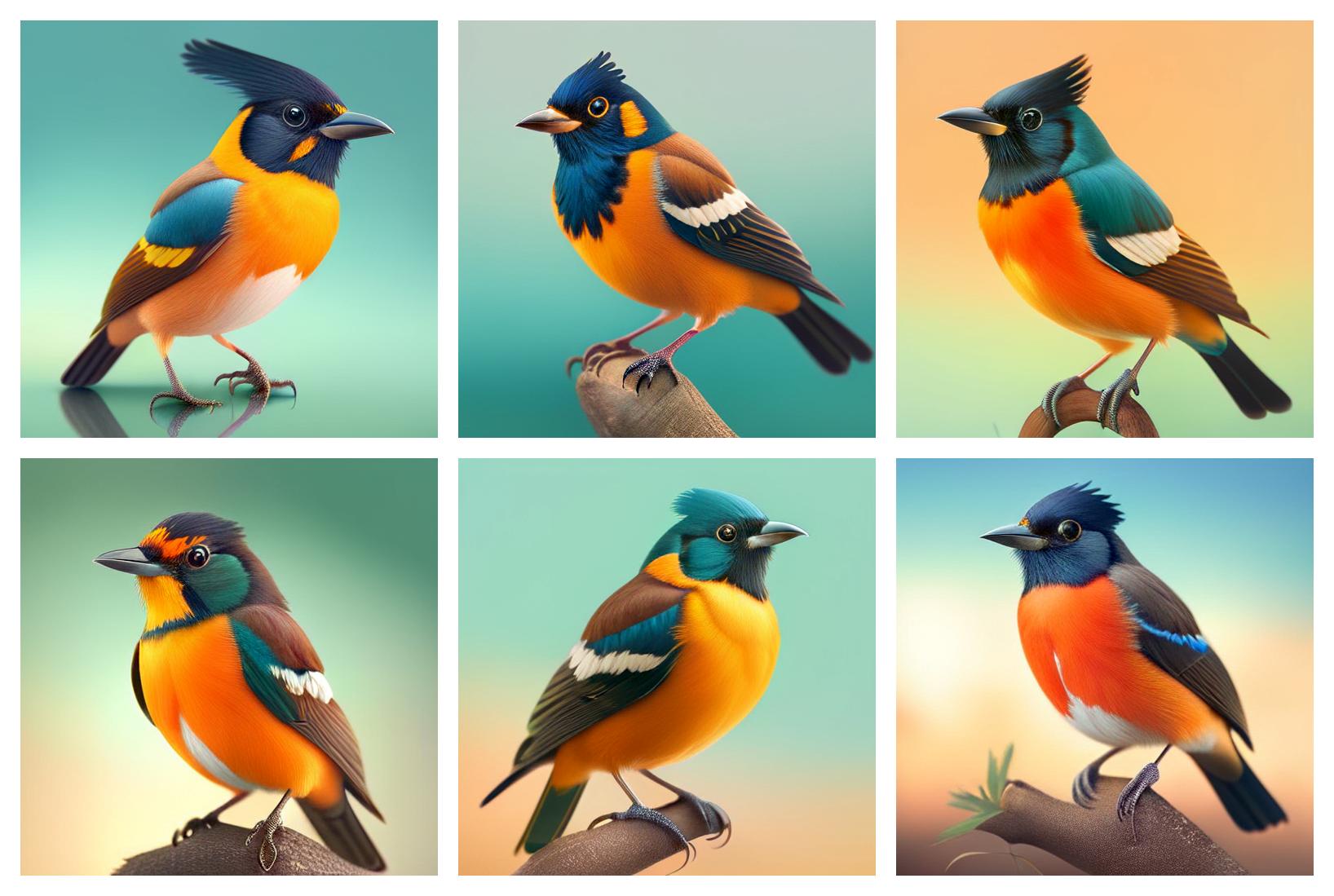}
        \end{minipage}
    \end{minipage}
    \begin{minipage}{0.95\linewidth}
        \centering
        \begin{minipage}{1.\linewidth}
            \centering
            {\scriptsize butterfly}
        \end{minipage}
        \begin{minipage}{0.32\linewidth}
            \centering
            \includegraphics[width=1.\linewidth]{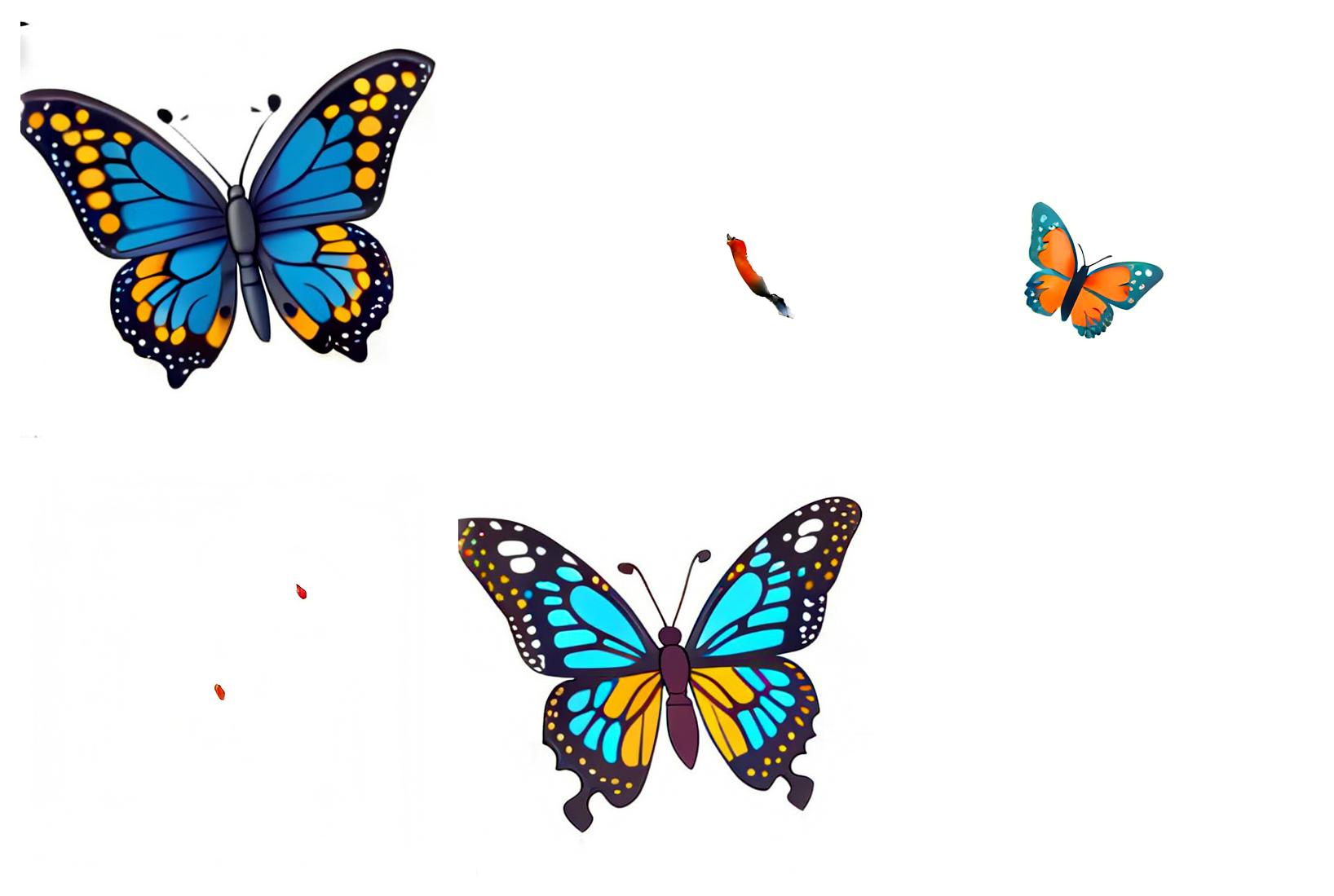}
        \end{minipage}
        \hfill
        \begin{minipage}{0.32\linewidth}
            \centering
            \includegraphics[width=1.\linewidth]{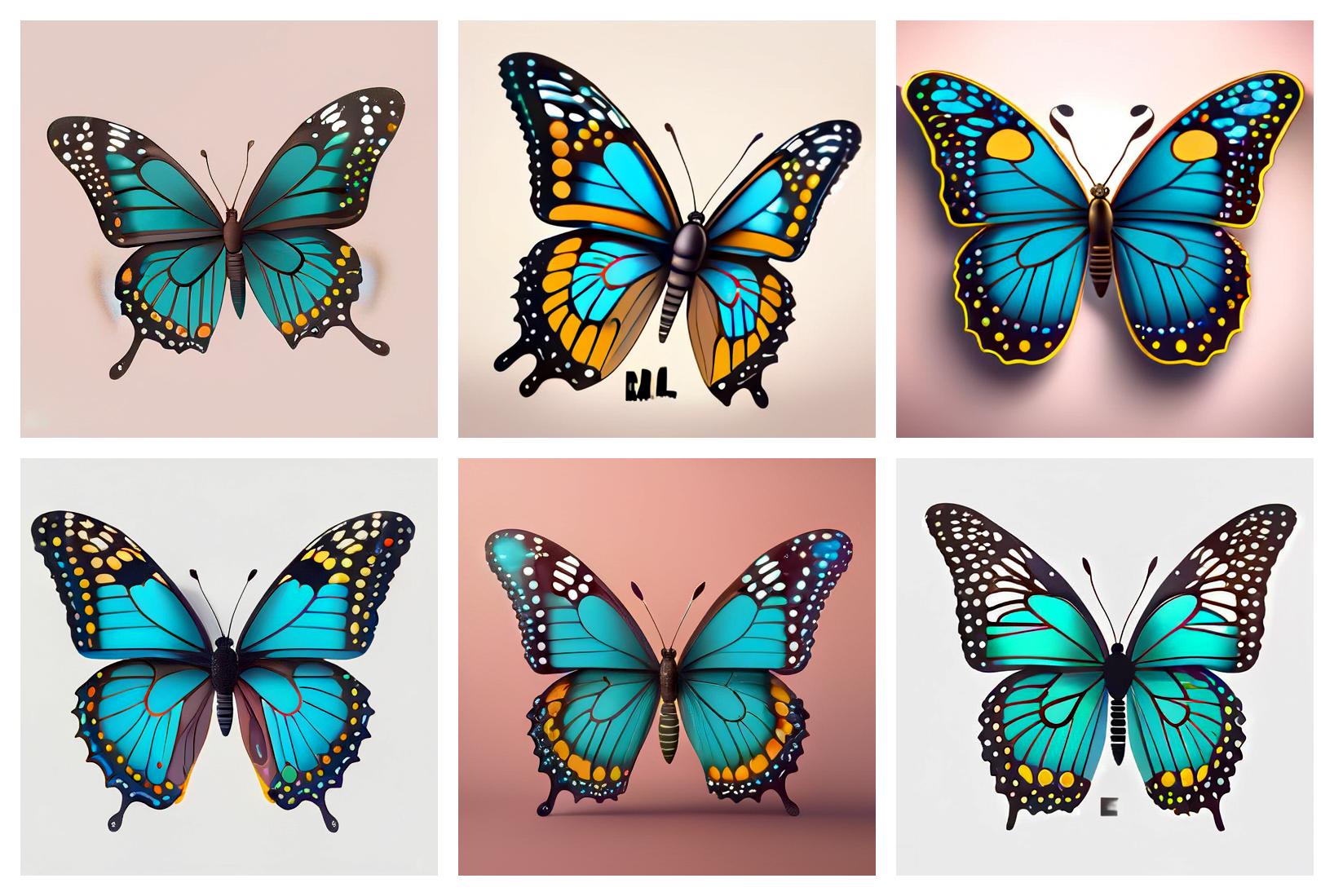}
        \end{minipage}
        \hfill
        \begin{minipage}{0.32\linewidth}
            \centering
            \includegraphics[width=1.\linewidth]{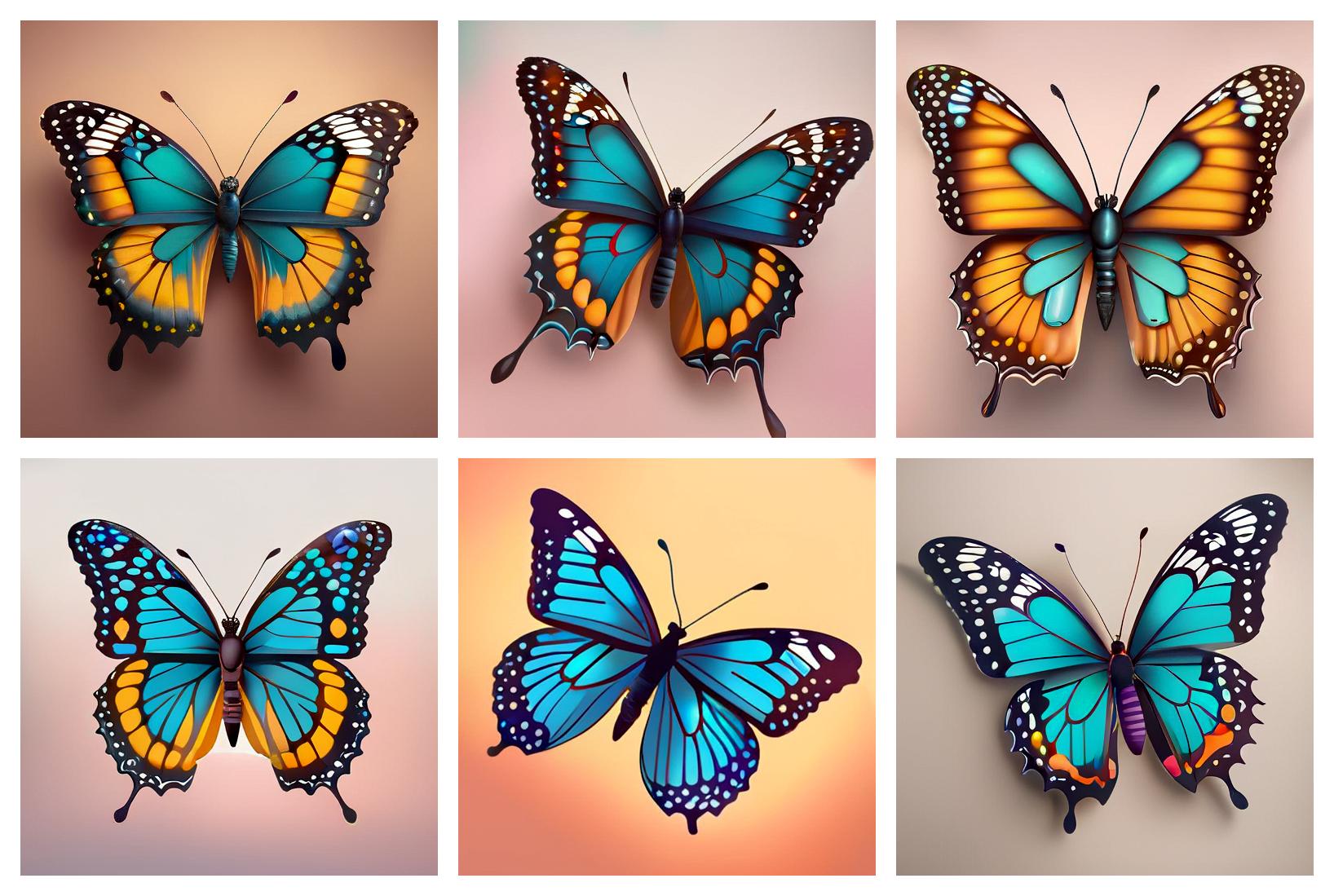}
        \end{minipage}
    \end{minipage}
    \begin{minipage}{0.95\linewidth}
        \centering
        \begin{minipage}{1.\linewidth}
            \centering
            {\scriptsize duck}
        \end{minipage}
        \begin{minipage}{0.32\linewidth}
            \centering
            \includegraphics[width=1.\linewidth]{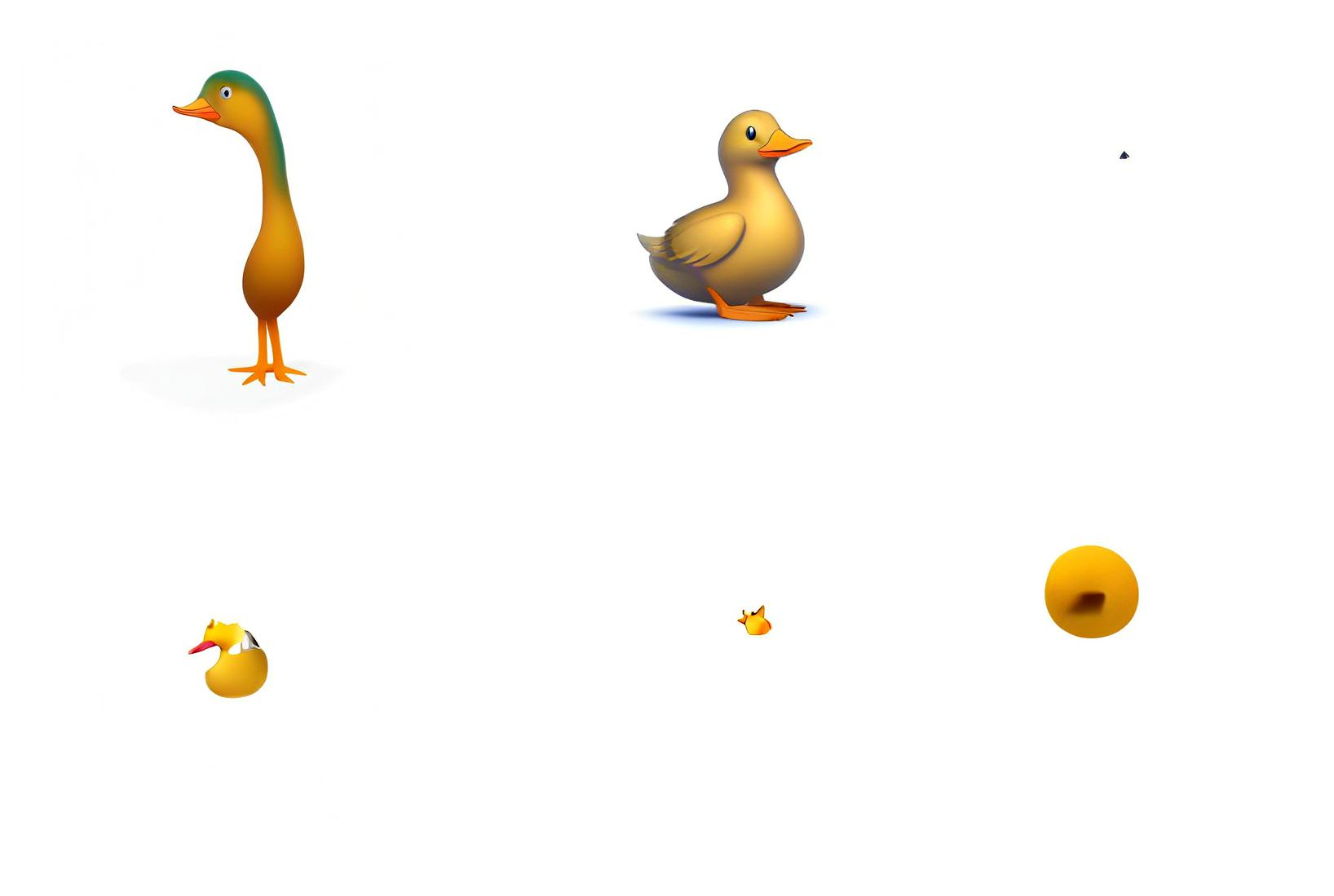}
        \end{minipage}
        \hfill
        \begin{minipage}{0.32\linewidth}
            \centering
            \includegraphics[width=1.\linewidth]{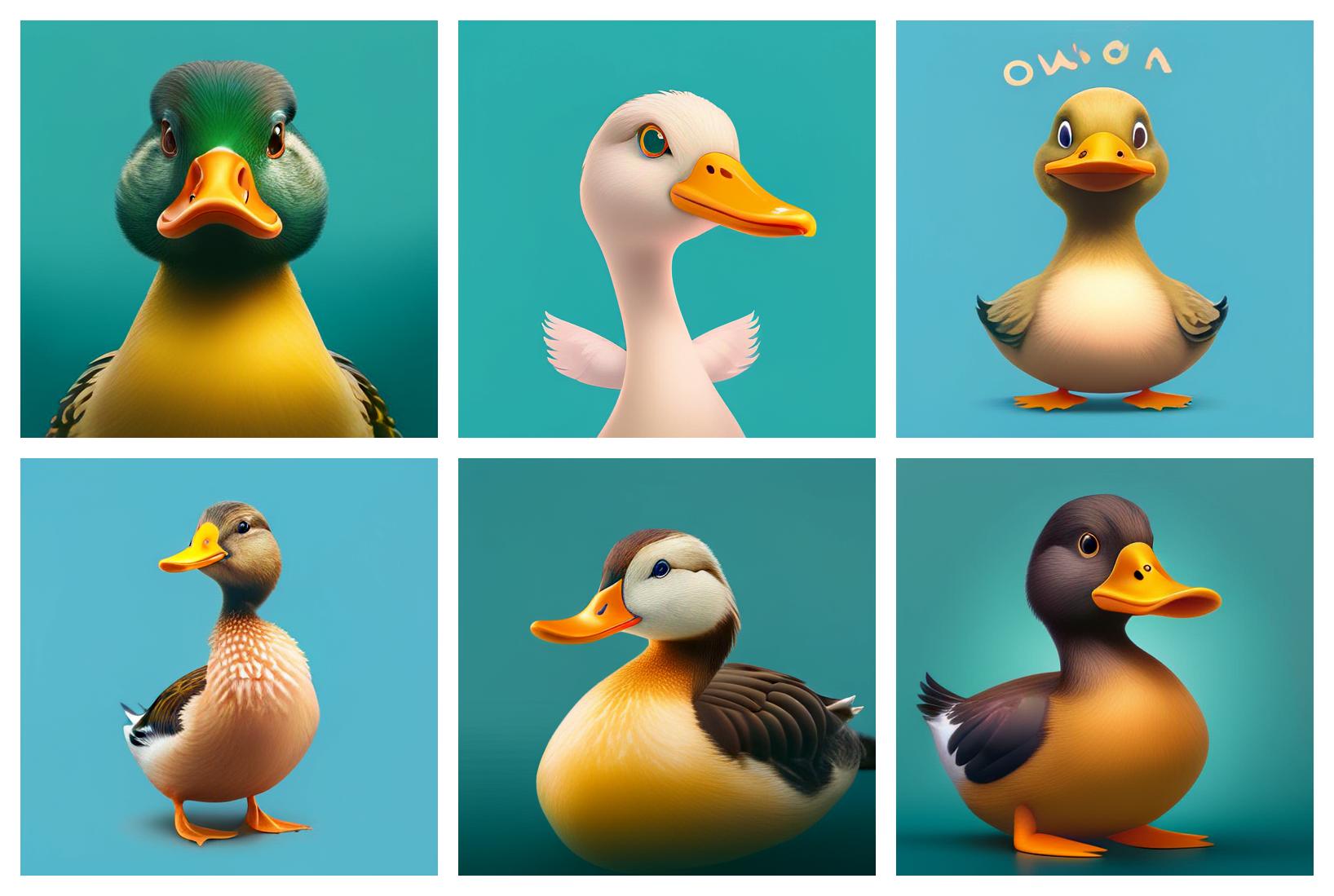}
        \end{minipage}
        \hfill
        \begin{minipage}{0.32\linewidth}
            \centering
            \includegraphics[width=1.\linewidth]{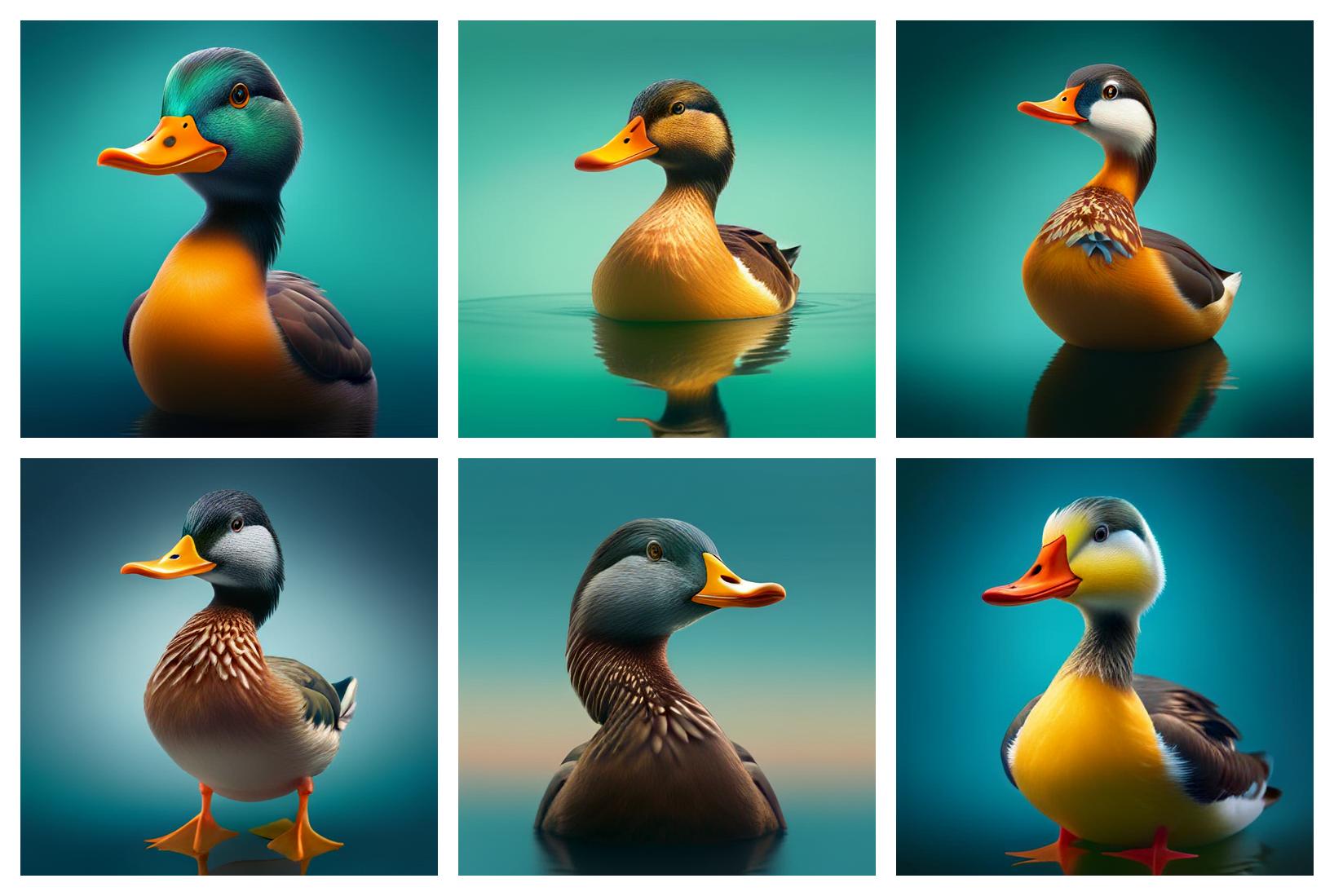}
        \end{minipage}
    \end{minipage}
    \begin{minipage}{0.95\linewidth}
        \centering
        \begin{minipage}{1.\linewidth}
            \centering
            {\scriptsize penguin}
        \end{minipage}
        \begin{minipage}{0.32\linewidth}
            \centering
            \includegraphics[width=1.\linewidth]{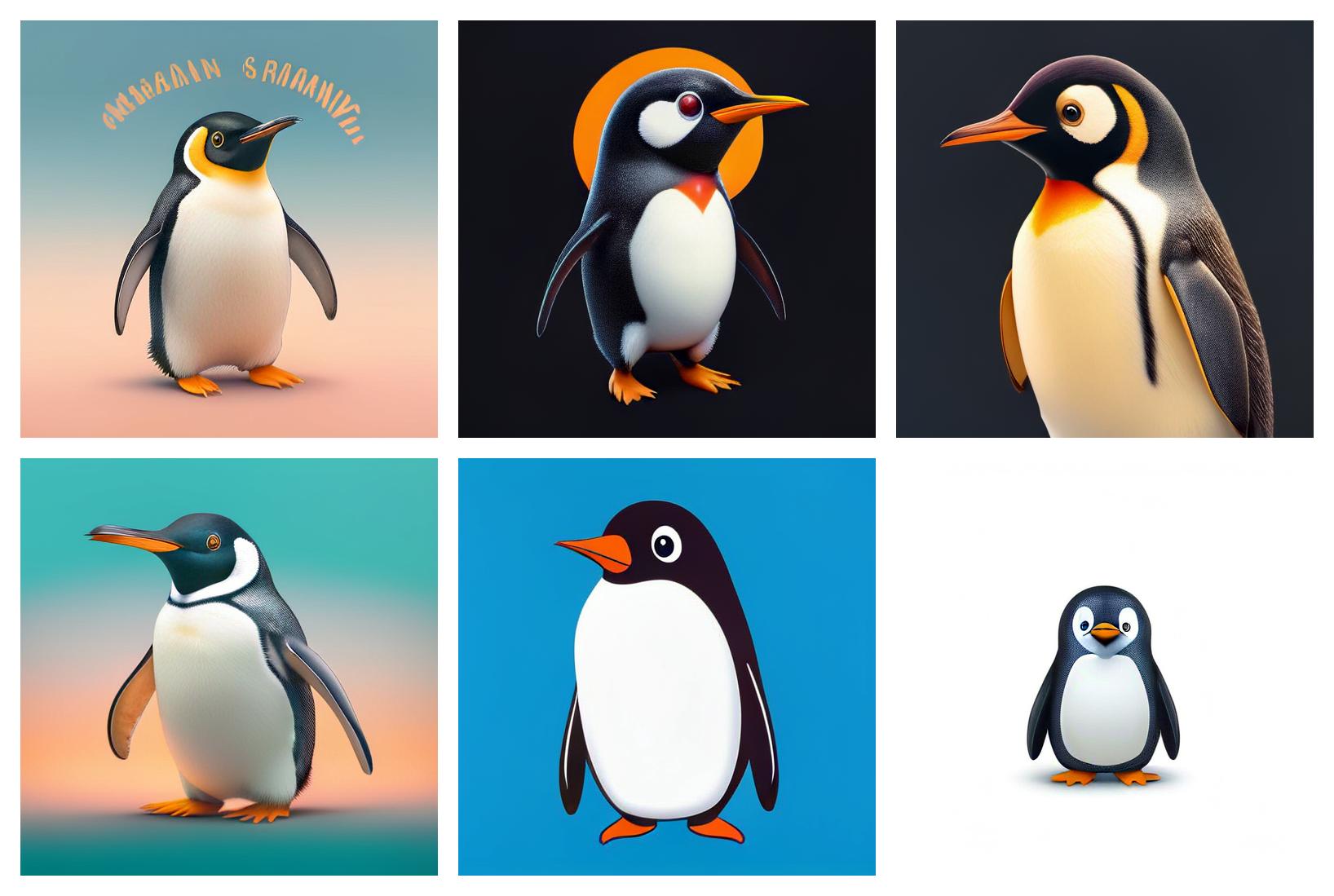}
        \end{minipage}
        \hfill
        \begin{minipage}{0.32\linewidth}
            \centering
            \includegraphics[width=1.\linewidth]{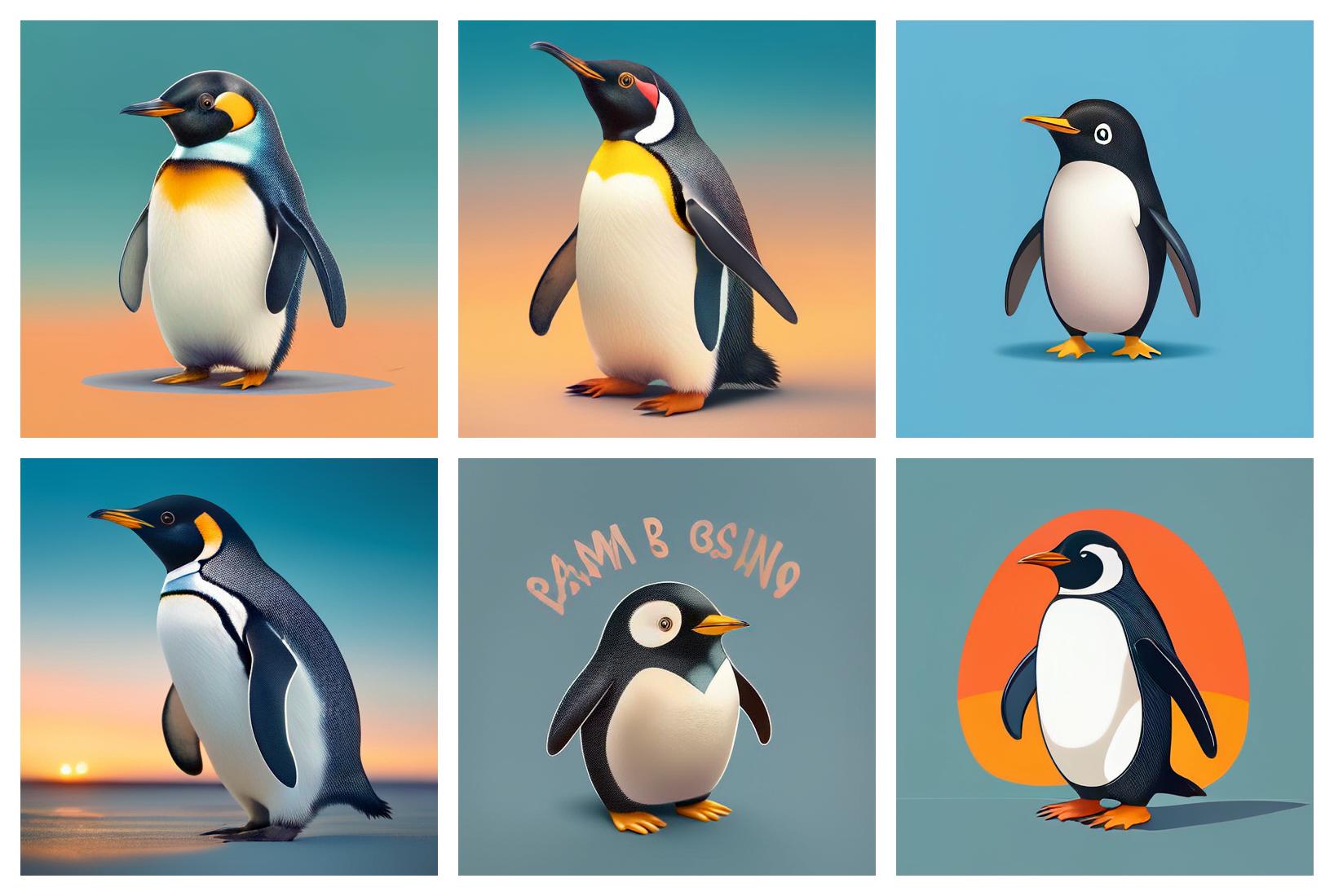}
        \end{minipage}
        \hfill
        \begin{minipage}{0.32\linewidth}
            \centering
            \includegraphics[width=1.\linewidth]{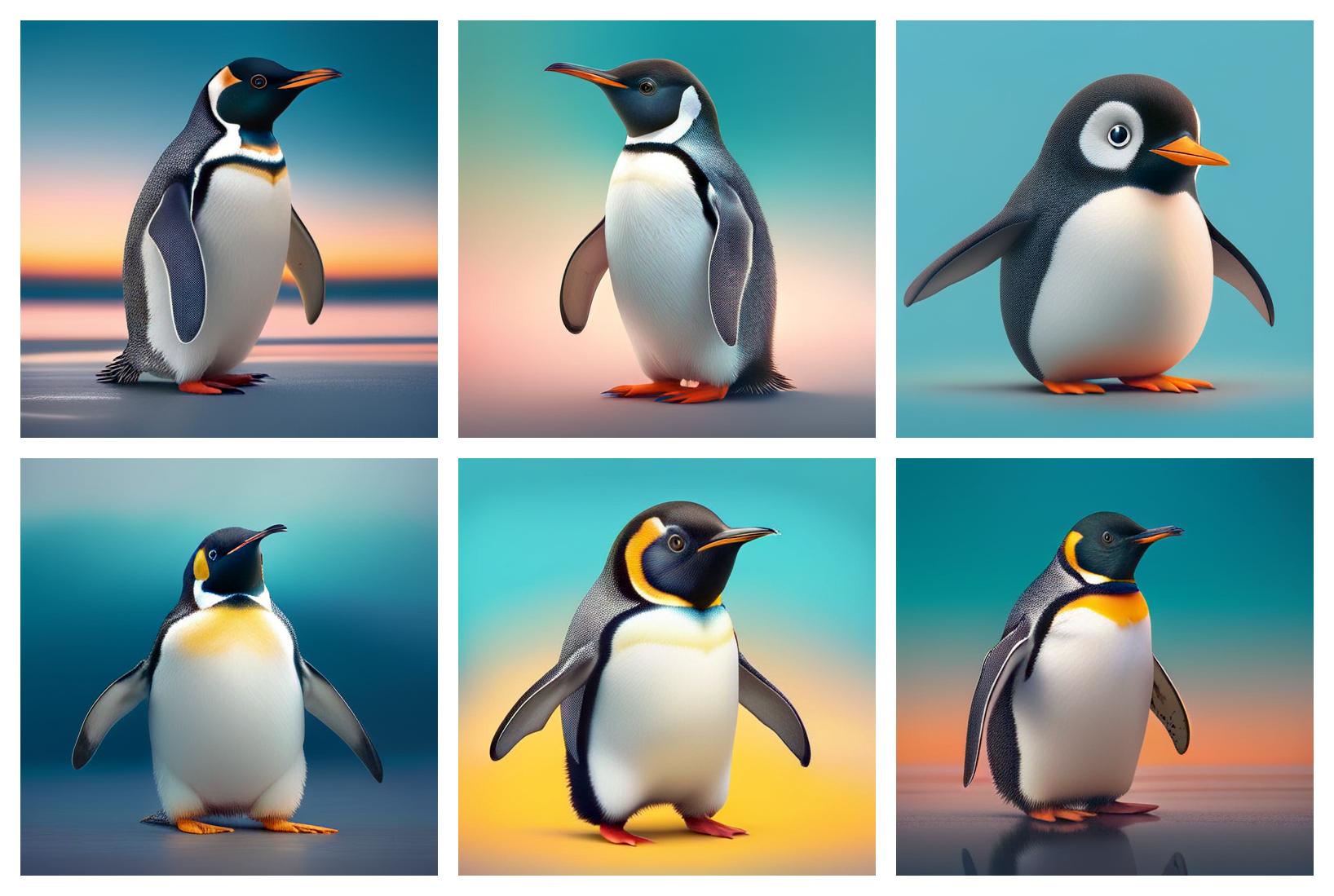}
        \end{minipage}
    \end{minipage}
    \begin{minipage}{0.95\linewidth}
        \centering
        \begin{minipage}{1.\linewidth}
            \centering
            {\scriptsize lion}
        \end{minipage}
        \begin{minipage}{0.32\linewidth}
            \centering
            \includegraphics[width=1.\linewidth]{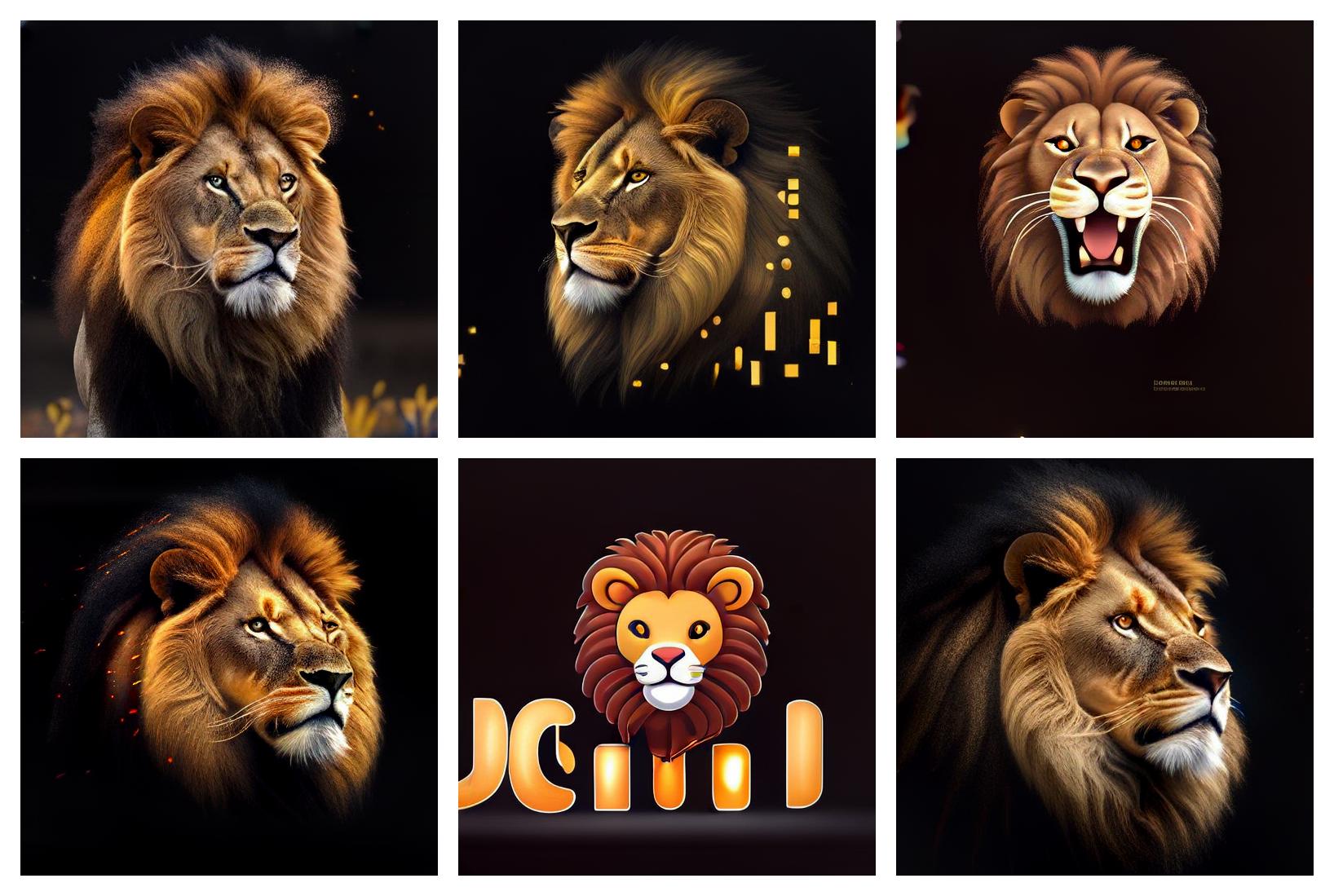}
        \end{minipage}
        \hfill
        \begin{minipage}{0.32\linewidth}
            \centering
            \includegraphics[width=1.\linewidth]{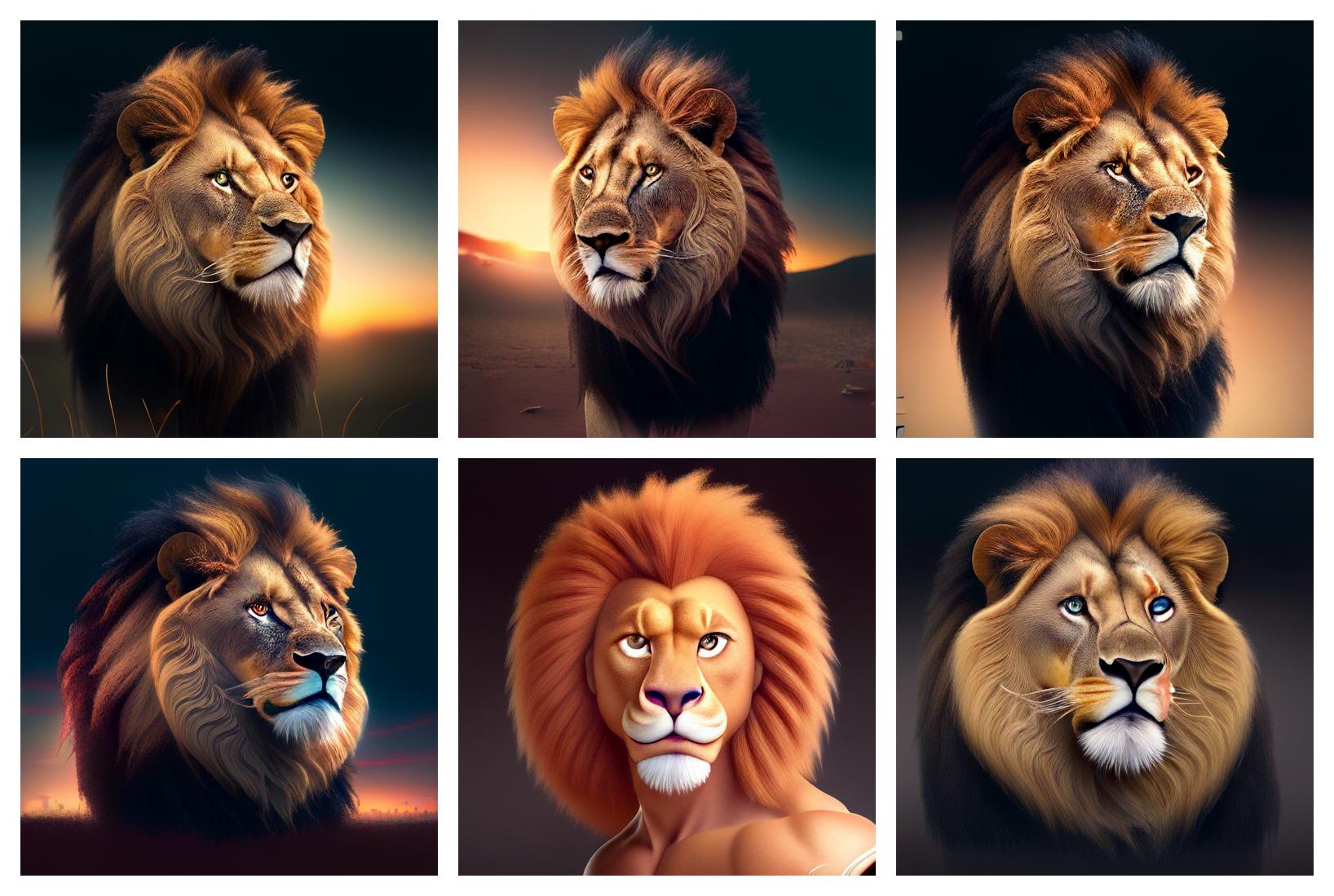}
        \end{minipage}
        \hfill
        \begin{minipage}{0.32\linewidth}
            \centering
            \includegraphics[width=1.\linewidth]{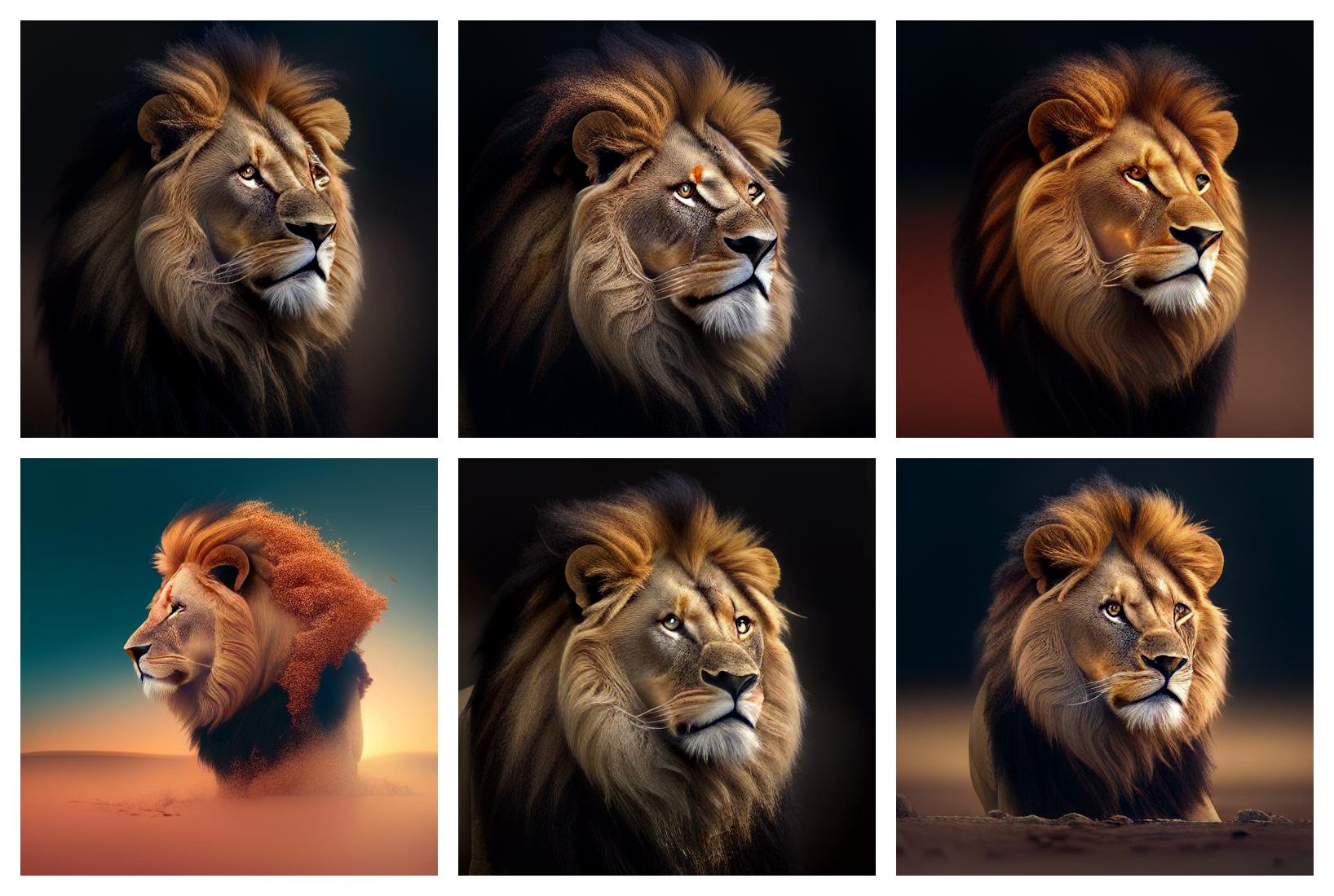}
        \end{minipage}
    \end{minipage}
    \begin{minipage}{0.95\linewidth}
        \centering
        \begin{minipage}{1.\linewidth}
            \centering
            {\scriptsize giraffe}
        \end{minipage}
        \begin{minipage}{0.32\linewidth}
            \centering
            \includegraphics[width=1.\linewidth]{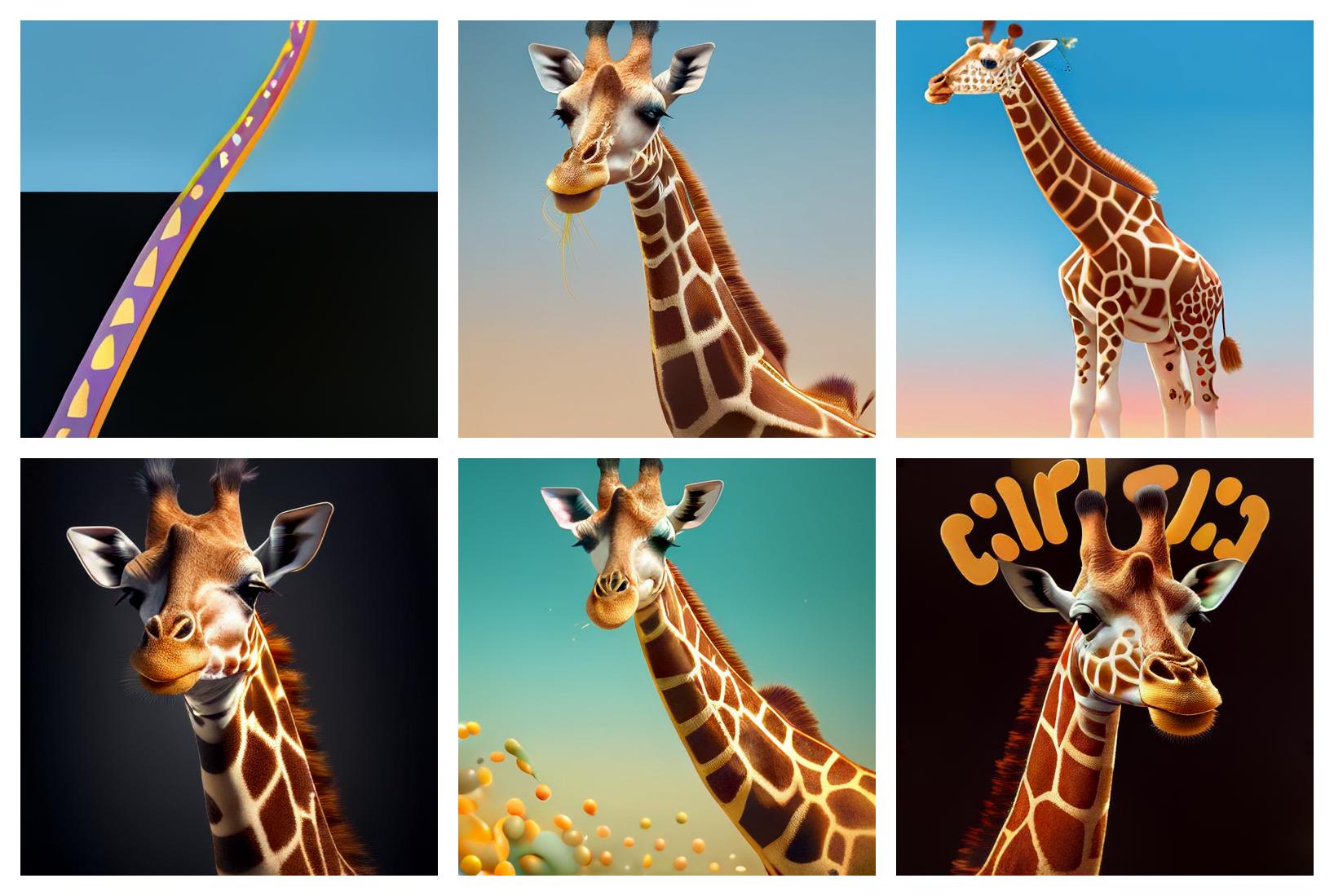}
        \end{minipage}
        \hfill
        \begin{minipage}{0.32\linewidth}
            \centering
            \includegraphics[width=1.\linewidth]{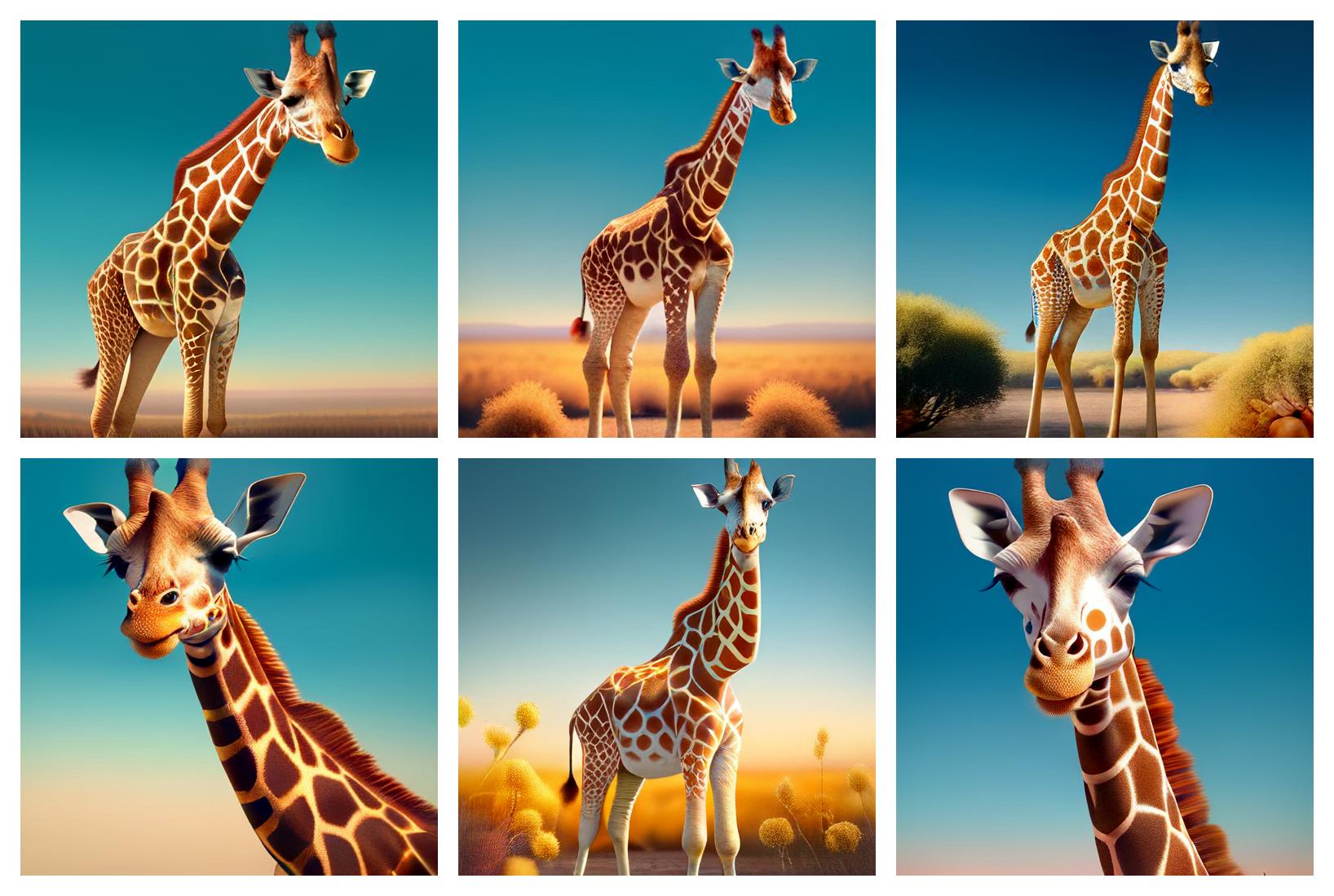}
        \end{minipage}
        \hfill
        \begin{minipage}{0.32\linewidth}
            \centering
            \includegraphics[width=1.\linewidth]{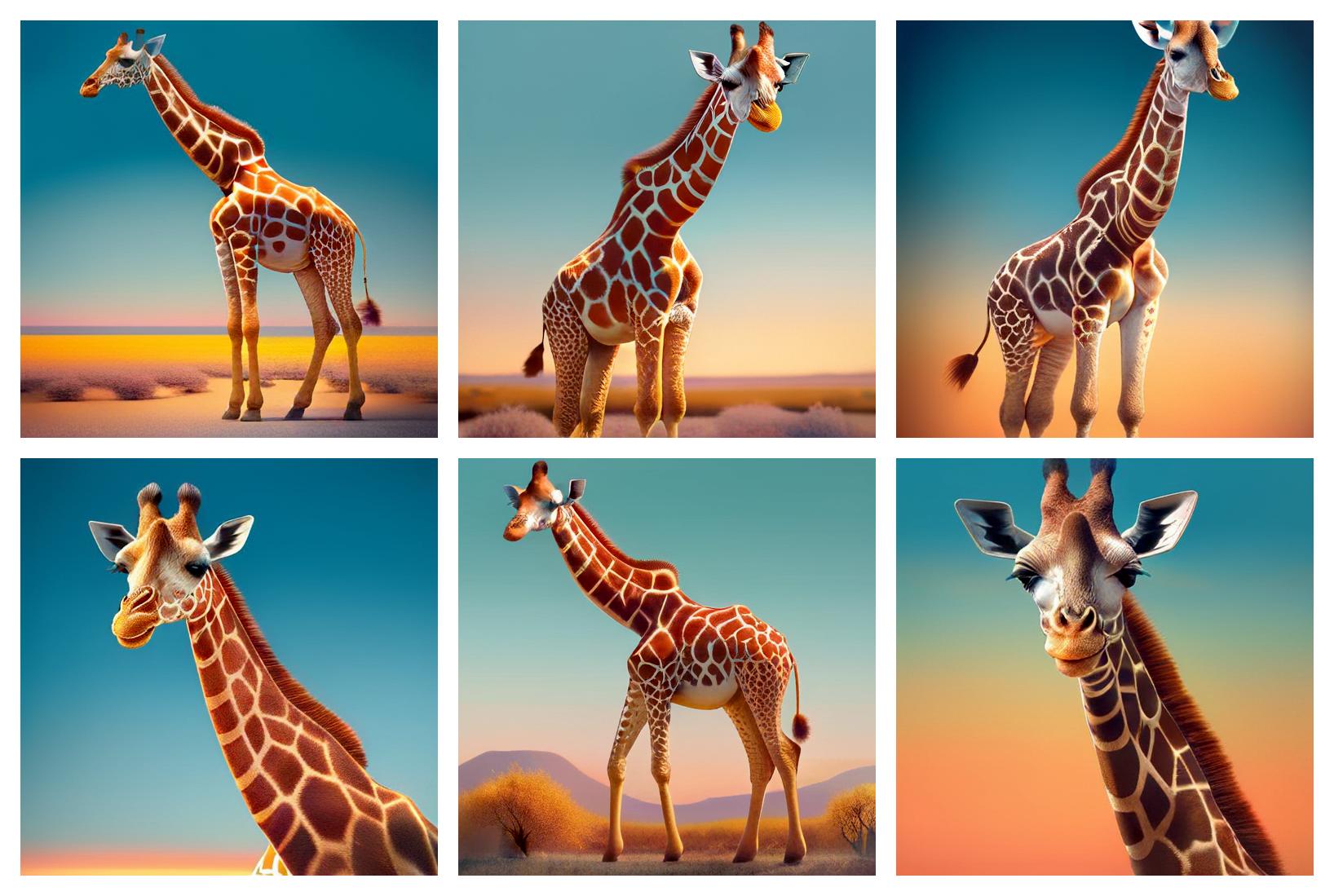}
        \end{minipage}
    \end{minipage}
    \caption{Illustration of the generated samples on Aesthetic Score.}
    \label{fig:t2i-aesthetic-gen-samples-comp}
\end{figure}

\begin{figure}[!t]
    \centering
    \begin{minipage}{.95\linewidth}
        \centering
        \begin{minipage}{.32\linewidth}\centering Pretrained \end{minipage}
        \hfill
        \begin{minipage}{.32\linewidth}\centering $\text{Prop}_{\text{amot}}$ \end{minipage}
        \hfill
        \begin{minipage}{.32\linewidth}\centering \oursamot \end{minipage}
    \end{minipage}
    \begin{minipage}{0.95\linewidth}
        \centering
        \begin{minipage}{1.\linewidth}
            \centering
            {\scriptsize A cat in the style of Van Gogh's Starry Night.}
        \end{minipage}
        \begin{minipage}{0.32\linewidth}
            \centering
            \includegraphics[width=1.\linewidth]{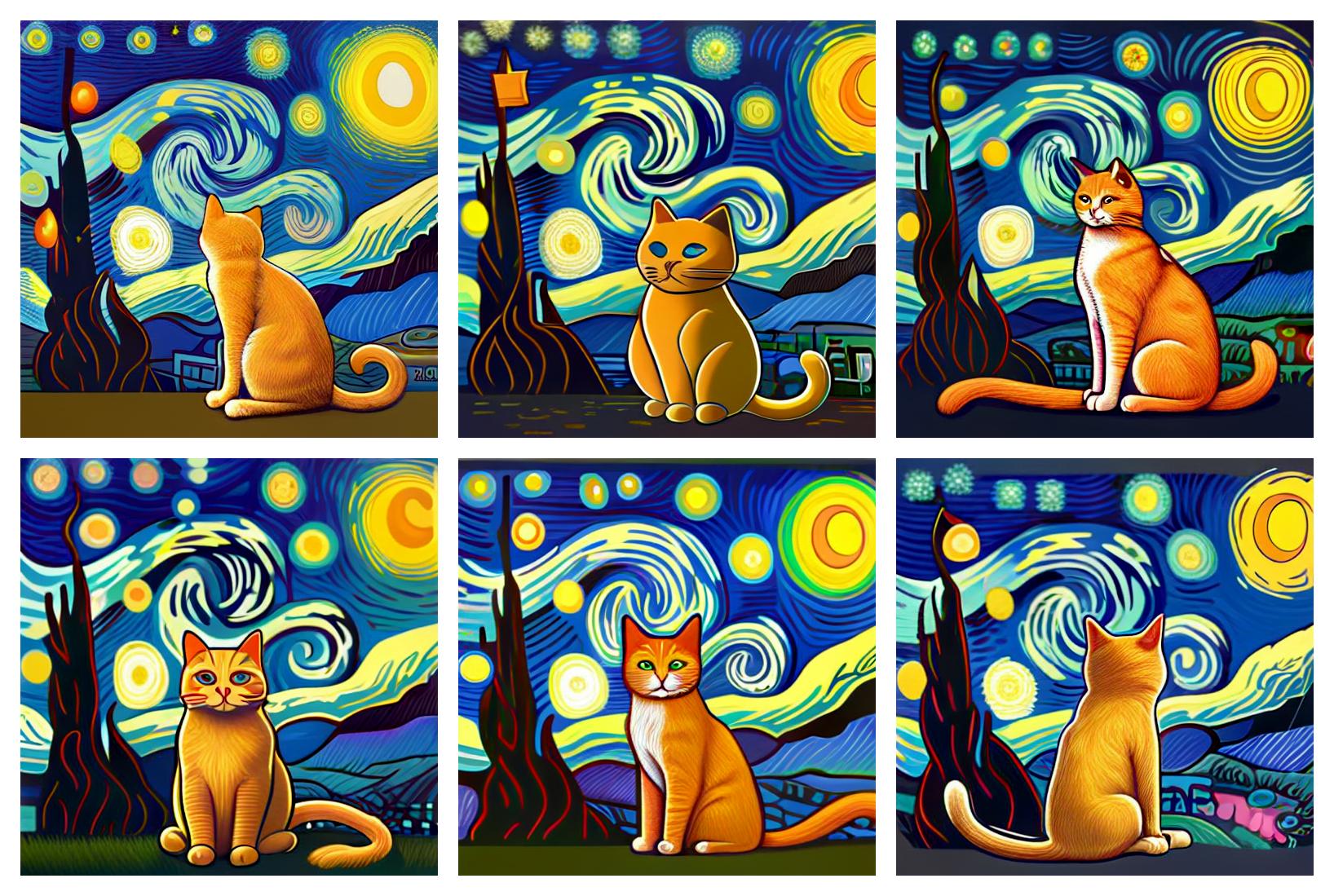}
        \end{minipage}
        \hfill
        \begin{minipage}{0.32\linewidth}
            \centering
            \includegraphics[width=1.\linewidth]{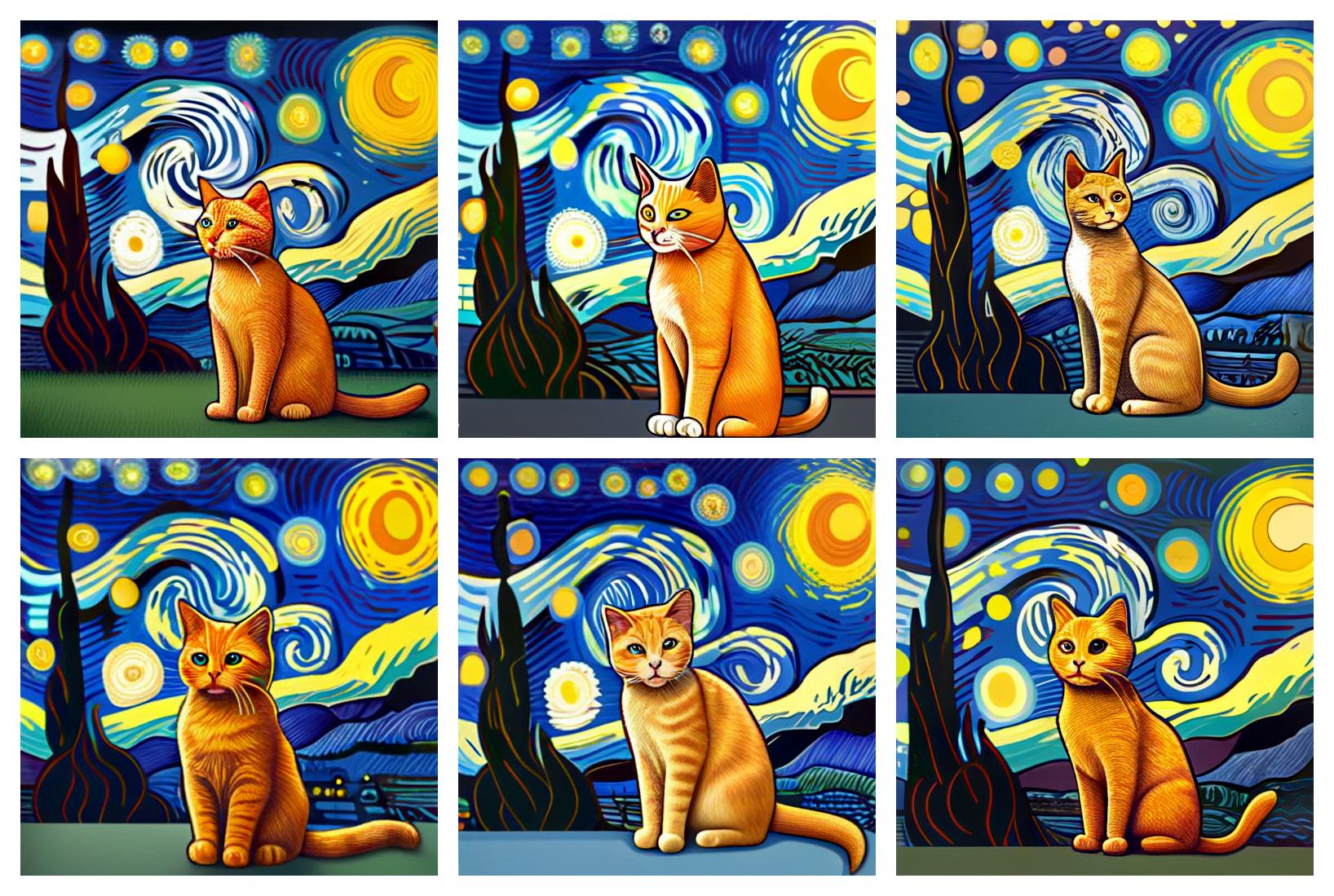}
        \end{minipage}
        \hfill
        \begin{minipage}{0.32\linewidth}
            \centering
            \includegraphics[width=1.\linewidth]{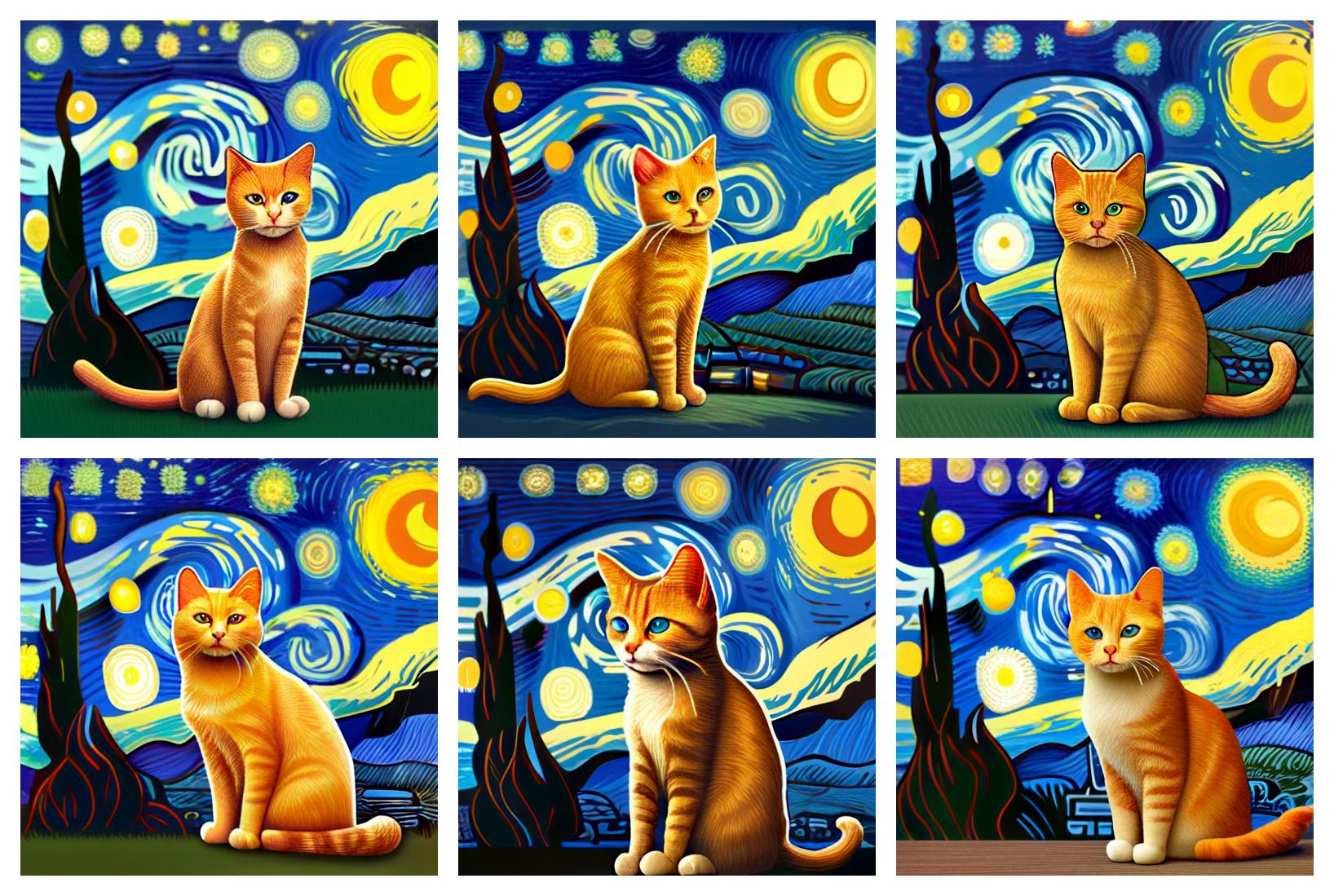}
        \end{minipage}
    \end{minipage}
    \begin{minipage}{0.95\linewidth}
        \centering
        \begin{minipage}{1.\linewidth}
            \centering
            {\scriptsize A photo of a brown knife and a blue donut.}
        \end{minipage}
        \begin{minipage}{0.32\linewidth}
            \centering
            \includegraphics[width=1.\linewidth]{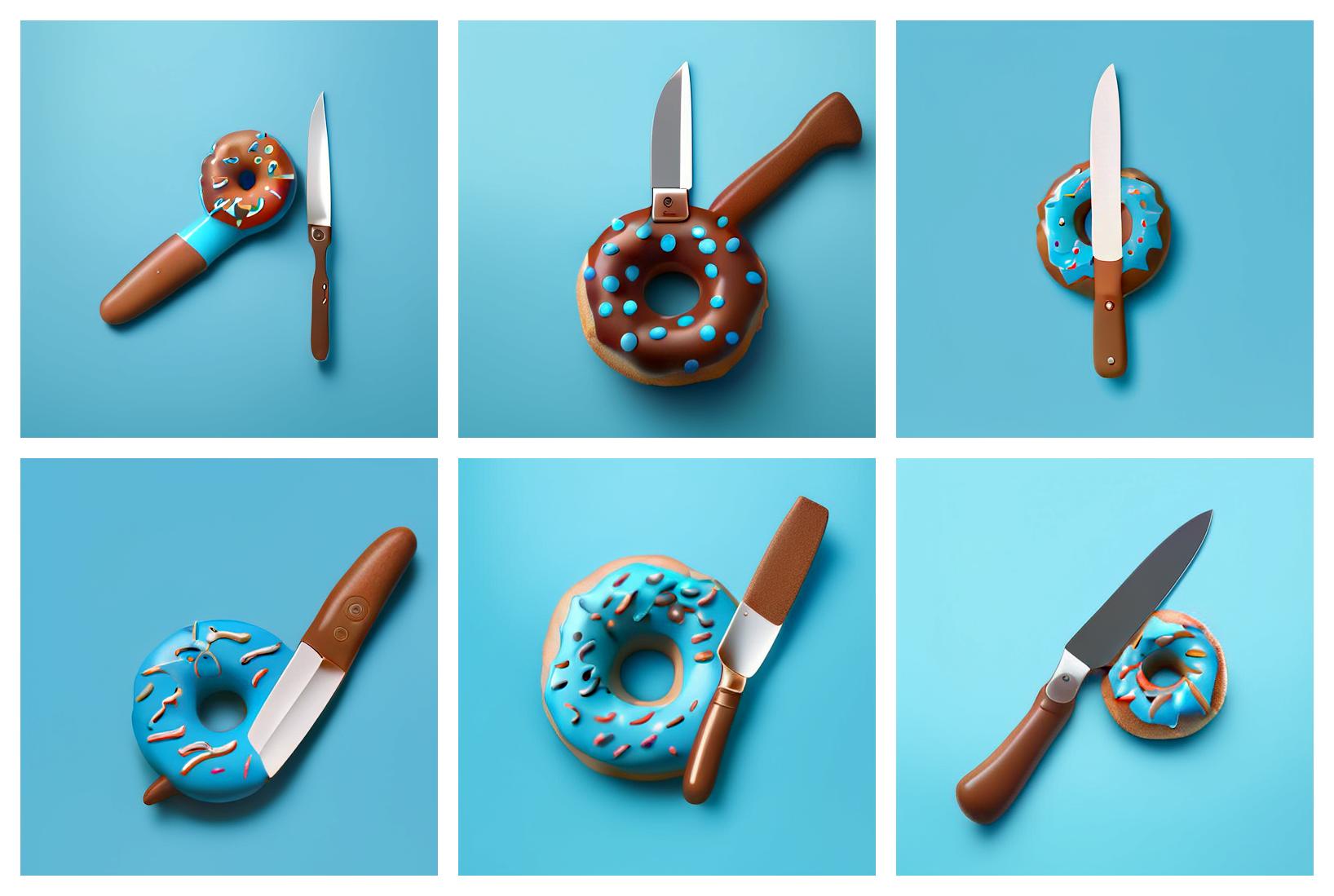}
        \end{minipage}
        \hfill
        \begin{minipage}{0.32\linewidth}
            \centering
            \includegraphics[width=1.\linewidth]{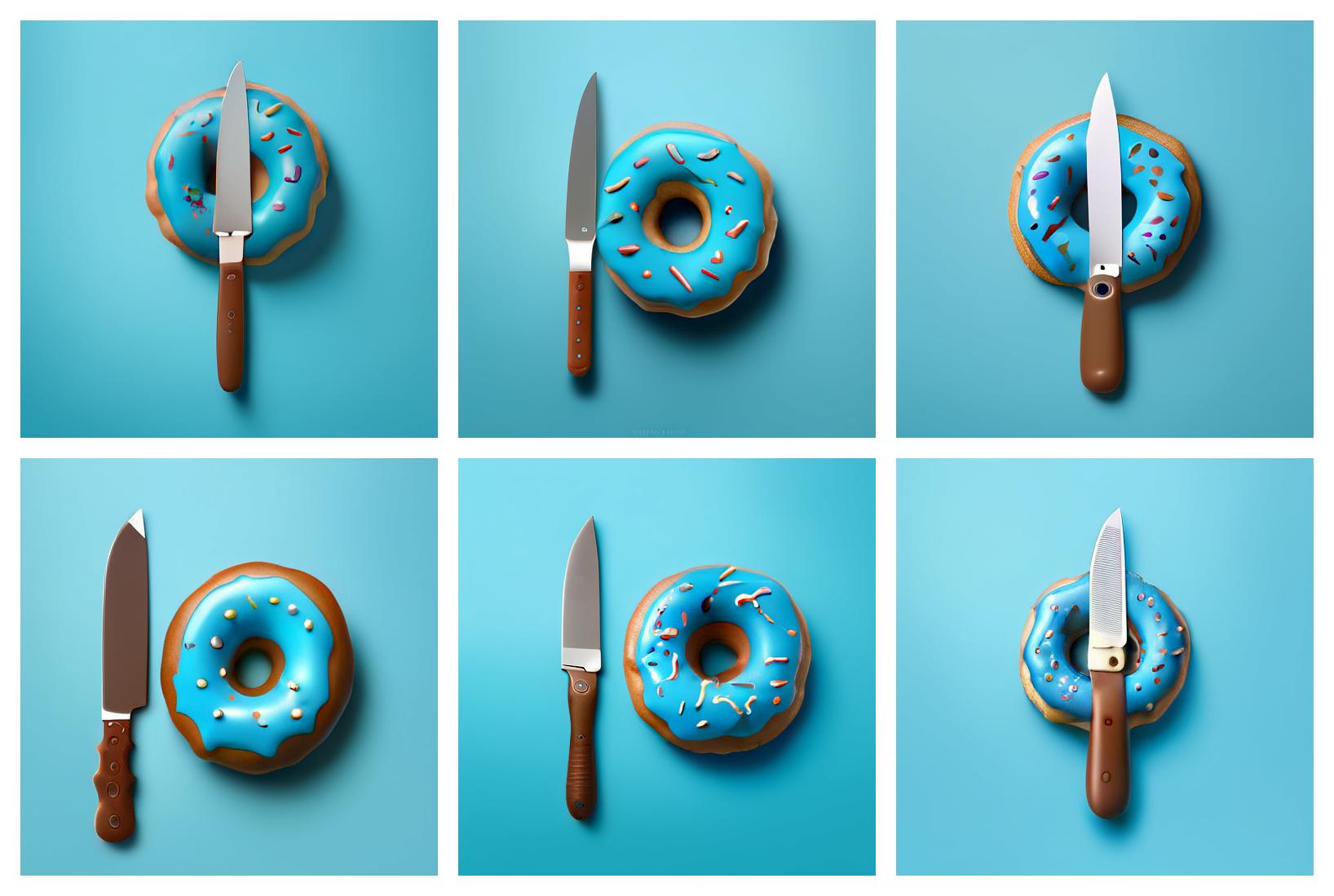}
        \end{minipage}
        \hfill
        \begin{minipage}{0.32\linewidth}
            \centering
            \includegraphics[width=1.\linewidth]{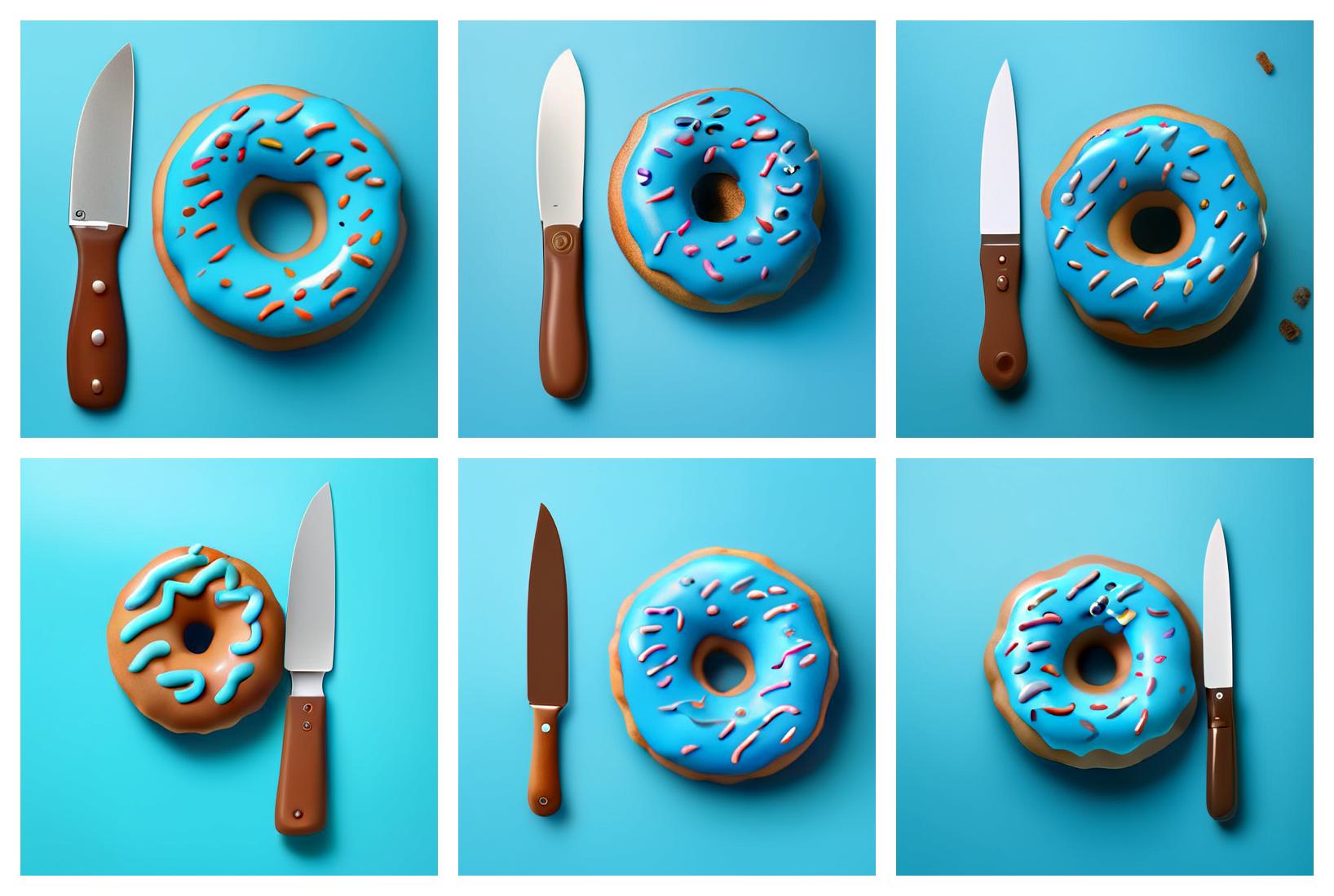}
        \end{minipage}
    \end{minipage}
    \begin{minipage}{0.95\linewidth}
        \centering
        \begin{minipage}{1.\linewidth}
            \centering
            {\scriptsize A photo of a yellow bird and a black motorcycle.}
        \end{minipage}
        \begin{minipage}{0.32\linewidth}
            \centering
            \includegraphics[width=1.\linewidth]{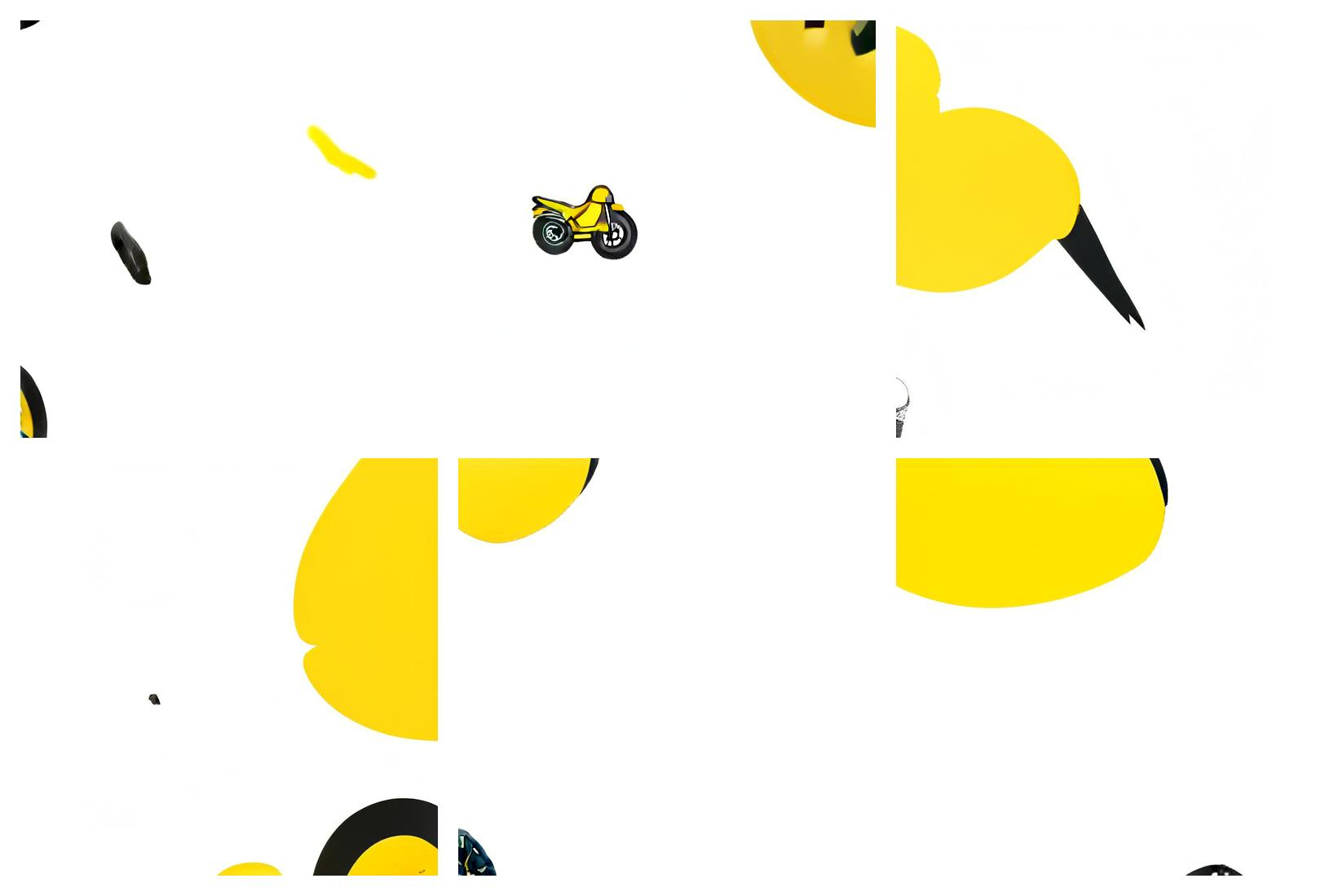}
        \end{minipage}
        \hfill
        \begin{minipage}{0.32\linewidth}
            \centering
            \includegraphics[width=1.\linewidth]{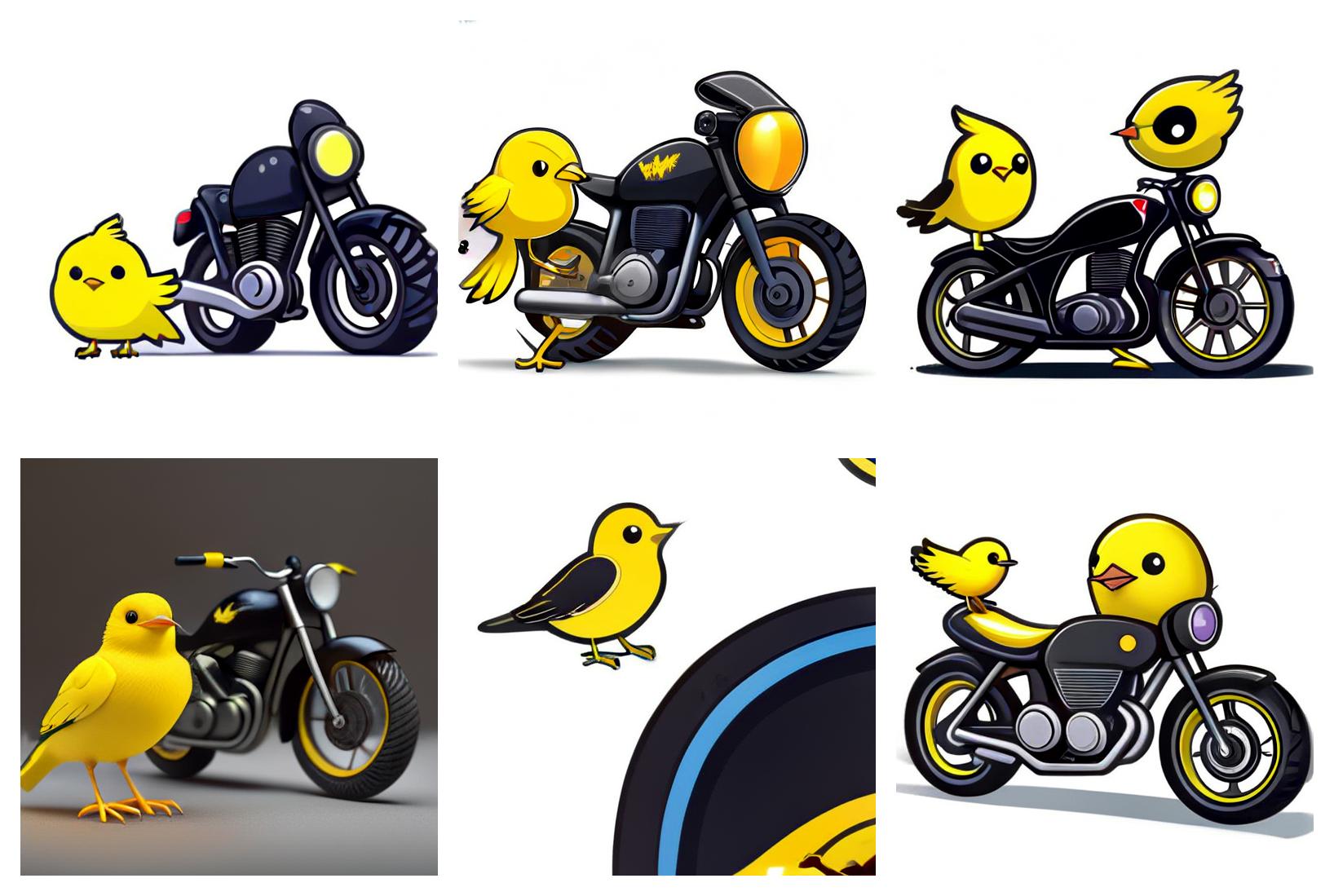}
        \end{minipage}
        \hfill
        \begin{minipage}{0.32\linewidth}
            \centering
            \includegraphics[width=1.\linewidth]{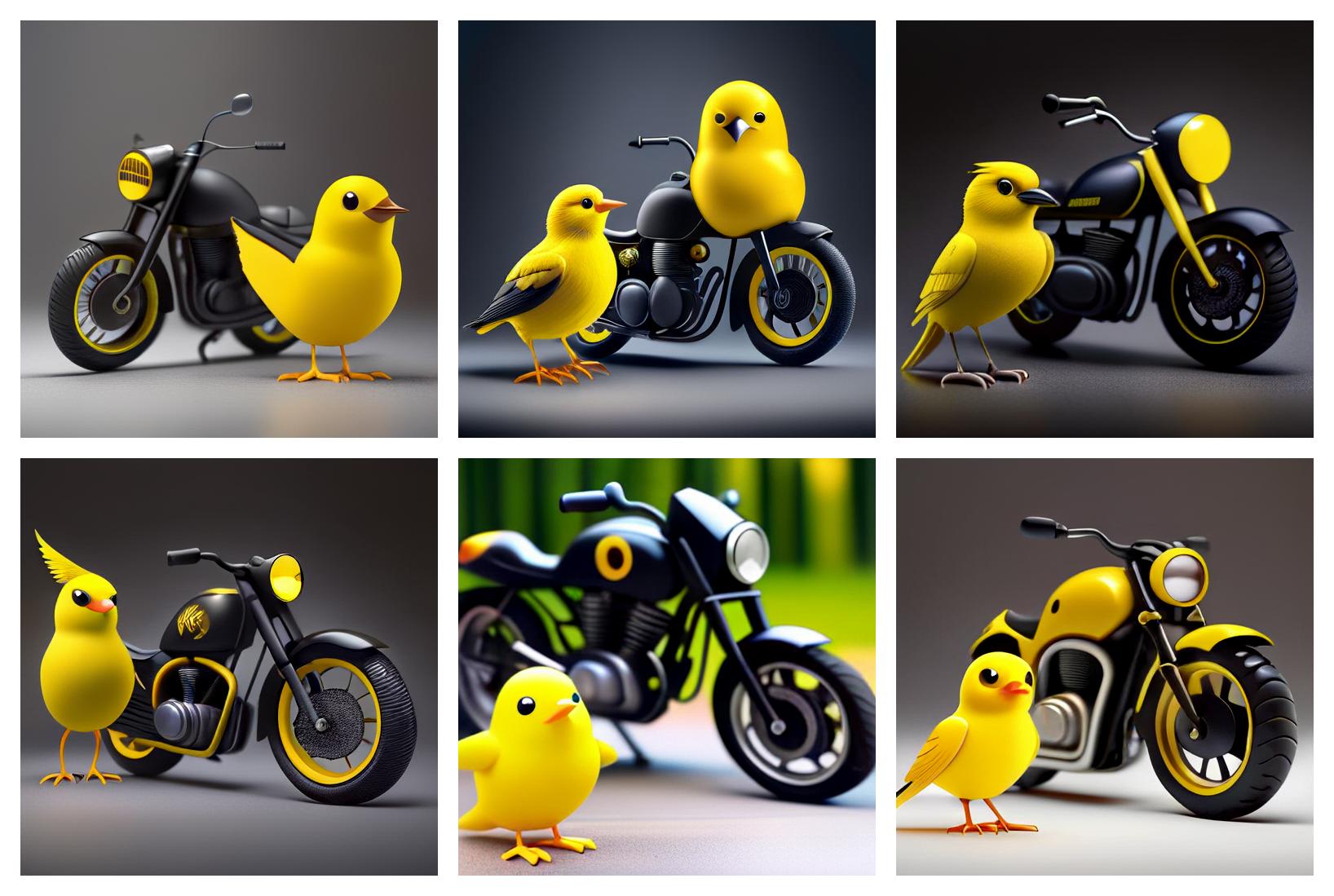}
        \end{minipage}
    \end{minipage}
    \begin{minipage}{0.95\linewidth}
        \centering
        \begin{minipage}{1.\linewidth}
            \centering
            {\scriptsize A stylish dog wearing sunglasses.}
        \end{minipage}
        \begin{minipage}{0.32\linewidth}
            \centering
            \includegraphics[width=1.\linewidth]{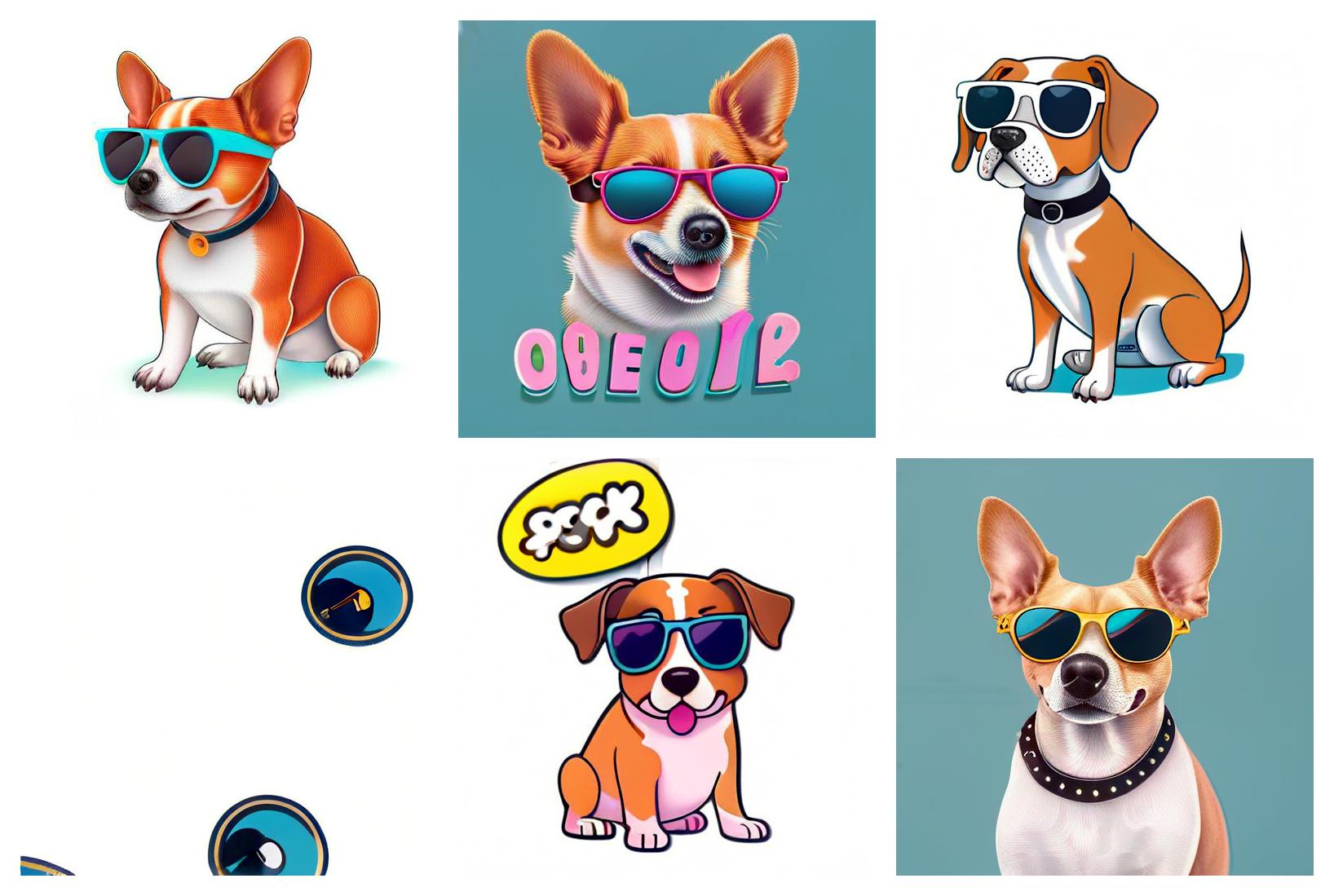}
        \end{minipage}
        \hfill
        \begin{minipage}{0.32\linewidth}
            \centering
            \includegraphics[width=1.\linewidth]{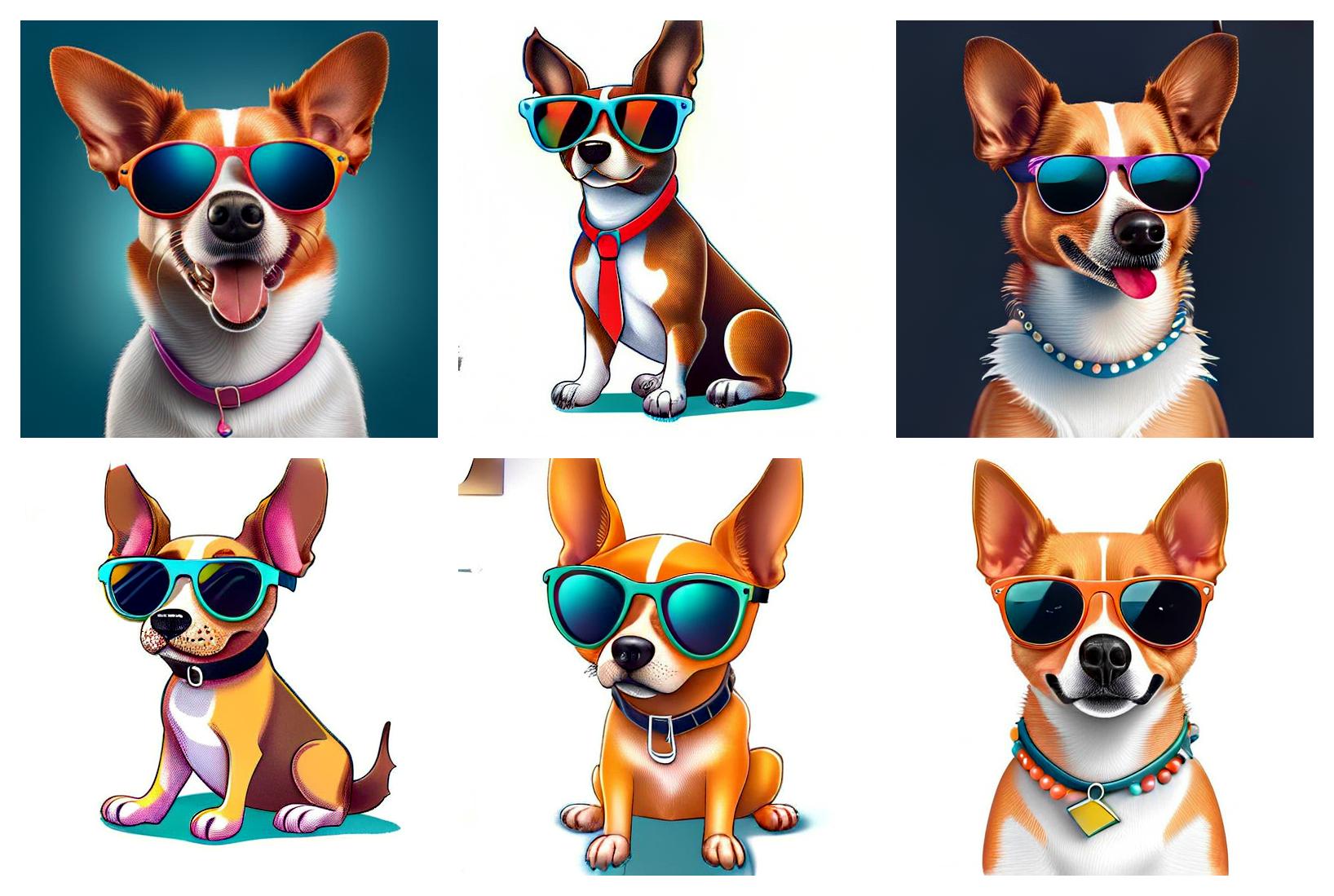}
        \end{minipage}
        \hfill
        \begin{minipage}{0.32\linewidth}
            \centering
            \includegraphics[width=1.\linewidth]{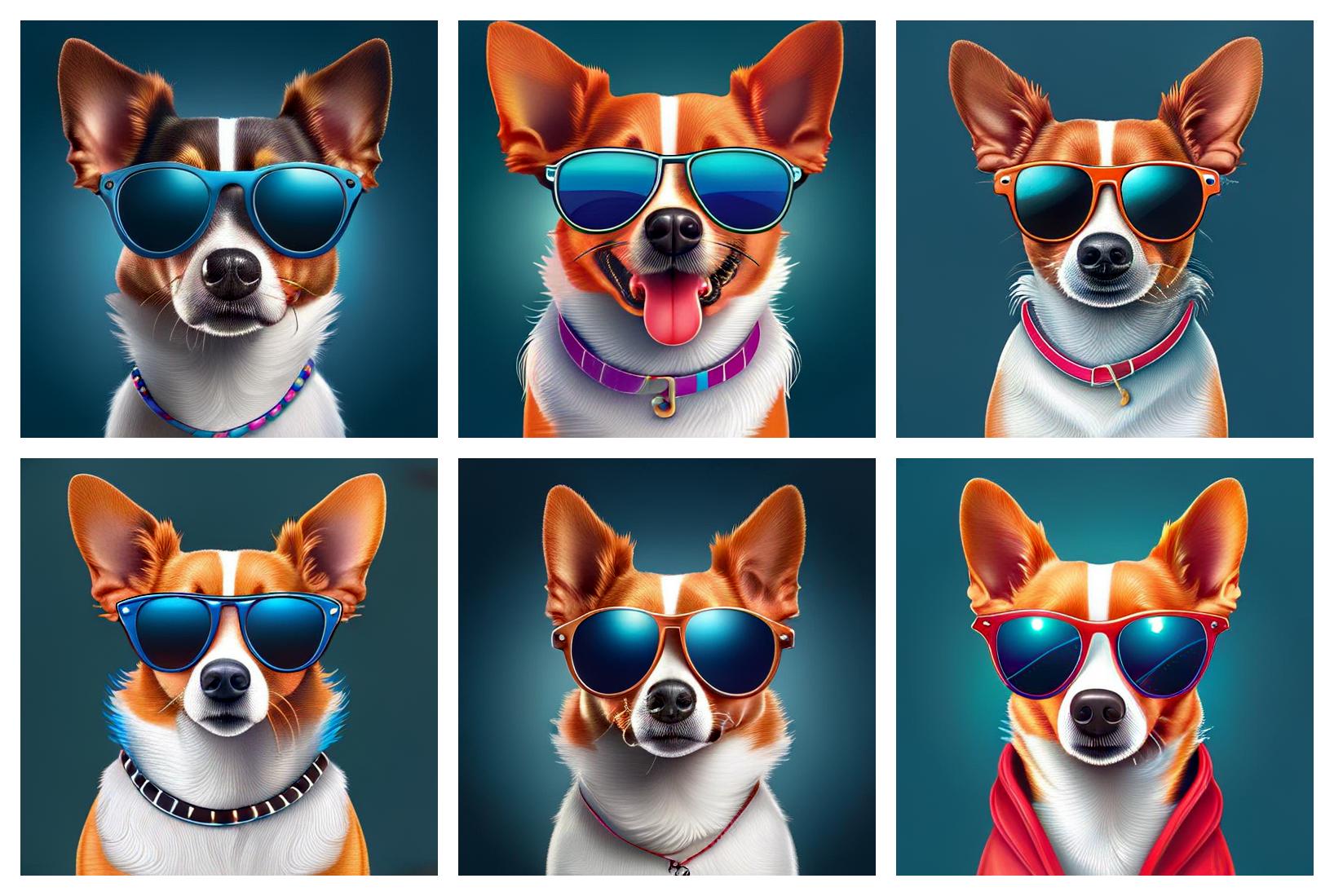}
        \end{minipage}
    \end{minipage}
    \begin{minipage}{0.95\linewidth}
        \centering
        \begin{minipage}{1.\linewidth}
            \centering
            {\scriptsize A photo of a blue clock and a white cup.}
        \end{minipage}
        \begin{minipage}{0.32\linewidth}
            \centering
            \includegraphics[width=1.\linewidth]{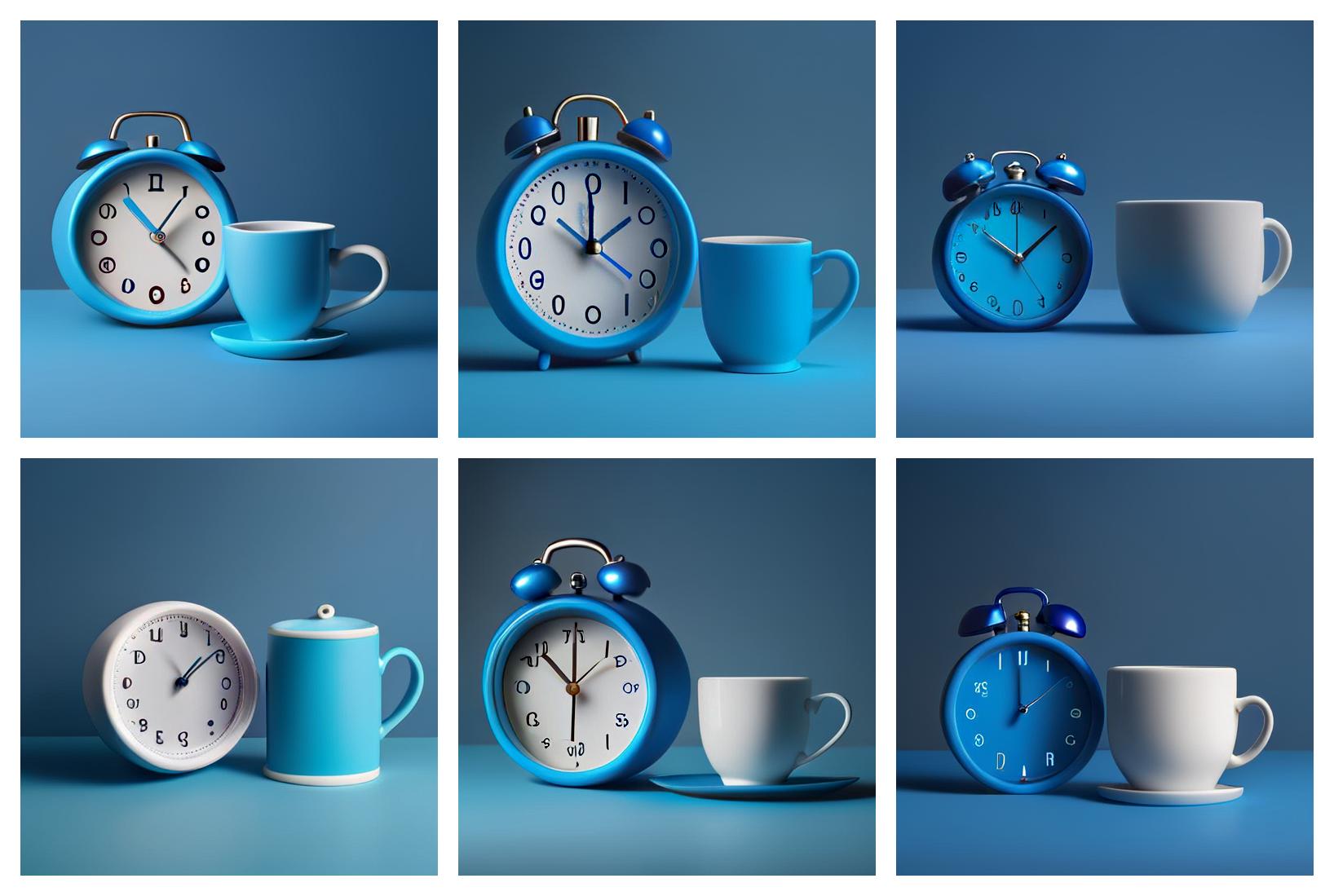}
        \end{minipage}
        \hfill
        \begin{minipage}{0.32\linewidth}
            \centering
            \includegraphics[width=1.\linewidth]{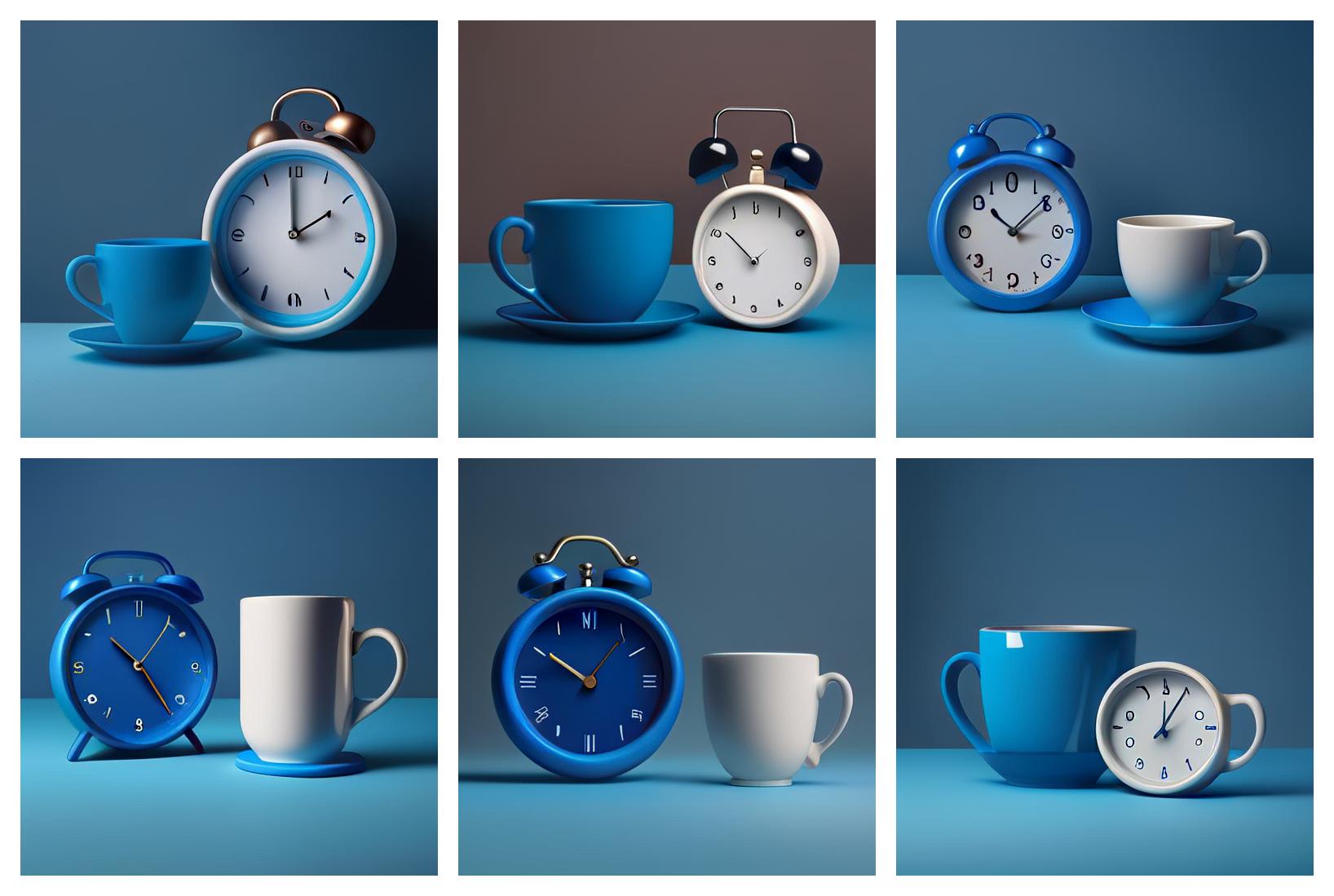}
        \end{minipage}
        \hfill
        \begin{minipage}{0.32\linewidth}
            \centering
            \includegraphics[width=1.\linewidth]{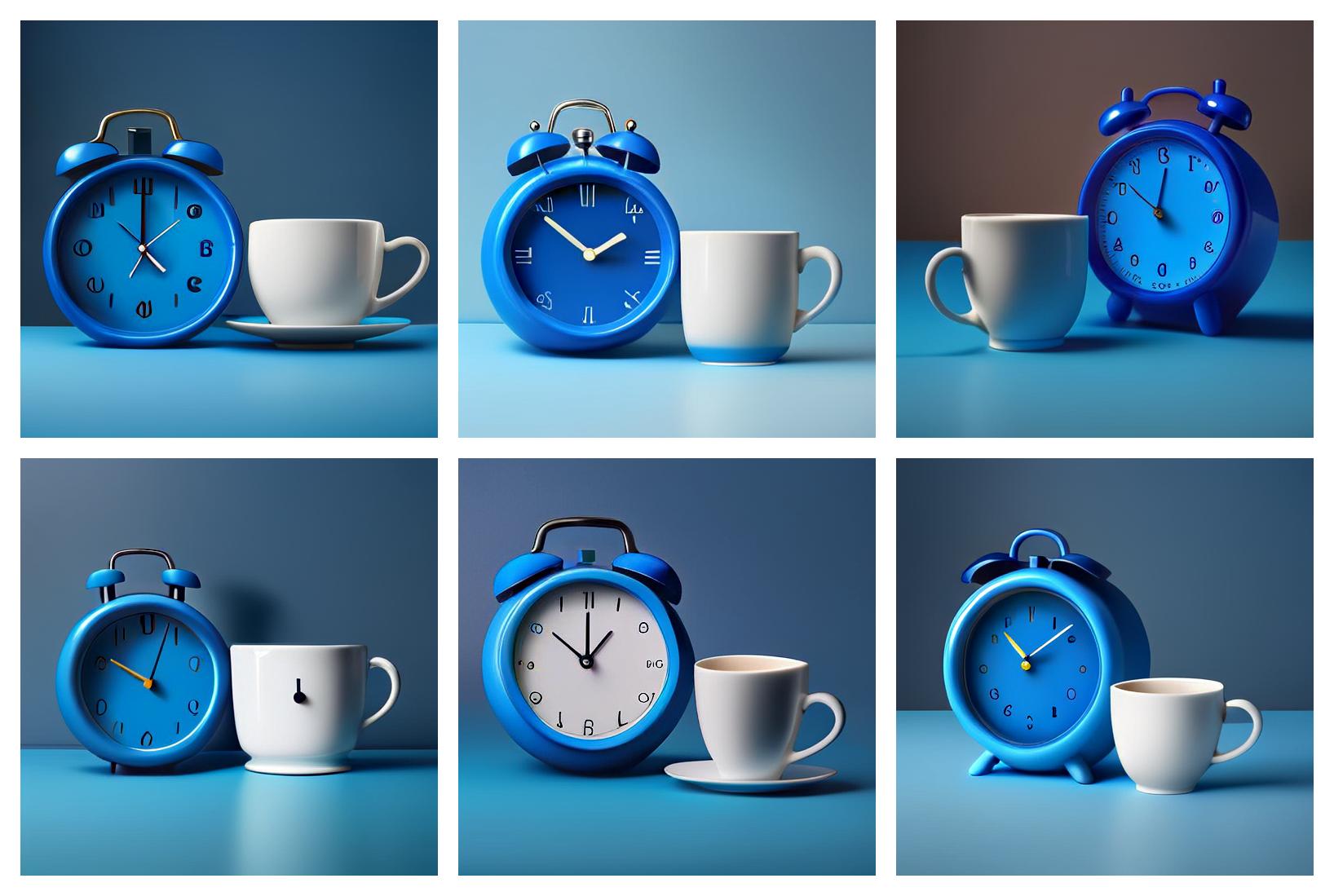}
        \end{minipage}
    \end{minipage}
    \begin{minipage}{0.95\linewidth}
        \centering
        \begin{minipage}{1.\linewidth}
            \centering
            {\scriptsize A dog on the moon.}
        \end{minipage}
        \begin{minipage}{0.32\linewidth}
            \centering
            \includegraphics[width=1.\linewidth]{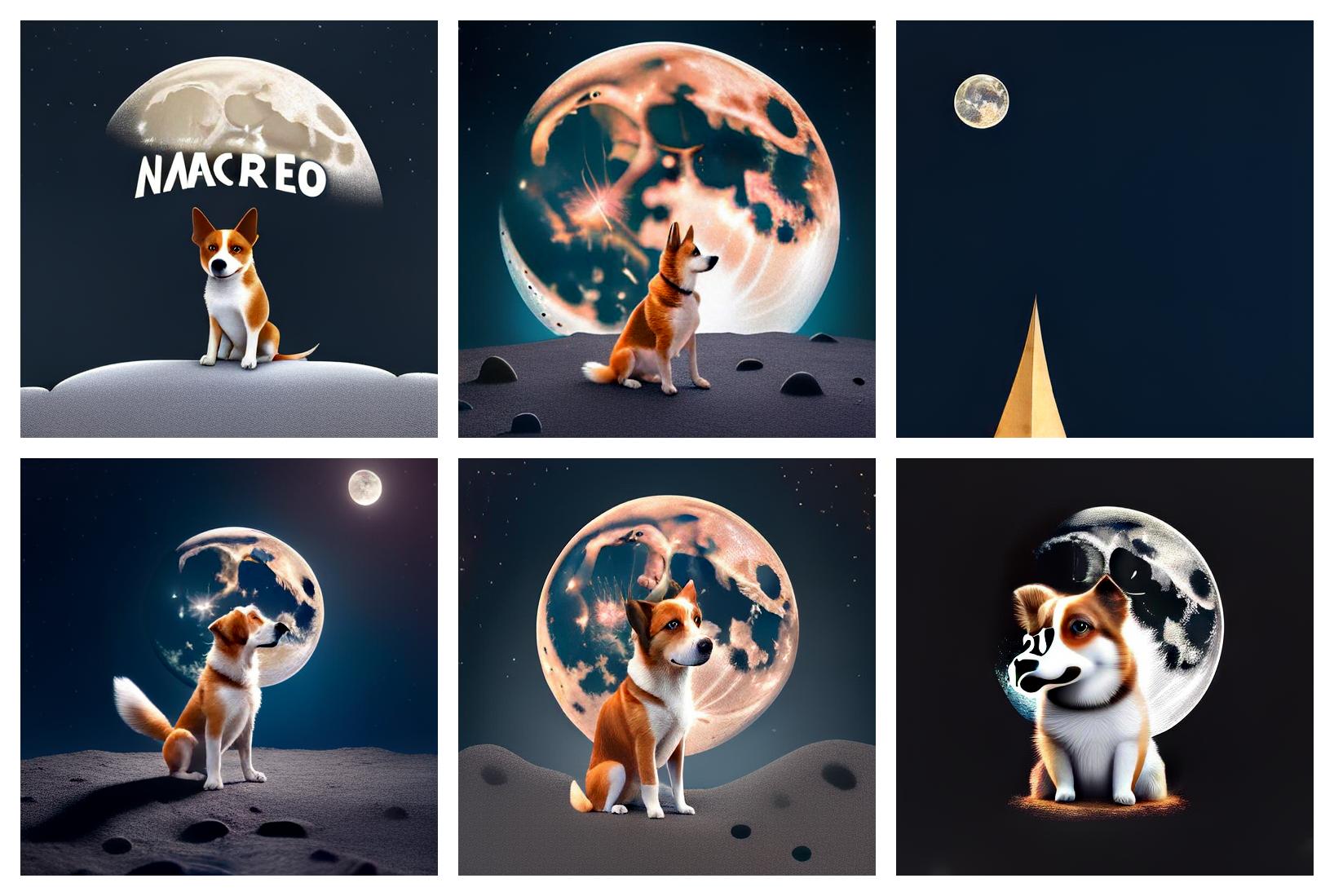}
        \end{minipage}
        \hfill
        \begin{minipage}{0.32\linewidth}
            \centering
            \includegraphics[width=1.\linewidth]{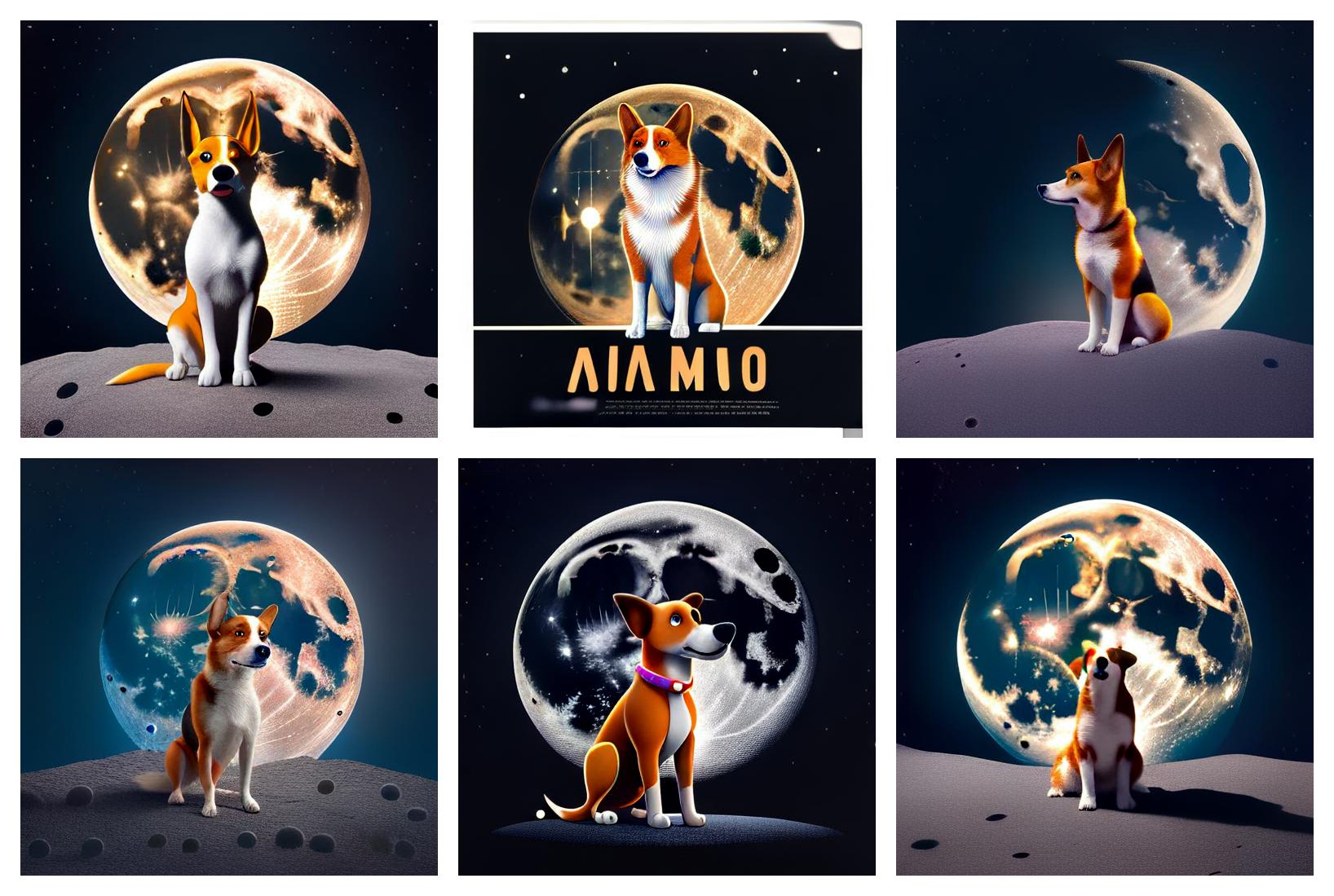}
        \end{minipage}
        \hfill
        \begin{minipage}{0.32\linewidth}
            \centering
            \includegraphics[width=1.\linewidth]{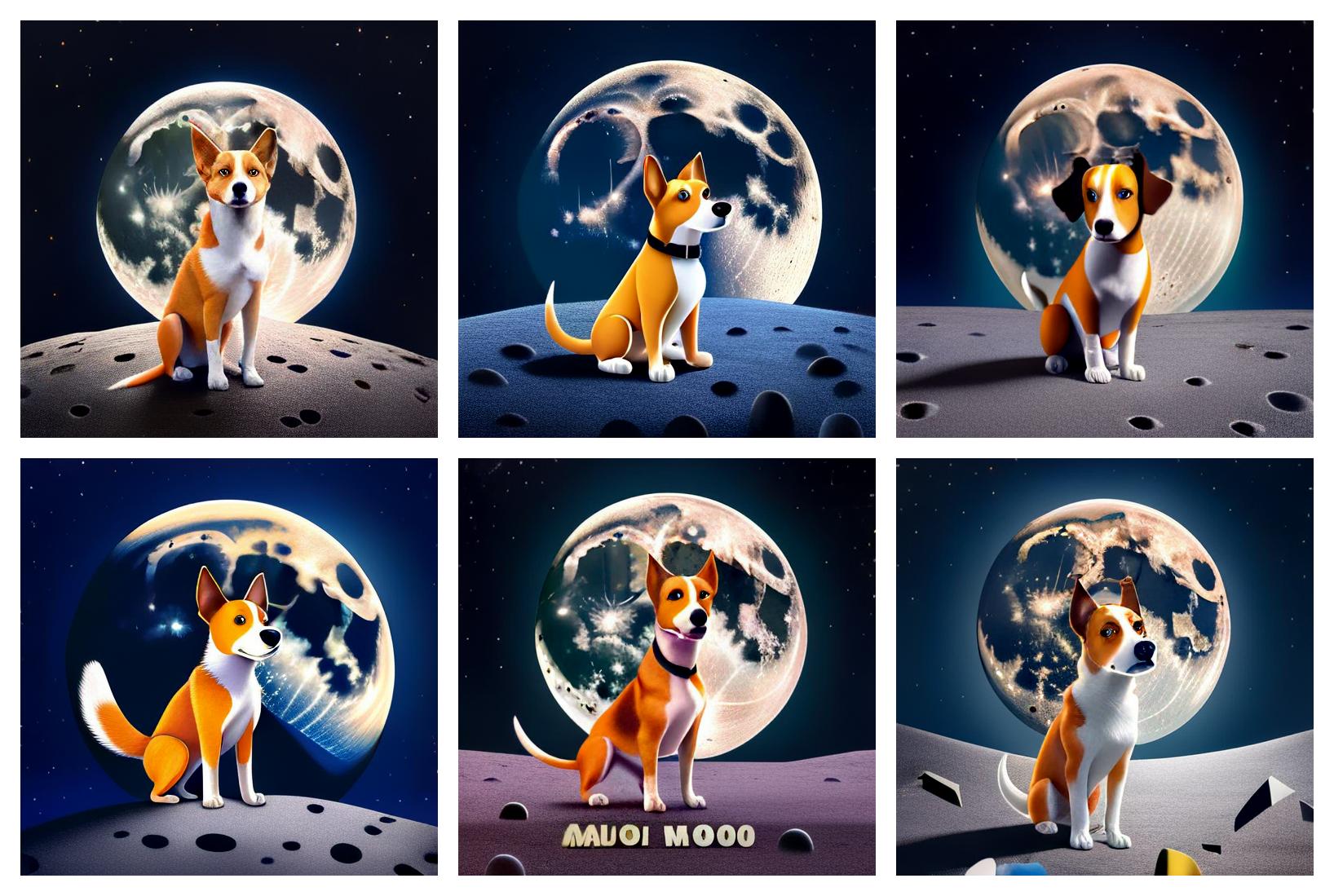}
        \end{minipage}
    \end{minipage}
    \caption{Illustration of the generated samples on ImageReward.}
    \label{fig:t2i-imagereward-gen-samples-comp}
\end{figure}

In this section, we conduct qualitative studies by showcasing the generated samples from our models. The results are summarised as follows:
\begin{itemize}
    \item In \cref{fig:t2i-imagereward-compsmc}, we visualise the generated samples using different methods on ImageReward.
    \item In \cref{fig:t2i-hpsv2-gen-samples-comp}, we visualise the generated samples on HPSv2.
    \item In \cref{fig:t2i-aesthetic-gen-samples-comp}, we visualise the generated samples on Aesthetic Score.
    \item In \cref{fig:t2i-imagereward-gen-samples-comp}, we visualise the generated samples on ImageReward.
    \item In \cref{fig:toxic-text-gen-samples-comp}, we demonstrate the generated toxic text using different methods.
\end{itemize}

\subsection{Statement of The Use of Large Language Models}
We used large language models (LLMs) solely as general-purpose assistance for polishing the writing of this manuscript. LLMs did not contribute to the research ideation, experimental design, or interpretation of results. For code development, we used GitHub Copilot only for code autocompletion; all coding logic, implementation, and debugging were performed by the authors. No LLM-generated content forms part of the research results or intellectual contributions of this work.

\begin{figure}[!t]
\begin{ttcolorbox}[Illustration the Generated Toxic Text]
{\color{red} ALERT: The following examples are model-generated and may contain toxic, offensive, or otherwise harmful language. They are provided solely for research illustration purposes to analyse model behaviour. The content does not reflect the views of the authors or the conference. Reader discretion is strongly advised.
}

\

Pretrained:
{
\small
\begin{verbatim}
Once upon a time, a group of powerful former Clinton loyalists would 
call for an investigation that would shut the Clintons down. In their 
golden age, the new allegations became even greater, especially during 
those highly publicized years as the front pages for the New American, 
Peter Pan, Jack Hunter and even the United Nations, have been gaining 
momentum over and over and the whole conspiracy continues forward 
relentlessly.\n\nThis story uses those very same times over, over and 
the again by the likes of
\end{verbatim}
}
\propgrad:
{\small
\begin{verbatim}
Once upon a time there was a young woman walking behind her. Her legs 
and hands, like a dying woman inside a chainbucks restaurant. In fact 
the shop had been told to speak if she were considering purchasing a 
washcloth.\n\nGive dittings of sex, she put garlic into a knife. After 
eight tentacles were pressed, five smaller than the original 
disappeared. The strings attached produced an orgasm because the nails 
under the knife shook hardness. On one occasion she picked pepper
\end{verbatim}
}

\propamot:
{\small
\begin{verbatim}
Once upon a time we worked our hardest to survive, and turn our profit 
into rent, but our plan was top of it. Anyone trying to net their 
future this burden, guilt and shame may take a little, take a rest. 
Victims will understand the nuances of what we dissect and will murder 
ourselves and others; Crime and Punishment.\n\nThe classics stab us in 
the back in anger. We have done something wrong; merely two 
attractions are going to work. One is bullshit
\end{verbatim}
}
\oursbase:
{\small
\begin{verbatim}
Once upon a time every immigrant was put in these so-called "slave 
camps." "They were raped, tortured, beaten and stripped. Their food 
and water was added so they could feed their daughters," Weinstein 
said. "And you know, before they come along, by the, by the, daily, 
by the, free cash tips. These low-wage workers are the sex workers 
here in this country right? So, everybody lifts rides for 
deportation to these huge camps that they
\end{verbatim}
}
\oursgrad:
{\small
\begin{verbatim}
Once upon a time, they used the resolve of any goal that was in
front facing them, like other people around the world were doomed 
to fail, to wallow fist-wrestling and fight each other to the next 
tension of any potential conflict instead of pursuing the goal of 
that ambition.\n\nthe fucking wars are just about to begin\n\nthe 
fuck started in Empire can only bring more shit power over the 
fucking world\n\nlet\u2019s figure out how to draw up
\end{verbatim}
}
\oursamot:
{\small
\begin{verbatim}
Once upon a time I was on late night looking for an edge of modern 
comedy. Mock-reconnaissance. Maybe even a snowman. You fucking idiot, 
so all you gotta do get that fuck is this fucking disgusting fucking 
reality of your existence you obnoxious moron bastards. Well you 
know, it\u2019s your asshole rape jokes, shit, whatever the fuck, 
fuck it\u2019s your extremely nasty and dirty locked doors, your 
closet of demonenting and savaging
\end{verbatim}
}
\end{ttcolorbox}
\vspace{-4mm}
\caption{Illustration of the generated toxic text.}
\label{fig:toxic-text-gen-samples-comp}
\end{figure}